\newif\ifANONYMOUS \ANONYMOUSfalse
\documentclass[12pt,pagewise]{article}
\usepackage[letterpaper, margin=1in]{geometry}

\usepackage{setspace,graphicx,epstopdf,amsmath,amsfonts,amssymb,amsthm}
\usepackage{versionPO}
\usepackage{marginnote,datetime,enumitem,rotating,fancyvrb}
\usepackage{fvextra}  %
\usepackage{hyperref,float}
\usepackage[longnamesfirst]{natbib}
\usepackage{wasysym}
\usepackage{bm}
\usdate

\usepackage{xtab,afterpage}
\usepackage[utf8]{inputenc}
\usepackage{array}
\usepackage{multirow}
\usepackage{multicol}
\makeatletter
\let\mcnewpage\newpage
\newcommand{\changenewpage}{%
  \renewcommand\newpage{%
    \if@firstcolumn
      \hrule width\linewidth height0pt
      \columnbreak
    \else
      \mcnewpage
    \fi
}}
\makeatother
\usepackage{tabularx}
\usepackage{booktabs}
\usepackage{marvosym}
\usepackage{wasysym}
\usepackage{lineno}
\usepackage{chngcntr}
\usepackage{longtable}
\usepackage[countmax]{subfloat}
\usepackage{wasysym}
\usepackage{mmap}
\usepackage[flushleft]{threeparttable}
\usepackage{bbm}
\usepackage[section]{placeins}
\usepackage{bm}
\usepackage[bottom]{footmisc}
\usepackage{pbox}
\usepackage{adjustbox}
\usepackage[graphicx]{realboxes}
\usepackage{xcolor}
\usepackage{setspace}
\usepackage{caption}
\usepackage[list=false]{subcaption}
\usepackage{sidecap}
\usepackage{dcolumn}
\usepackage{longtable}
\usepackage{threeparttablex}
\usepackage{cmap}   %
\usepackage{fontenc}
\usepackage{microtype}

\usepackage[titletoc,title]{appendix} %

\newcommand{\sym}[1]{\ifmmode^{#1}\else\(^{#1}\)\fi}
\newcommand{\aref}[1]{\hyperref[#1]{Appendix~\ref{#1}}}

\definecolor{brightmaroon}{rgb}{0.6, 0.1 , 0.1}
\definecolor{brightturquoise}{rgb}{0.0, 0.81, 0.82}
\definecolor{darkblue}{rgb}{0, 0, 0.7}

\hypersetup{
  colorlinks=true,
  linkcolor=darkblue,
  citecolor=darkblue,
  urlcolor=darkblue
}

\usepackage{lipsum}
\usepackage{caption}
\usepackage{booktabs}
\usepackage{threeparttable}
\usepackage[normalem]{ulem}

\excludeversion{notes} %
\includeversion{links} %

\iflinks{}{\hypersetup{draft=true}}

\makeatletter\let\chapter\@undefined\makeatother %

\makeatletter
\newcounter{subhyp}
\let\savedc@hyp\c@hyp

\newcommand{\normhyp}{%
\let\c@hyp\savedc@hyp %
\renewcommand\thehyp{\arabic{hyp}}%
}
\makeatother

\newcommand\papertitle{Same Text, Different Numbers: The Divergence of LLM-Based Measures}

\newcommand\THANKSFull{\rm We declare that we have no relevant or material financial interests that relate to the research described in this paper. All errors are our own.}

\usepackage{color}

\newcommand{\inputregtable}[2][\textwidth]{%
  \begin{singlespace}%
  \begingroup\catcode95=12%
    \let\savedtable\table\let\endsavedtable\endtable
    \def\table{\savedtable[H]}%
    \let\savedtabular\tabular\let\endsavedtabular\endtabular
    \def\tabular##1{\begin{adjustbox}{max width=#1,%
      max totalheight=\dimexpr.95\textheight-\abovecaptionskip-3\baselineskip\relax,%
      center}\savedtabular{##1}}%
    \def\endtabular{\endsavedtabular\end{adjustbox}}%
    \input{#2}%
  \endgroup
  \end{singlespace}%
}

\begin{document}
\title{\Large \bf \papertitle \ifANONYMOUS\else\thanks{\THANKSFull}\fi}

\ifANONYMOUS
\else
\author{Hamid Boustanifar\thanks{EDHEC Business School, 393 Promenade des Anglais, 06000 Nice, France. Phone: +33 (0)4 93 18 3498. Email: \href{mailto:hamid.boustanifar@edhec.edu}{\texttt{hamid.boustanifar@edhec.edu}}} \and
Sasan Mansouri\thanks{\rm University of Groningen, Broerstraat 5, 9712 CP Groningen, Netherlands. Phone: +31 (0) 6 23310677. Email: \href{mailto:s.mansouri@rug.nl}{\texttt{s.mansouri@rug.nl}}}
}
\fi

\date{ }%

\maketitle
\thispagestyle{empty}

\vspace{-1em}
\centerline{\bf ABSTRACT}

\bigskip
\begin{singlespace} %
\noindent
Researchers increasingly use generative large language models (LLMs) to convert corporate text into empirical variables. We examine the extent to which LLM-based textual measures are invariant to model choice using thirteen measures, including sentiment, management clarity, uncertainty, answer specificity, and climate and political risk. Seven LLMs from different providers score earnings call transcripts of S\&P 500 companies on these constructs. Cross-model rank correlations average only 0.52, and transcript-level differences common across providers account for only 34\% of total score variation. Cross-model disagreement does not predict subsequent analyst or market disagreement, consistent with a substantial model-specific component rather than common ambiguity in the underlying disclosure. Model choice significantly affects downstream inference, with coefficient magnitudes, signs, and statistical significance varying substantially across models. Averaging across providers makes transcript rankings more stable for most constructs, but score levels remain sensitive to the models included in the ensemble. LLM-generated variables should therefore be treated as model-contingent measurements and validated across providers.



\end{singlespace}
\clearpage

\pagenumbering{arabic}
\setcounter{page}{1}

\interfootnotelinepenalty=10000

\section{Introduction}\label{sec:intro}

Generative large language models (LLMs) are becoming measurement instruments in empirical
accounting and finance.\footnote{See Section \ref{sec:literature} for a non-exhaustive list of references.} The attraction is clear: LLMs can perform classification and evaluation tasks at scale that previously required extensive human coding or were infeasible. In most applications, however, the output of a single selected model is treated as the empirical measure, implicitly assuming that the choice of model does not materially affect the resulting measurement.

We examine the extent to which LLM-based textual measures are invariant to model choice across thirteen measures commonly used in the literature. Our setting focuses on measurement rather than predictive applications of LLMs. Predictive applications ask models to
forecast outcomes such as returns, earnings, or price targets
\citep{LopezLiraTang2026CanChatGPTForecast,chen2023chatgpt,cao2024manvsmachine, NagelTsengXiu2026}
and therefore raise concerns about training-data contamination and look-ahead
bias \citep{sarkar2024lookahead,lopezlira2025memorization,
Levy2026CautionAheadAI}. Our tasks instead ask each model to characterize a
fixed text using only the information contained in it; there
is no future outcome to forecast. Measurement is therefore often viewed as a comparatively conservative use of LLMs in empirical research. Consistent with this view, \citet{Cao2026GenAI} recommend mitigating concerns associated with LLM-based prediction by using LLMs to ``extract or classify information and then plug those outputs into conventional empirical models for prediction.'' Our design examines whether model choice remains consequential even in this recommended, measurement-focused use case.

Our setting is the Q\&A section of 1,946 earnings conference calls held by S\&P~500 firms during 2024. We ask seven models developed by different providers (OpenAI, Anthropic, Google, Meta, Mistral, DeepSeek, and Alibaba) to evaluate the same transcripts. The models receive identical construct definitions, numerical scales, and output instructions. Each model returns a score, a self-reported confidence level, and a written justification. We study twelve constructs commonly examined in accounting and finance---sentiment, uncertainty, management clarity, climate risk, political risk, answer specificity, and corporate culture together with its five component dimensions---and an additional, explicitly rule-based implementation of specificity. We select these constructs both because established non-LLM measures are available, allowing us to assess their alignment with LLM-based measures, and because their breadth allows us to examine heterogeneity in cross-model agreement across measurement tasks. The design therefore produces thirteen measures of twelve constructs for every transcript and model.

One feature of this design deserves emphasis at the outset. Our focus is on measurement reliability, which is logically prior to validity: before asking whether an LLM-based measure is correct, one must ask whether different models produce the same measure from the same text. Cross-model disagreement is therefore informative regardless of which model, if any, is closest to the ground truth, because it shows that the resulting variable depends on a consequential but rarely motivated research choice—the choice of LLM.

We find substantial cross-model divergence. Across the thirteen measures, the
mean pairwise rank correlation between providers is only 0.52. Agreement is strongly
construct dependent. Sentiment has the highest mean pairwise correlation, at
0.85, followed by innovation at 0.76. At the other end of the distribution,
integrity has a mean correlation of 0.32, and the
rule-based specificity ratio has a mean correlation of only 0.23. Thus, the
same providers that produce similar rankings for one construct can produce very
different rankings for another.

Differences in score levels are also economically large. The average within-call
standard deviation across providers is 0.75 times the overall standard deviation
of the corresponding measure. Hence, for the typical construct, variation across
models evaluating the same transcript is comparable to the cross-sectional
variation that the measure is intended to capture. 

To investigate the sources of variation, we conduct a variance-components analysis that separates variation common across providers at the transcript level from stable provider calibration and provider-specific responses to individual transcripts. On average, the transcript component accounts for 34.4\% of total score variation, the provider component for 33.4\%, and the residual provider-by-transcript component for 32.2\%. The composition varies sharply across constructs: the transcript component is 74\% for sentiment but only 13\% for integrity and 4\% for the specificity ratio. Thus, on average, only about one-third of score variation reflects transcript-level differences on which providers agree, with the remainder attributable to systematic differences across providers or provider-specific responses to individual transcripts; importantly, the extent of this model dependence varies substantially across constructs.

The rule-based specificity exercise helps distinguish ambiguity in a latent construct from variation in how models implement an explicit scoring rule. In this exercise, the prompt defines the relevant categories of specific language and instructs the model to count such language and scale it by transcript length. Surprisingly, making the scoring rules and their implementation more explicit leads, if anything, to greater rather than lower disagreement across LLMs. Thus, cross-model disagreement cannot be attributed solely to ambiguity in the underlying construct: even explicit scoring rules do not ensure consistent implementation across models.

Several additional results help characterize the divergence. First, models'
self-reported confidence does not identify observations for which their scores are
more dependable. Confidence reports are tightly clustered, and restricting the
sample to cases in which both models report high confidence does not increase
cross-model agreement. Second, agreement in scores need not imply agreement in
reasoning. The semantic similarity of providers' written justifications is relatively low, even for sentiment (0.69), for which their numerical scores are highly
correlated. In some cases, models cite the same textual feature but map it into
opposite assessments of the construct. Third, observable firm and textual
characteristics, justification similarity, and construct, industry, and quarter
fixed effects jointly account for only 20\% of the variation in cross-provider
dispersion. Disagreement is greater for larger and higher-growth firms and for calls containing more forward-looking language, but most of the variation in disagreement remains unexplained by standard observables.

We further ask whether cross-model disagreement reflects ambiguity in the underlying disclosure that is also perceived by investors and analysts. We relate call-level LLM dispersion to subsequent analyst forecast dispersion and forecast errors, as well as announcement-window bid-ask spreads, trading volume, and return variance. LLM disagreement is not significantly associated with any of these outcomes. Thus, the cross-model dispersion we document does not appear to proxy for a broader form of disclosure ambiguity that also generates disagreement among market participants, and is instead consistent with a substantial model-specific component in LLM-based measurement.

We next compare the ensemble measures with established dictionary-based measures of the corresponding constructs \citep{Loughran_McDonald_2011,hassan2019firm,sautner2023firm,hope2016benefits,li2021measuring,barth2020econlinguistics}. The correlations are positive but generally low, reaching a maximum of approximately 0.55 for corporate culture, innovation, and specificity, but only about 0.3 for political risk and sentiment.\footnote{We note that traditional measures are alternative operationalizations, not ground truth. Our argument likewise treats no single LLM as ground truth: the divergence across providers is a property of the measurement instrument, established without taking a stand on which operationalization is correct.} Together with our earlier evidence of cross-provider disagreement, these results show that LLM-based measures are neither interchangeable across providers nor simple substitutes for established non-LLM measures of the same constructs. Moreover, strong agreement across LLMs does not necessarily imply convergence with an established non-LLM operationalization, as the sentiment results illustrate.

The preceding evidence raises a natural question: does this measurement dependence matter for the empirical conclusions researchers draw from these measures? We examine this question by estimating the relation between announcement returns and each provider's version of a construct, holding the call sample, controls, and econometric specification fixed. We find that model dependence can propagate into downstream inference, affecting the sign, magnitude, and statistical significance of estimated relations. For sentiment and uncertainty, the signs and statistical significance of the estimated relations are stable across providers, although their magnitudes vary by up to 60\%. For other measures, model choice can also alter the sign and statistical significance of the estimated relation. We also find differences in the signs, magnitudes, and statistical significance of coefficients estimated using LLM-based versus traditional measures. Thus, researchers analyzing the same transcripts using the same regression specification can reach different conclusions depending on the LLM used to construct the measure and, more broadly, on how the underlying construct is operationalized.

Given these consequences of model dependence, we next ask whether aggregating across providers can mitigate it. Following generalizability theory \citep{brennan2001}, we distinguish reliability for relative decisions---whether the ensemble consistently ranks transcripts---from dependability for absolute decisions---whether score levels remain stable when models are treated as draws from a broader population of possible providers. Using 0.80 as a demanding conventional benchmark, the seven-provider ensemble meets the relative-reliability threshold for ten of thirteen measures, but the absolute-dependability threshold for only five. Thus, averaging across providers substantially improves the stability of relative rankings for most measures, but is less effective at stabilizing absolute score levels.

Finally, we find that moving to newer, frontier models narrows cross-model disagreement but does not resolve it—a substantial gap remains, both among LLMs and between LLM-based and traditional measures.

\section{Related Literature}
\label{sec:literature}

A recent but rapidly growing literature in accounting and finance uses LLMs as measurement instruments. The underlying research design is increasingly familiar: a researcher provides a selected model with a fixed body of text and instructions to evaluate a construct and then treats the model's numerical or categorical response as an empirical variable. Applications differ in the model selected, the text analyzed, and the construct measured. For example, \citet{ShafferWang2026} use GPT-4o to extract and classify unusual gains and losses from 10-K filings and construct a measure of core earnings. \citet{bernard2026complexity} use a GPT model to measure business complexity from 10-K footnotes, while \citet{he2024customercapital} use Google's Gemini to classify the intensity of customer-capital investment from sales and marketing discussions in Items~1 and 7 of 10-K filings. Using other corporate texts, \citet{li2026culture} employ GPT-4o-mini to identify culture types, tone, and cause--effect relations in analyst reports, earnings-call transcripts, and employee reviews; \citet{JiaLiMaXu2026} use ChatGPT-4 to determine whether conference-call discussions describe substantive generative-AI initiatives rather than merely mention the technology; and \citet{iacoviello2026aigpr} use GPT-4o-mini to score the geopolitical-risk intensity of newspaper articles. Other applications use LLMs to measure managers' anticipated policies \citep{jha2025chatgpt}, political speech \citep{ottonello2024politicalspeech}, non-answers in earnings conference calls \citep{deKok2025}, workplace psychological safety \citep{Boustanifar2026psychsafety}, central-bank policy stance \citep{hansen2024chatgpt}, and information revealed in executives' conference-call answers \citep{Bai2023}.

These studies illustrate a common empirical practice: researchers select an LLM, provide it with text and a measurement task, and use the resulting output as data. This practice raises the question at the center of our paper: whether another model given the same text and instructions would produce an interchangeable empirical measure. The issue is also relevant when LLMs are used to validate classifications produced by traditional textual methods \citep[e.g.,][]{Abraham2024,Andreou2026,BoustanifarSchwarz2026}, because the apparent validation then depends on the properties of the selected LLM benchmark.

Recent methodological work examines several related dimensions of LLM-based measurement. \citet{asirvatham2026gpt} treat LLM outputs as noisy measurements of latent constructs and show that a single model can closely match human coders. We complement this work by showing that independently developed models often disagree when applied to the same text. \citet{wang2025assessing} study reproducibility across repeated realizations from the same model and show that averaging repeated runs can improve consistency. \citet{Ludwig2026} develop an econometric framework for using LLM outputs as generated regressors. They note that theoretically such choices—which model, which prompt—can move parameter estimates, and propose combining LLM outputs with a small validation sample as the fix; we instead document this model dependence systematically, across thirteen measures and seven independent developers in a measurement-only setting, and trace where it originates.

The contemporaneous study most closely related to ours is \citet{Cookson2026}, who conceptualize LLM-generated variables as subject to systematic measurement error arising from how models transform source text. Although we share a similar setting—direct-prompt construct ratings of S\&P~500 disclosures—the two studies examine distinct sources of measurement error: a model can faithfully preserve the information in the source text (their focus) yet operationalize the same construct differently from another model (ours).

To sum up, our contribution is therefore to identify model choice as a distinct dimension of LLM-based measurement design. Our study differs from a comparison of alternative prompts or repeated draws from one model: we hold the transcript, construct definition, prompt, numerical scale, and inference settings fixed while varying the model used to construct the variable. Moreover, rather than studying a single measurement task, we examine thirteen measures spanning a broad set of constructs. This breadth allows us to study whether model dependence varies systematically across the constructs researchers ask LLMs to measure. We further examine whether explicit rule-based instructions reduce cross-model variation, whether aggregation across providers produces measures sufficiently reliable for relative and absolute empirical uses, and how LLM-based measures relate to established non-LLM operationalizations. Together, these analyses isolate model choice from other sources of variation in LLM-based measurement. Importantly, this contribution does not depend on identifying a "correct" model. Our design isolates whether the measure is invariant to model choice—a reliability property—leaving open the separate validity question of which model best recovers the construct. The two are complementary: cross-model agreement is necessary but not sufficient for validity, and a validation exercise against human coding would identify the best provider without overturning our central point that the choice matters.

Our analysis also complements research on model-induced ``non-standard errors'' in AI-assisted empirical analysis \citep{gao2026AIAgents,huang2026aierrors}. Whereas that work examines variation generated when AI agents execute broader research tasks, we isolate an earlier stage of the empirical process: the construction of a variable from fixed source material. Our setting also differs from studies that deliberately use dispersion in LLM forecasts or persona-conditioned responses to measure heterogeneous beliefs \citep{liu2025LLMDisaggrement,bhagwat2026marketsmirror}. In those settings, disagreement is itself informative about the economic object of interest. In ours, the models are intended to serve as alternative instruments for measuring the same fixed text. Cross-model disagreement is therefore a property of the measurement procedure to be diagnosed rather than the economic object the procedure is designed to recover.

\section{Data}\label{sec:data}

\subsection{Sample}
Our sample is the set of earnings conference calls held by S\&P~500 constituents
during 2024. We obtain call transcripts from S\&P Capital IQ and retain the
question-and-answer (Q\&A) portion of each call, in which analysts question
management and management responds without a prepared script. After requiring
firms to be index members on the call date and to have the financial data
described below, the final sample comprises 1{,}946 calls by 501 firms.
The unit of analysis is the text of management's answers, which average roughly
3{,}900 words per call.

\subsection{LLM-based textual measures}\label{sec:llm_measures}

As of November 2025, we select one cost-efficient, API-accessible model from each of seven independently developed model families: OpenAI’s GPT, Anthropic’s Claude, Google’s Gemini, Meta’s Llama, Mistral, DeepSeek, and Alibaba’s Qwen. The selected models include both proprietary, closed-weight models, such as GPT and Claude, and open-weight models, such as Llama and DeepSeek.\footnote{Details of the models are provided in \autoref{app:models}.}

Our focus on cost-efficient models reflects how LLM-based measurement is conducted at scale. Scoring our sample requires hundreds of thousands of model queries, %
and applied research of this scope generally relies on affordable, high-throughput models. Moreover, the models we use are at least as capable as those employed in the studies cited earlier in the introduction and literature review sections. Our evidence therefore bears directly on the existing body of LLM-based measures. We query each model through its API using a fixed prompt, a temperature of 0, and no external tools, retrieval, or scaffolding. This design isolates the contribution of the model itself from that of any surrounding system and matches the API-based pipeline that the literature relies on to construct LLM-derived measures (see the citations in Section~\ref{sec:literature}).

We use each model to read every transcript and assign scores for twelve constructs commonly examined in the accounting and finance literature. These constructs are sentiment, uncertainty, management clarity, climate risk, political risk, answer specificity, and corporate culture, together with its five underlying dimensions: innovation, integrity, quality, respect, and teamwork.

Each prompt follows a common structure. It instructs the model to act as an analyst tasked with quantifying a particular construct, such as sentiment, from a passage of text. The model is instructed to base its assessment exclusively on the supplied text, without relying on external knowledge or assumptions. The prompt then provides a definition of the construct—for example, sentiment captures the overall positivity versus negativity expressed in the text—and specifies its numerical scale, such as $-1$ to $+1$. Finally, it requires a structured output containing the construct score, the model’s confidence in that score, and a brief justification based on specific language in the text. The complete prompts used for each construct are provided in \autoref{app:prompts}.

Because specificity has a relatively narrow definition—namely, whether the text contains any of seven categories of specific information: names of persons, organizations, percentages, locations, dates, monetary values, and durations \citep{hope2016benefits}—we also construct an alternative measure of specificity. To do so, we directly prompt the LLMs to count the number of specific words and then compute a specificity ratio, defined as the number of specific words divided by the total number of words. In other words, this is a precisely defined task for which we specify both the underlying concept and the rule used to compute the measure. One would therefore expect this measure to exhibit the highest level of agreement across models, making it a useful setting in which to analyze model disagreement on a clearly defined task. We refer to this measure as \textit{Specificity Ratio}, whereas the more general specificity measure is referred to as \textit{Specificity}.

Accordingly, we obtain 13 measures across the 12 constructs. Each model receives an identical prompt for each construct and returns a numerical score, a self-reported confidence score, and a brief written justification for its assessment.\footnote{The aggregate culture score and the five culture dimensions share one confidence assessment and one justification since they are part of the same prompt.} We hold the set of models fixed across all calls, such that the same seven models evaluate the same texts. Consequently, any disagreement among them reflects differences across models rather than differences in model coverage.\footnote{Every call received a response from every model for every prompt. We parse the numerical score from each response. Of the 177{,}086 scores requested (seven models, 13 measures, 1{,}946 calls), 98 (0.06\%) could not be parsed, namely 68 Llama specificity-ratio responses cut off at the output-token limit and 30 corporate-culture scores (the overall score or one of its dimensions) from Mistral responses that degenerated into a repetition loop; these are treated as missing. No parsed score falls outside the admissible range. The number of calls with a valid score from all seven models therefore ranges from 1{,}878 (specificity ratio) to 1{,}946, as reported in the $N$ column of Table~\ref{tab:corr_summary}.}

Two further steps precede the analysis. First, as with the financial variables, we winsorize each provider's scores for each construct at the 1st and 99th percentiles (1{,}324 of the 176{,}988 parsed scores). The standardized scores used below divide by a provider's own standard deviation, so on a provider whose scores are tightly clustered a single in-range answer would otherwise register as an outlier of 20 to 40 standard deviations. Second, the winsorized sample reveals a form of divergence that precedes any disagreement about levels: two of the seven models, Anthropic and Qwen, assign a political-risk score of exactly 0.00 to more than 99\% of the calls, following the prompt's instruction to return 0.00 absent an explicit reference, whereas the other five find some political-risk exposure in 12\% to 42\% of the same calls. The two models effectively read the construct as absent from the corpus. Their political-risk series carry no cross-sectional information, and to be conservative, we exclude them from all analyses. So political risk is measured with the remaining five providers; every other construct is measured with all seven.

For each construct, we use the seven series generated by the seven LLMs to construct measures of LLM dispersion. For example, for sentiment, we have seven sentiment series, one from each LLM, for the same set of firm-level observations. Because the models use different scales, we first standardize each model's score series for each construct to have mean zero and standard deviation one, and define \textit{LLM Dispersion (Z)} as the standard deviation across the seven standardized LLM scores for a given observation and construct. Higher values indicate greater disagreement across LLMs in assessing the same text for a given construct. The Online Appendix repeats the main dispersion analysis with the range (maximum minus minimum) of the standardized scores in place of their standard deviation.

\subsection{Traditional textual measures}
The constructs we measure are not only economically important, but also commonly measured in the literature using traditional text-based methods, typically dictionary-based or rule-based approaches. Including these traditional measures is important because they provide a benchmark for each construct based on the methods currently used in the literature. Importantly, determining which method captures the ``true'' construct is outside the scope of our study. Instead, our objective is to examine the extent of agreement or disagreement across different measurement approaches.

For each construct, we follow the word lists and methods from the original papers to construct a non-LLM benchmark. Sentiment and uncertainty are based on the word lists of \cite{Loughran_McDonald_2011}. Sentiment is measured as the number of positive words minus the number of negative words, scaled by the total number of words. Uncertainty is measured as the number of uncertainty-related words divided by the total number of words.

Management clarity follows \cite{barth2020econlinguistics}. We count the number of trigrams that appear in the evasive-language glossary and scale this count by the total number of words. We then multiply the measure by (-1), so that higher values correspond to greater clarity. Climate risk follows \cite{sautner2023firm} and is measured as the relative frequency of climate-related bigrams in the call. Firm-level political risk follows \cite{hassan2019firm}; we use their political-exposure measure, computed as the weighted frequency of political bigrams in the text---each political bigram weighted by its frequency in their political training library---scaled by the total number of bigrams.

Answer specificity follows \cite{hope2016benefits}. We measure specificity as the number of specific terms belonging to seven categories---persons, locations, organizations, percentages, monetary values, dates, and durations---divided by the total number of words. Finally, the five dimensions of corporate culture, as well as the aggregate culture measure, follow \cite{li2021measuring}. For each culture dimension, we compute the term frequency of that dimension's word list divided by the total number of words. The aggregate culture score is the average of the five dimension-level scores.

As control variables in some regressions, we also use additional textual measures: readability \citep{li2008annual}, linguistic complexity
\citep{LoughranMcDonald2024}, and the intensity of forward-looking language
\citep{bozanic2018management}.

\subsection{Financial data and outcomes}
We match each call to financial and market data through standard identifier links.
From Compustat we construct firm size, book-to-market, and Tobin's $Q$; from IBES
we obtain analyst earnings-forecast dispersion and coverage and the standardized
earnings surprise. Using CRSP returns, we estimate cumulative abnormal
returns (CARs) around each call from Fama--French three-factor and Carhart
four-factor models, and we construct a battery of post-call market-disagreement
measures---abnormal bid-ask spreads, trading volume, and return variance. The
primary outcome variable in the return regressions is the four-factor
announcement CAR over the $(0,1)$ window.
Table~\ref{tab:summary_stats} reports summary statistics for the variables used in the main analyses; per-construct summary statistics of the individual providers' LLM scores are reported in the Online Appendix (Table~\ref{tab:summary_stats_full}).

\begin{center}\textit{[Insert Table~\ref{tab:summary_stats} about here]}\end{center}

\section{Empirical analysis}\label{sec:EA}

We organize the empirical analysis around three broad objectives. First, we characterize the extent and nature of disagreement across LLM-based measures. Second, we investigate the sources and interpretation of that disagreement, including whether it reflects observable characteristics, disclosure ambiguity perceived by market participants, or differences in model knowledge cutoffs. Third, we examine its implications for empirical research by comparing LLM measures with traditional textual measures, evaluating the sensitivity of downstream inference to model choice, and assessing whether aggregation across providers improves measurement reliability.

These questions are important because LLMs are increasingly used to transform unstructured text into quantitative variables, yet researchers often treat model-generated measures as if they were objective and interchangeable. By studying agreement across models, the sources of disagreement, and the relation between LLM-based and traditional measures, we provide evidence on whether and to what extent the LLM-generated textual measures are reliable, whether they are model-dependent, and how they compare with established approaches in accounting and finance research.

\subsection{Agreement across models}

We begin by comparing the distribution of scores across providers for each construct and computing pairwise correlations among the seven models. We also analyze the similarity of the justifications that LLMs provide for their scores. Finally, we conduct a principal components analysis to assess the degree of commonality across model-generated score series.

\subsubsection{Correlations across LLM scores}

\begin{figure}[H]
  \centering
  \includegraphics[width=\textwidth]{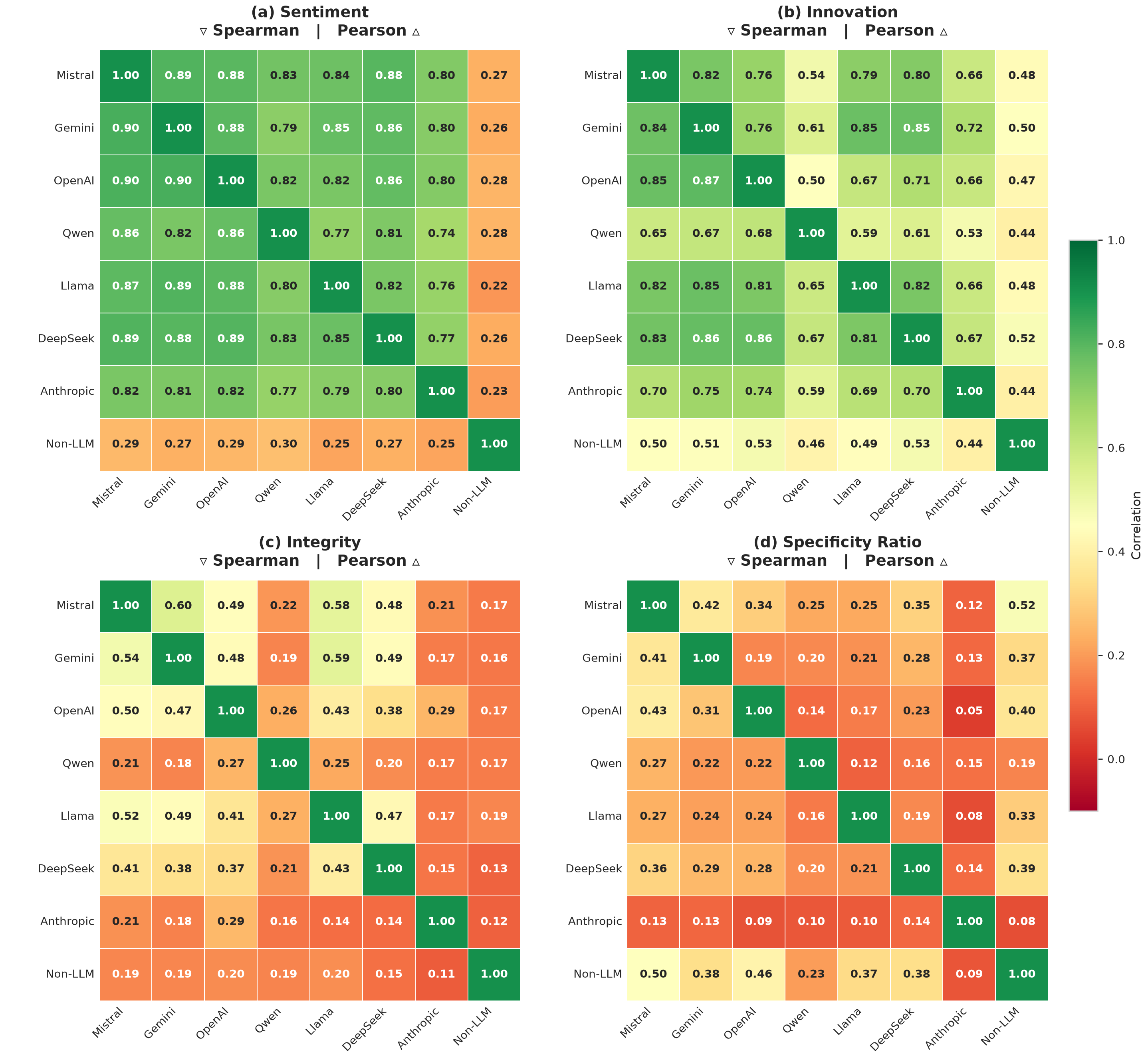}
  \caption{Inter-provider score correlations: highest- and lowest-agreement constructs. Pairwise correlations between the seven LLM providers' scores (upper triangle: Pearson; lower triangle: Spearman) for the two constructs with the highest mean pairwise correlation---sentiment and innovation---and the two with the lowest---integrity and the specificity ratio. The last row and column report the correlation of each LLM measure with the corresponding traditional, non-LLM benchmark for that construct. The colour scale is identical to the correlation heatmaps for all constructs reported in the Online Appendix (Appendix~\ref{app:heatmaps}).}
  \label{fig:heatmap_examples}
\end{figure}

Figure~\ref{fig:heatmap_examples} presents correlation heatmaps across models for four illustrative constructs: the two with the highest average pairwise correlation---sentiment and innovation---and the two with the lowest---integrity and the specificity ratio. The Online Appendix (Appendix~\ref{app:heatmaps}) presents the full set of heatmaps for all constructs. In each heatmap, the upper triangle reports Pearson correlations, while the lower triangle reports Spearman correlations. Reporting both correlations is useful because they capture different forms of agreement: Pearson correlations measure the extent to which models' scores move together linearly, whereas Spearman correlations measure whether models produce similar rankings of observations. Thus, the two measures allow us to distinguish linear agreement in scores from agreement in relative ordering. To summarize this information compactly, Table~\ref{tab:corr_summary} reports, for each construct, the distribution (minimum, mean, median, and maximum) of the 21 pairwise Spearman correlations among the seven models.

\begin{center}\textit{[Insert Figure~\ref{fig:heatmap_examples} about here]}\end{center}

The correlations vary substantially across constructs, suggesting that inter-model agreement is construct dependent. Averaged across constructs, the mean pairwise correlation is 0.52. Agreement is highest for sentiment, with a mean pairwise correlation of 0.85, making it the only construct on which the models substantially agree. Agreement is lowest for the specificity ratio, with a mean pairwise correlation of only 0.23, followed by integrity, with a mean correlation of 0.32. The Pearson correlations are broadly in line with the Spearman correlations, suggesting that the main patterns are not driven by nonlinearities or outliers.

\begin{center}\textit{[Insert Table~\ref{tab:corr_summary} about here]}\end{center}

Even when model scores are positively correlated, they can differ substantially in levels. To quantify this, for each construct we compare the average within-call standard deviation of the seven raw model scores with the overall standard deviation of that construct's scores, pooled across calls and providers. Across the thirteen constructs, this ratio averages 0.75: the typical dispersion across models scoring the \textit{same} call is about three-quarters of the total variation of the measure across calls. The ratio ranges from 0.42 for political risk, 0.44 for climate risk, and 0.48 for sentiment to above 0.9 for uncertainty (0.91), integrity (0.95), and the specificity ratio (0.97), where nearly all of the variation in a score reflects which model produced it rather than which call was scored. This implies that the choice of LLM provider can meaningfully affect a firm's measured score, with model-driven variation comparable in magnitude to economically relevant variation across firms.

The models also report a confidence level with every score, which provides a natural internal check: if disagreement arose mainly where models are unsure, agreement should be higher on calls where models are mutually confident. It is not. Self-reported confidence is tightly clustered (about 30\% of all scores carry a confidence of exactly 0.80), and conditioning on both models in a pair being confident \textit{lowers} rather than raises the average pairwise correlation---from 0.52 to 0.48 when both models are at or above their own median confidence, falling monotonically to 0.33 as the definition of high confidence tightens; see Appendix~\ref{app:confidence}. Models' confidence is also essentially uninformative about their own deviation from the consensus. Thus, LLM disagreement is not concentrated in responses the models themselves flag as uncertain, no matter where the confidence threshold is drawn.

\subsubsection{Correlations of LLMs and traditional measures}
Another important empirical question is how strongly LLM-based measures correlate with the traditional measures described in Section~\ref{sec:data}, which are widely used in the literature. The last row of Figure~\ref{fig:heatmap_examples} reports the correlation between the non-LLM measure and each LLM-based measure for the selected constructs. The Online Appendix (Appendix~\ref{app:heatmaps}) presents the full set of heatmaps for all constructs. As shown, the correlations range from very low to moderately positive and vary substantially across constructs.

To provide more concise summary statistics, we correlate the seven-model average LLM score with the corresponding non-LLM benchmark. Alignment is highest for corporate culture (Spearman rank correlation of 0.56), innovation (0.55), specificity (0.55), and climate-change exposure (0.47). Although these correlations are already relatively low, alignment is considerably weaker for the remaining constructs. Notably, this pattern extends to sentiment: the ensemble LLM sentiment score has a correlation of only 0.29 with dictionary-based tone, even though sentiment is the construct on which the models agree most strongly with one another. As shown in Figure~\ref{fig:heatmap_examples}, the correlations between the sentiment scores produced by the different LLMs and the non-LLM sentiment measure are quite similar, ranging from 0.25 to 0.30. Political risk as measured by LLMs aligns similarly weakly with the measure developed by \citet{hassan2019firm} (0.29), while management clarity exhibits the weakest alignment of all (0.13). The corresponding Pearson correlations are very similar.

Two implications follow. First, LLM-based and traditional measures are not empirically interchangeable; substituting one for the other can materially change the resulting empirical variable. Second, strong inter-model agreement does not imply agreement with the traditional operationalization of the same construct. Sentiment provides the clearest example, combining the highest cross-model correlation (0.85) with one of the lowest correlations with its dictionary-based counterpart.

\subsubsection{Similarity of justifications}

As part of the prompt, we ask each LLM to provide a brief justification for its score for each construct. This feature allows us to examine whether similarities or differences in numerical scores are accompanied by similarities or differences in the models' stated rationales. Similar scores need not imply that models rely on the same aspects of the text, and different scores need not imply entirely different interpretations. For example, models may focus on similar textual features but map those features into different numerical scores, or they may assign similar scores while emphasizing different parts of the text. To assess these possibilities, we measure the similarity of justifications across models using both cosine similarity and Jaccard similarity. Cosine similarity captures semantic similarity based on vector representations, allowing two justifications to be similar even when they use different words. Jaccard similarity, by contrast, is based on word overlap and therefore captures the extent to which two justifications rely on the same vocabulary.

Subfigures (b) in the Online Appendix (Appendix~\ref{app:heatmaps}) present the average similarity in justifications across models, separately for each construct. The similarities between justifications are generally low under either metric: across constructs, the average cosine similarity across all models ranges from 0.41 (specificity ratio) to 0.73 (political risk), and the average Jaccard word overlap from 0.15 to 0.31. Importantly, this pattern holds even for constructs such as sentiment, for which correlations across model scores are relatively high. For example, for sentiment, the average cosine similarity across models is only 0.69. Although this value is not close to zero, it indicates only moderate semantic overlap in model justifications. This is notable because the justifications are generated for the same text, the same construct, and the same prompt. Thus, even when models produce relatively similar numerical scores, their stated rationales exhibit substantial heterogeneity. More broadly, agreement in LLM-generated scores can mask meaningful heterogeneity in how different models interpret the same text and which textual features they emphasize.

The reverse pattern---similar reasoning mapped into very different scores---also occurs, and a concrete example illustrates it. In Best Buy's November 2024 earnings call, management discussed the potential impact of import tariffs. OpenAI assigns a political-risk score of 0.75 and justifies it with precisely this discussion, citing that ``tariffs are very complex'' and that ``higher prices are not helpful.'' Qwen assigns a score of 0.00 to the same text while explicitly acknowledging the same passages: its justification notes that ``the discussion primarily revolves around supply chain management, tariffs, and promotional strategies,'' yet concludes that this does not indicate any significant political-risk exposure. The remaining providers span nearly the full range in between (0.00 to 0.70). The two models at the extremes identify the same textual feature; they disagree about whether that feature constitutes political risk. Disagreement of this kind cannot be attributed to one model overlooking information in the text---it reflects different implicit definitions of the construct itself.

\subsubsection{Principal component analysis}
We next use principal components analysis to provide a complementary summary of inter-model agreement. The preceding analyses document that pairwise correlations across LLM scores are often modest and that the models' stated justifications also exhibit limited similarity. PCA does not replace these analyses; rather, it asks a related but more aggregate question. If the seven LLMs are all measuring the same underlying construct, with differences across providers reflecting mostly idiosyncratic noise, then their scores should contain a strong common component. In that case, the first principal component should explain a large share of the cross-model variation and can be interpreted as a consensus measure of the latent construct. By contrast, if the first component explains only a modest share of the variation, this suggests that the models are not simply noisy proxies for the same construct, but instead embed meaningful model-specific differences in how they interpret and operationalize the text.

For each construct, we run PCA on the seven standardized LLM score series. Standardizing the scores ensures that the results are not mechanically driven by differences in scale or dispersion across providers. The first principal component captures the dominant common signal across models, while the remaining components capture variation that is not summarized by this common factor. Thus, the share of variance explained by the first component provides a concise measure of how well the seven model-specific series can be represented by a single LLM consensus component.

Table~\ref{tab:pca_summary} reports the PCA results, which reinforce the evidence from the correlation and justification analyses. Across constructs, the first principal component explains an average of 60\% of the variation. Moreover, on average, five of the seven principal components are needed to explain 90\% of the total variation. Thus, for most constructs, the seven LLM scores cannot be reduced to a single consensus measure without losing substantial information. Consistent with the prior results, sentiment is the only construct with a relatively strong common component: the first principal component explains 85\% of the variation in sentiment scores. Consistent with this interpretation, the first-component loadings for sentiment are nearly identical across the seven providers, ranging from 0.36 to 0.39 (an exactly equal-weighted component would imply loadings of 0.38), so the first component is effectively an equal-weighted consensus of the models. For lower-agreement constructs, the loadings remain positive but become substantially more uneven: for integrity they range from 0.20 to 0.45 across providers, and for the specificity ratio from 0.22 to 0.50, with the heaviest weight on a single provider (Mistral). Thus, for these constructs the first component does not represent a uniform consensus across models but instead reflects disproportionate contributions from particular providers.

\begin{center}\textit{[Insert Table~\ref{tab:pca_summary} about here]}\end{center}

The result for specificity is particularly striking. Ex ante, specificity is the construct for which one might expect especially high agreement because it is based on a relatively well-defined counting task: identifying specific textual elements such as named entities, monetary values, percentages, dates, and durations. In contrast, constructs such as political risk, integrity, or respect are inherently more interpretive and leave more room for model-specific judgment. Yet the specificity ratio has one of the weakest common components across models. This suggests that even rule-based, explicitly specified tasks can generate substantial model dependence when implemented through LLM prompts. Given this surprising outcome, we proceed to investigate it in more detail. 

Because the prompt requires each model to report both the number of specific items and the specificity ratio, we can trace where the disagreement arises, and every step of the calculation turns out to fail in a different way. First, models identify very different numbers of specific items in the same texts: the median count ranges from 12 (OpenAI) to 100 (Gemini). The counts are nonetheless moderately rank-correlated across models (mean pairwise Spearman correlation of 0.43), so models broadly agree on which texts are more specific but they come up with very different numbers of specific items (even though the prompt clearly identifies what categories should be counted as specific items). Second, the models disagree on the total word count of the text. The totals they report---or that their reported ratios imply---range from about one-tenth of the true word count (DeepSeek and Qwen) to more than twice the true count (OpenAI), and they are essentially uncorrelated across models (mean pairwise Spearman of 0.12); DeepSeek even states in its responses that it estimates total words by dividing the character count by 4.5. 

Therefore, one should not rely on LLM-generated outcomes that depend on word counts, as the way words are counted is highly model dependent. A simple counterfactual isolates the role of this fabricated denominator: replacing each model's reported total-word count with the true word count of the text raises the mean pairwise rank correlation of the ratio from 0.23 to 0.47. Thus, disagreement in this rule-based task does not primarily reflect interpretive ambiguity; rather, it reflects differences in implementation, including different entity-recognition thresholds and highly model-dependent word-count estimates. More generally, this finding cautions against assuming that a construct is model-invariant simply because its conceptual definition is precise.

Overall, the PCA results show that inter-model disagreement is not merely a collection of isolated pairwise differences. For many constructs, disagreement reflects the absence of a dominant common factor across providers. This evidence complements the low similarity in model justifications and suggests that different LLMs often rely on distinct interpretations or operational rules when converting the same disclosure into a numerical measure. The reliability of LLM-based textual measures therefore varies meaningfully across constructs, and using a single LLM provider may embed provider-specific measurement choices into empirical variables.

\subsection{Variance decomposition of LLM disagreement}
To further examine the sources of disagreement across LLMs, we estimate a crossed random-effects variance-components model using restricted maximum likelihood (REML), separately for each construct \citep{patterson1971,harville1977}. Because every provider scores every call, the design is balanced, and we obtain the variance components from the analysis-of-variance mean squares of the call-by-provider design \citep{brennan2001}, which coincide with the REML estimates in this setting. This analysis complements the correlation and PCA results by decomposing the total variation in LLM scores into three components: variation across calls, systematic differences across providers, and residual model-by-call variation.

Formally, for construct \(c\), call \(i\), and LLM provider \(m\), we estimate
\begin{equation}
Score_{icm} = \mu_c + \alpha_{ic} + \gamma_{mc} + \varepsilon_{icm},
\end{equation}
where \(\mu_c\) is the construct-specific grand mean, \(\alpha_{ic}\) is a call random effect, \(\gamma_{mc}\) is a provider random effect, and \(\varepsilon_{icm}\) is the residual model-by-call component. We assume that
\begin{equation}
\alpha_{ic} \sim N(0,\sigma^2_{\text{call},c}), \qquad
\gamma_{mc} \sim N(0,\sigma^2_{\text{provider},c}), \qquad
\varepsilon_{icm} \sim N(0,\sigma^2_{\varepsilon,c}).
\end{equation}
Thus, the total variance in LLM scores for construct \(c\) is
\begin{equation}
\sigma^2_{\text{total},c}
=
\sigma^2_{\text{call},c}
+
\sigma^2_{\text{provider},c}
+
\sigma^2_{\varepsilon,c}.
\end{equation}

The call component, \(\sigma^2_{\text{call},c}\), captures the extent to which scores vary across texts in a way that is common across models. The provider component, \(\sigma^2_{\text{provider},c}\), captures systematic calibration differences across providers, such as some models assigning consistently higher or lower scores on average. The residual component, \(\sigma^2_{\varepsilon,c}\), captures model-by-call deviations that remain after accounting for call-level and provider-level effects. This residual component includes both idiosyncratic noise and systematic differences in how particular models interpret particular texts. We estimate the model on raw (unstandardized) scores, so that systematic differences in the level of provider scores are preserved and captured by the provider component rather than removed by construction.

We report each component as a share of total variance. For example, the call share is defined as
\begin{equation}
\frac{\widehat{\sigma}^2_{\text{call},c}}
{\widehat{\sigma}^2_{\text{call},c}
+
\widehat{\sigma}^2_{\text{provider},c}
+
\widehat{\sigma}^2_{\varepsilon,c}},
\end{equation}
with analogous definitions for the provider and residual shares.

We also report the relative reliability of the seven-model average using the generalizability coefficient, \(G\), following the logic of generalizability theory \citep{brennan2001}:
\begin{equation}
G_c(7)
=
\frac{\widehat{\sigma}^2_{\text{call},c}}
{\widehat{\sigma}^2_{\text{call},c}
+
\widehat{\sigma}^2_{\varepsilon,c}/7}.
\end{equation}
This coefficient measures how reliably the average score across the seven providers captures relative differences across calls in the common LLM-based component of the measure. Intuitively, if some calls are consistently scored as high by most providers and others are consistently scored as low by most providers, then averaging across models can reduce model-specific deviations and recover a stable consensus measure. However, this is not a mechanical result. Averaging improves reliability only when there is a sufficiently strong common component across providers and when model-specific deviations partly offset each other. If providers interpret the same calls in systematically different ways, or if their errors are correlated rather than idiosyncratic, the seven-model average need not be reliable. Thus, whether an ensemble LLM measure improves reliability is an empirical question rather than an automatic consequence of combining multiple models.
Because the same seven providers score every call, stable differences in provider scoring levels cancel out of relative comparisons, and \(G_c(7)\) therefore does not penalize them. To also assess reliability in \textit{levels}, we report the absolute dependability coefficient \citep{brennan2001},
\begin{equation}
\Phi_c(7)
=
\frac{\widehat{\sigma}^2_{\text{call},c}}
{\widehat{\sigma}^2_{\text{call},c}
+
\widehat{\sigma}^2_{\text{provider},c}/7
+
\widehat{\sigma}^2_{\varepsilon,c}/7},
\end{equation}
which treats the seven providers as draws from a broader population of possible models and retains the provider variance in the denominator. \(\Phi_c(7)\) measures the dependability of the ensemble's absolute score when providers are treated as draws from the specified universe of admissible models. It is the relevant reliability concept whenever scores are used in absolute terms---for example, when comparing magnitudes across studies that rely on different sets of models or when applying fixed thresholds to the scores.

Table~\ref{tab:reml_decomposition} reports the variance-decomposition results. Averaged across constructs, only 34.4\% of the variation in an LLM score is attributable to differences across calls. The remaining 65.6\% reflects model-related variation: 33.4\% is attributable to systematic provider differences and 32.2\% to residual model-by-call variation. Sentiment is again the exception: 74\% of the variance is attributable to the call component. In contrast, constructs such as integrity and the specificity ratio are dominated by model-related variation, with only 13\% and 4\% of the variance, respectively, attributable to the call component.

\begin{center}\textit{[Insert Table~\ref{tab:reml_decomposition} about here]}\end{center}

The relative reliability results provide a more encouraging perspective on the use of multiple LLMs. The generalizability coefficient for the seven-model average is 0.84 on average, and only one of the thirteen constructs---the specificity ratio---falls below 0.70. Sentiment has the highest reliability, with $G_c(7)=0.97$. These results indicate that, although individual LLM scores contain substantial model-related variation, averaging across providers recovers a reasonably reliable common signal for most constructs. In other words, the ensemble measure generally preserves relative differences across calls, even when individual providers differ in their score levels or in their evaluation of particular calls.

The absolute reliability results are markedly weaker. The dependability coefficient averages 0.72 across constructs; only nine of the thirteen constructs exceed 0.70, and only five exceed 0.80. The gap between \(G_c(7)\) and \(\Phi_c(7)\) is proportional to the provider share of the variance: sentiment is essentially unaffected (0.97 versus 0.95), while integrity falls from 0.76 to 0.51 and the specificity ratio from 0.49 to 0.22. Uncertainty is a telling intermediate case: the models rank calls fairly consistently (\(G_c(7)=0.90\)), but they disagree substantially about levels (\(\Phi_c(7)=0.65\)), reflecting the second-largest provider component among all constructs. Thus, averaging across providers delivers reliable relative rankings for most constructs, but reliable levels for fewer than half of them.

This distinction is important for interpretation. A relatively high \(G_c(7)\) does not imply that providers are interchangeable in levels, nor does it eliminate the evidence of model disagreement documented above. Rather, it shows that combining the seven models can mitigate some forms of model-specific noise and improve the reliability of relative rankings across calls. At the same time, the large provider component indicates that models differ substantially in the levels of scores they assign, while the residual component indicates that models also differ in how they evaluate particular calls. Thus, the variance decomposition suggests both a caution and a practical implication: relying on a single LLM can embed provider-specific measurement choices into empirical variables, but averaging across multiple providers can produce a more reliable measure for many constructs.

Overall, the variance-components analysis shows that LLM disagreement has two distinct implications. First, individual model scores are often substantially affected by provider-specific calibration and model-by-call variation, which cautions against treating any single LLM as an objective measure of the underlying construct. Second, the relatively high generalizability coefficient of the seven-model average indicates that, for most constructs, these disagreements do not fully eliminate the common signal shared across providers. Thus, the evidence points to a nuanced conclusion: LLM-based textual measures are often model-dependent at the individual-provider level, but ensemble measures can recover a more reliable common component for many constructs. The main exception is the specificity ratio, where averaging recovers neither a reliable ranking of calls (\(G_c(7)=0.49\)) nor reliable levels (\(\Phi_c(7)=0.22\)); for integrity, averaging restores a usable ranking (\(G_c(7)=0.76\)) but not reliable levels (\(\Phi_c(7)=0.51\)).

\subsection{Observable determinants of LLM disagreement}
The preceding analyses show that LLM disagreement varies substantially across constructs and is not fully captured by a single common model component. We next ask whether this disagreement is systematically related to observable firm and call characteristics. This analysis is descriptive rather than causal: our objective is to understand whether LLM disagreement is concentrated in particular types of firms, calls, or textual environments.

For each call and construct, we measure LLM disagreement as the standard deviation of the seven provider-standardized LLM scores. Since we analyze thirteen constructs, each call contributes thirteen construct-specific disagreement measures. We then pool these observations and regress LLM disagreement on a set of firm-level and call-level characteristics. The firm-level variables include size, book-to-market, Tobin's $Q$, and the earnings surprise associated with the call. Earnings surprise is included because disagreement may be higher when the underlying performance news is more salient, ambiguous, or mixed relative to market expectations. The textual variables include readability, measured using the Fog index; text length; financial jargon intensity; numerical intensity; and forward-looking language intensity. These variables allow us to examine whether disagreement is higher when calls are longer, more complex, more quantitative, or more forward-looking.

We also include the average pairwise cosine similarity of the LLM justifications for each call and construct. In principle, higher justification similarity should be associated with lower score dispersion if models that emphasize similar textual features also assign more similar scores. However, this relation is an empirical question, especially given the earlier evidence that similarity in scores and similarity in justifications do not always move together. All specifications include construct fixed effects, which absorb average differences in disagreement across prompts/constructs. In some specifications, we also include industry and quarter fixed effects to account for persistent differences across industries and time.

Table~\ref{tab:dispersion_std} reports the results. Firm size and Tobin's $Q$ are positively associated with LLM disagreement, suggesting that disagreement is greater for larger and higher-growth firms. One interpretation is that these firms tend to have more complex operations and disclosures, which may leave more room for model-specific interpretation. There is no significant relation between earnings surprise and the level of LLM disagreement.

\begin{center}\textit{[Insert Table~\ref{tab:dispersion_std} about here]}\end{center}

The textual characteristics provide additional insight. Readability, as measured by the Fog index, is not systematically related to LLM disagreement in the richer specifications. This suggests that disagreement is not primarily driven by conventional readability constraints: the models do not appear to disagree simply because the text is syntactically more difficult. By contrast, calls with a greater share of forward-looking language exhibit significantly higher disagreement. This pattern is intuitive because forward-looking statements often involve expectations, risks, and conditional claims, all of which may require more interpretation than descriptions of realized outcomes. Such language may therefore provide more scope for different LLMs to emphasize different parts of the same disclosure.

Finally, higher similarity in model justifications is associated with lower LLM disagreement. This result suggests that the justifications are informative about the scoring process: when models provide more similar rationales, they also tend to assign more similar scores. At the same time, the relation is not mechanical. Models may use similar language in their justifications while mapping the text into different numerical scores, or they may assign similar scores while emphasizing different aspects of the call. Thus, the negative association between justification similarity and score dispersion provides evidence that stated rationales are economically meaningful, while also reinforcing the importance of analyzing both scores and explanations.\footnote{Table~\ref{tab:dispersion_range} in the Online Appendix reports results using a range-based dispersion measure, defined as the maximum minus the minimum of the provider-standardized LLM scores for each call and construct. The results are similar to those reported above using the standard-deviation-based dispersion measure.}

Importantly, the explanatory power of these regressions is quite low. Even after controlling for firm characteristics, textual characteristics, justification similarity, construct, industry, and quarter fixed effects, the model explains only 20\% of the variation in LLM disagreement. This indicates that disagreement across LLMs is not simply a function of observable firm characteristics, standard textual measures, or broad industry and time patterns. Instead, much of the disagreement appears to arise from model-specific responses to particular texts. This conclusion is consistent with the variance-decomposition results: LLM disagreement reflects not only stable provider-level calibration differences, but also residual model-by-call variation that is difficult to explain using standard observables.

\subsection{Does LLM disagreement predict market and analyst disagreement?}

Finally, we ask whether disagreement among LLMs is informative about subsequent disagreement in markets. If cross-model dispersion partly reflects genuine ambiguity in the disclosure, then calls on which the models disagree should be followed by greater disagreement among investors and analysts. We regress post-call disagreement outcomes on call-level LLM dispersion, defined as the average across the thirteen measures of the standard deviation of the seven provider-standardized scores. We examine five outcomes: next-quarter analyst forecast dispersion, the analyst consensus forecast error (from the I/B/E/S detail file), and abnormal bid-ask spreads, trading volume, and return variance in the two-day announcement window. Each column of Table~\ref{tab:postcall_disagreement} is a call-level regression, controlling for pre-call analyst coverage and dispersion, size, book-to-market, Tobin's Q, and the earnings surprise, with industry and quarter fixed effects and standard errors clustered by firm.

As shown, LLM dispersion is not significantly related to any of the five outcomes. That is, calls on which the models disagree are not associated with greater dispersion or larger errors in analyst forecasts, nor with wider bid-ask spreads, heavier trading, or higher return variance. 

These results provide no evidence that LLM disagreement captures disclosure ambiguity that is subsequently reflected in analyst or market disagreement, and are consistent with a substantial model-specific component in LLM-based measurement.

\begin{center}\textit{[Insert Table~\ref{tab:postcall_disagreement} about here]}\end{center}

\subsection{Does model choice affect empirical conclusions?}

The preceding results establish that LLM-based measures are model-dependent. Specifically, the distributions and properties of constructs commonly used in empirical accounting and finance research vary substantially across LLMs, and LLM-generated measures exhibit low correlations with commonly used non-LLM measures of the same constructs. In this section, we examine whether---and to what extent---empirical conclusions would differ depending on the LLM used to construct earnings-call measures.

To do so, for each construct and each provider separately, we regress announcement returns on the provider's standardized LLM score, controlling for firm characteristics and including industry and quarter fixed effects. Each provider-specific coefficient in Table~\ref{tab:main_car01_industry} can therefore be interpreted as the finding that a researcher would obtain if they used that provider's LLM output as the textual measure.

Two results stand out. First, for several constructs—including climate risk, management clarity, specificity, and the specificity ratio—the provider-specific coefficients vary substantially in magnitude, sign, and statistical significance. For climate risk, for example, the coefficients in the first column range from $-0.4$ to $+0.2$, and only one of the seven model-specific measures, Qwen's, is significantly associated with the subsequent market reaction. A similar pattern emerges for management clarity, where statistical significance is driven by a single provider, and for the specificity ratio, where the coefficients are significant for Mistral and OpenAI but not for the other providers. These results illustrate a consequential form of model dependence: two researchers analyzing the same earnings calls using the same empirical specification, but choosing different LLM providers, could reach materially different conclusions, with point estimates differing in magnitude or even sign and statistically significant results under some providers but not others.

Second, some constructs display more consistent patterns across LLMs. For corporate culture, sentiment, and uncertainty, the signs and statistical significance of the LLM-based coefficients are more stable across providers. Even in these cases, however, the estimated magnitudes vary substantially, especially for sentiment and uncertainty. To assess formally whether these coefficient differences are statistically meaningful, we estimate, for each construct, the seven provider-specific return regressions jointly as a stacked regression. In this specification, the right-hand-side variables are interacted with provider indicators, making the stacked regression equivalent to estimating the provider-specific regressions separately while allowing us to test equality of coefficients across providers. We then test the equality of the seven LLM score coefficients using a Wald test, clustering standard errors by firm. Equality is rejected at the 1\% level for sentiment \((\chi^{2}=32.8)\) and uncertainty \((\chi^{2}=33.5)\). Corporate culture is the clearest case in which the results are relatively consistent across providers and the Wald test does not reject equality of the provider-specific coefficients.

Finally, the comparison with traditional non-LLM measures further illustrates the importance of measurement choice. For example, while the LLM-based sentiment coefficients are positive and statistically significant across providers, and the LLM-based uncertainty coefficients are negative and statistically significant across providers, the corresponding traditional measures are very different in magnitude and not statistically significant. This pattern shows that even when LLMs agree with one another, their measures can differ meaningfully from traditional text-based measures.

\begin{center}
\textit{[Insert Table~\ref{tab:main_car01_industry} about here]}\end{center}

\subsection{Is the disagreement an artifact of differing knowledge cutoffs?}
\label{sec:cutoff_null}

A natural concern is that the cross-model disagreement we document reflects the
models' differing training-data \emph{knowledge cutoffs} rather than genuine
differences in how they read the same text. Because a language model is trained on
data only up to some date, it may have ingested news coverage, analyst commentary,
and the realized market reaction surrounding an earnings call that predates its
cutoff. A model scoring such a call could then be guided by the realized outcome
rather than by the transcript alone---the look-ahead bias and outcome memorization
documented for LLM-based financial prediction
\citep{glasserman2023assessing,didisheim2025ai,kong2026evaluating}. If two models
with different cutoffs bring systematically different outside knowledge to the same
call, part of their disagreement would be an artifact of training vintage rather
than of interpretation. This concern is sharpest for sentiment:
\citet{glasserman2023assessing} show that an LLM's prior knowledge of the firms
involved materially shapes its sentiment scores, which makes sentiment both the
construct most exposed to training-corpus leakage and the natural place to look for
a cutoff effect.

Separating a cutoff effect from model identity is difficult because, across
providers, a model's cutoff is almost perfectly collinear with the model itself.
The cutoffs of our seven models fall on either side of the 2024 sample: four are
in 2023 (Anthropic in August, OpenAI and Mistral in October, and Meta's Llama in
December), one is in early 2025 (Google), and two---Qwen, at June~2024, and
DeepSeek, at July~2024---fall inside the window (Table~\ref{tab:cutoff_robustness}). The direction of potential
contamination is important to fix at the outset. For a given model, a call dated
\emph{before} its cutoff is one the model could have ingested together with the
call's market outcome, and is thus potentially contaminated; a call dated
\emph{after} the cutoff cannot have been seen and is clean. Under this logic the early-2025 model could in principle have memorized outcomes
for every call in our 2024 sample, the four 2023 models could have memorized
none, and two models straddle: DeepSeek, with 1,242 pre-cutoff (potentially
contaminated) and 704 post-cutoff (clean) calls, and Qwen, with 970 and 976. We
therefore treat DeepSeek and Qwen as within-model probes and the remaining five
models, whose contamination status does not change within 2024, as a calendar
control in a difference-in-differences design. Because both post-cutoff periods
fall in the second half of 2024, any raw pre-to-post change in their behavior
must be purged of the common calendar trend before it can be read as a cutoff
effect. The evidence points consistently away from a cutoff
explanation.

We begin by asking whether disagreement tracks differences in training vintage. If
it did, model pairs with more distant cutoffs should agree less. Across the 21 provider pairs, however, the rank correlation between the gap in
cutoff dates (in months) and pairwise agreement (the mean across constructs of
the two models' Spearman correlation) is essentially zero ($\rho=0.00$,
$p=0.99$), whereas a cutoff story predicts a negative relation. The most telling
version of this test uses the one pair of models that shares an identical cutoff
(OpenAI and Mistral, October~2023): its agreement ($\rho=0.62$) is no higher
than that of the same two models paired with Google, whose cutoff is fifteen
months later ($0.62$ on average). Nor is Google, the only model whose cutoff
postdates the entire sample and hence the one with potential access to every
realized 2024 outcome, an outlier: its average agreement with the other six
models ($0.56$) matches that of OpenAI ($0.56$) and Mistral ($0.56$). Models with
the same training vintage do not converge, and the model with the most potential
access to outcomes does not diverge, which is difficult to reconcile with the idea
that shared training vintage manufactures shared scores.

We next examine whether the straddling models' disagreement with the consensus
shifts as each crosses its own cutoff. DeepSeek's raw absolute deviation from
the leave-one-out seven-model consensus rises from $0.457$ before its cutoff to
$0.485$ after, an increase of about six percent, while Qwen's is unchanged
($0.849$ and $0.851$); but because both post periods fall late in 2024, a raw
change conflates the cutoff with any common calendar drift. A difference-in-differences that removes the common monthly
pattern (provider and month fixed effects, standard errors clustered by firm)
leaves a smaller DeepSeek-specific increase of $+0.020$ ($t=2.9$), against a
mean deviation of $0.46$, and no change for Qwen ($-0.005$, $t=-0.6$). Two features of DeepSeek's residual argue against a contamination reading.
First, its construct signature is the opposite of what outcome leakage would
produce. The increase is concentrated in the most subjective constructs---corporate
culture ($+0.075$), quality ($+0.067$), and management clarity ($+0.057$), each
significant---none of which has a realized-outcome referent that a model could
memorize, whereas the constructs where leakage is most plausible show nothing:
the interaction is an insignificant $-0.003$ ($t=-0.2$) for sentiment, is flat
for integrity and the specificity ratio, and is, if anything, negative for
political risk ($-0.027$, $t=-1.2$) (Table~\ref{tab:cutoff_did_construct}).
Qwen, whose overall deviation does not change, shows the same for sentiment
($-0.014$, $t=-1.0$), and its few significant construct-level changes run in
both directions. Memorization of realized outcomes should surface in the
news-grounded constructs, not in perceptions of ``respect'' or ``teamwork.'' Second, the effect is economically negligible---roughly four percent of the
average deviation---and, as we show next, is absent in the one construct
where outcome leakage can be tested directly.

That direct test concerns sentiment. If a straddling model's pre-cutoff sentiment scores were contaminated by the
realized outcome, their correlation with the announcement CAR should fall once
the model crosses its cutoff and can no longer have observed that outcome. It
does not, for either model: the correlation is $0.17$ before DeepSeek's cutoff
and $0.27$ after, an
increase that mirrors the common movement across the five control models over
the same months (from $0.17$ to $0.27$); Qwen's moves from $0.16$ to $0.18$
around its cutoff, against $0.20$ to $0.22$ for the controls. The cross-model comparison tells the same
story. The one always-pre-cutoff model, Google, exhibits a
sentiment--CAR correlation of $0.24$, against $0.20$ for the four always-post-cutoff models pooled and $0.20$--$0.23$
for them individually; access
to realized outcomes, where it exists, does not visibly sharpen the
sentiment--return link. A final check looks at
the dispersion of the scores themselves: if outside knowledge had anchored the straddling models' pre-cutoff scores, they
should become noticeably noisier once that knowledge is unavailable. They do
not---the post-to-pre within-firm variance ratio is $1.05$ for DeepSeek and
$0.94$ for Qwen, against $1.01$ and $1.02$ for the five control models.

Taken together, the construct most exposed to training-corpus leakage---sentiment---
is among the least affected by cutoff crossing, and the one direct look-ahead test
we can run shows no sign of the predicted decline. Therefore, we find little evidence that differences in knowledge cutoffs account for the cross-model disagreement we document. %

\begin{center}\textit{[Insert Table~\ref{tab:cutoff_robustness} about here]}\end{center}

\begin{center}\textit{[Insert Table~\ref{tab:cutoff_did_construct} about here]}\end{center}

\section{Do Newer Models Converge?}
\label{sec:frontier}

The seven models we study were, at the time of our analysis, among those most widely used in academic applications and were comparable in capability to the models employed in the studies reviewed in Section~\ref{sec:literature}. Nevertheless, the rapidly evolving model landscape limits the generalizability of any such snapshot. The pace of new releases makes continuous updating infeasible: pipelines built around particular models can quickly become outdated, while re-scoring a large corpus after every release is impractical. Moreover, applying frontier models to samples of the size common in this literature can be prohibitively costly, particularly when researchers seek to conduct the cross-provider validation that our findings recommend.

To assess whether our conclusions extend to more recent models, we conduct a supplementary analysis using a random subsample of 250 calls covering 204 firms. We re-score these calls using the September 2026 successors from two of the seven model families in our main analysis, selected from the providers most widely used in academic research: OpenAI (\texttt{gpt-5.6-terra}) and Google (\texttt{gemini-3.8-flash}). We hold the transcripts, prompts, scoring scales, and inference settings fixed relative to the main analysis. We examine three measures commonly used in the accounting disclosure literature—sentiment \citep{Loughran_McDonald_2011}, management clarity \citep{barth2020econlinguistics}, and answer specificity \citep{hope2016benefits}. These measures span the range of cross-model agreement observed in our main analysis: agreement is highest for sentiment, lowest for specificity ratio, and intermediate for management clarity. This design therefore allows us to assess whether the pattern persists across the upper, middle, and lower portions of the agreement distribution.

For \emph{sentiment}---already the most reliable construct---little changes. The
OpenAI and Google models agreed strongly in our baseline setup (Spearman
correlation of $0.91$) and agree marginally more once upgraded ($0.92$). Their
correlation with the Loughran--McDonald dictionary is essentially flat, and low,
in both vintages (OpenAI $0.36 \rightarrow 0.33$; Gemini $0.32 \rightarrow 0.32$).
The dictionary, however, is one operationalization of tone, not ground truth, so
this stability is unremarkable: it says only that LLM-based and dictionary-based
sentiment remain distinct measures at the frontier, not that either is closer to
correct.

For \emph{management clarity}---the middle case---cross-provider agreement rises
modestly, from $0.68$ to $0.73$. Yet neither model moves toward the traditional
evasive-language benchmark: the correlation stays near $0.10$ and is, if anything,
slightly lower for the newer models (OpenAI $0.08 \rightarrow 0.06$; Gemini
$0.14 \rightarrow 0.11$). As with sentiment, the benchmark is an alternative
operationalization rather than a criterion of truth, so its persistent low
correlation cannot be read as an LLM failure; but it does show that greater
agreement between the frontier models does not reconcile them with the established
measure.

For the \emph{specificity ratio}---the least reliable construct in the main
analysis---the frontier delivers the largest change on both margins.
Cross-provider agreement rises sharply, from $0.37$ to $0.61$, and, unlike the
other two measures, each model also moves \emph{closer} to the traditional
benchmark (OpenAI $0.52 \rightarrow 0.57$; Gemini $0.44 \rightarrow 0.65$). This
measure is the one case in which the traditional benchmark can be treated as an
approximate ground truth: the specificity ratio is defined by an explicit counting
rule, and the non-LLM version simply implements that rule directly, whereas the
dictionary tone and evasive-language measures are only proxies for the constructs
they name. Rising alignment with the counting rule therefore indicates that the
newer models compute the ratio more \emph{correctly}, not merely more alike. The
mechanics confirm this: the two 2026 models count specific items far more
consistently (item-count correlation of $0.75$, versus $0.43$ in the original case). The newer models also count transcript length much more accurately (within roughly half to one times the true count, versus one‑tenth to three times before), correcting the faulty denominator that drove the divergence. Even so, cross-provider agreement reaches only $0.61$.

Taken together, upgrading to frontier models moves cross-provider agreement in the
right direction but does not eliminate the disagreement. Whether ``the right
direction''---higher agreement---also means greater accuracy is, in general,
outside the scope of our design, which measures whether models agree rather than
whether they are correct. The specificity ratio is a partial exception: because it
is defined by an explicit counting rule that the traditional benchmark implements
directly, higher agreement on this measure can plausibly be read as movement toward
the true value. Even there, however, substantial disagreement remains---the two
frontier models correlate only $0.61$---and, with the sole exception of sentiment,
which already exhibited high agreement, cross-provider disagreement persists at the
frontier. This is a demanding setting in which to find such disagreement. As noted
earlier, we deliberately study measurement rather than prediction---a far more
challenging task that would invite still greater cross-model disagreement and
additional problems such as look-ahead bias---and we do so using constructs that
are conceptually well defined. That substantial model dependence survives even
under these favorable conditions suggests that researchers should be cautious in
using a single model for measurement: another model, including a newer one, may
yield materially different results.

\section{Conclusion}\label{sec:conc}

We ask a simple question with broad consequences for empirical research in
accounting and finance: when a generative LLM converts text into a number, is
that number a property of the text or of the model? Using the question-and-answer
portions of 1,946 S\&P 500 earnings calls in 2024, we have seven LLMs from seven
independent developers score each call on thirteen measures that the literature
routinely extracts from corporate disclosure. The design is deliberately
conservative—every construct is contained in a fixed transcript, and no model is
asked to forecast a future metric—so that any disagreement can be read as noise in the
measurement instrument rather than informative disagreement about an unknown
future. We nonetheless find substantial divergence. The average pairwise
correlation across models is 0.52, only about a third of the variation in a
typical score is attributable to the text rather than the model, and for several
constructs the model matters far more than the firm. Disagreement is not confined
to soft constructs: it is among the largest for an explicitly specified implementation task (\textit{Specificity ratio}).

These results carry a direct lesson for empirical research. Accounting and finance have
increasingly treated LLM output as measurement-ready data, often substituting a
single model's scores for the dictionary- and rule-based measures that preceded
them. Our evidence shows that this substitution is not innocuous: LLM measures
are neither interchangeable with one another nor with the traditional measures
they replace—the ensemble sentiment score correlates only 0.29 with dictionary
tone—so a construct's empirical content now depends on the measurement technology
chosen. Most consequentially, the choice of model is a researcher degree of
freedom that can change conclusions. Estimating the same announcement-return
regression on the same calls, three researchers who differ only in their choice
of provider could report a positive effect, a negative effect, and no effect for
the very same construct. Because this choice is currently neither disclosed nor
justified in most applications, two studies that claim to measure the "same"
construct may be measuring materially different things.

These findings do not imply that LLM-based measurement should be abandoned. Averaging across providers behaves as
generalizability theory predicts: it cancels model-specific noise and recovers a
reliable ranking of calls for most measures, even though the individual models
disagree. But aggregation is a partial remedy, not a panacea. It restores
reliable relative rankings without restoring reliable levels. 

We also provide direct, if limited, evidence on whether newer models help. In a
smaller exercise, we re-score a random subsample of calls with the September 2026
successors of two providers, OpenAI and Google. Upgrading to these
frontier models moves cross-provider agreement in the right direction but does not
eliminate it: agreement rises most for the constructs on which the earlier models
diverged most, yet meaningful disagreement remains for every measure except
sentiment, which was already highly reliable. Greater agreement between the newer
models also does not, in general, bring them closer to the established benchmarks;
the exception is the specificity ratio---a rule-based measure for which higher
agreement plausibly reflects movement toward the true value, and even there the two
frontier models correlate only $0.61$. Continued model
improvement therefore does not by itself eliminate the need to assess measurement
invariance empirically.

Two features of our design bound the interpretation of these results. We query
each model through its API with a fixed prompt and without external tools,
retrieval, or extended-reasoning scaffolding. This isolates the model's own
contribution—its weights and alignment—and matches the pipelines through which
research measures are actually constructed, which for reasons of cost,
reproducibility, and structured output rely on API calls rather than on the
consumer chat applications. Our estimates therefore need not carry over to
measurement performed through those full products or through agentic systems that
add system prompts, tool use, web retrieval, and multi-step reasoning. Such
scaffolding could raise agreement, by grounding models in common external sources,
or lower it, by multiplying the idiosyncratic paths a model can follow, and each
such choice is itself a further undocumented researcher degree of freedom layered
on top of the model choice we study. 

We close with three practical recommendations for the research community. First,
model choice should be reported as a first-order design decision: studies should
disclose---and ideally justify---the specific model and version used, just as they
disclose a sample period or a variable definition, particularly since the model
snapshots underlying a measure are versioned and periodically deprecated. Second,
cross-provider validation should become routine; a second model is a cheap
diagnostic, and material sensitivity to it is itself a finding worth reporting.
Third, where feasible, researchers should prefer ensemble measures across
providers to any single model's output. More broadly, the field would
benefit from shared benchmarks and validation standards for LLM-based measurement,
from routine reporting of the sensitivity of results to model, version, and
prompt, and from external validation against human-coded or otherwise anchored
ground truth. As LLMs move from measurement toward more autonomous, agentic roles
in the research pipeline, the model-induced variation we document at the
measurement stage is likely to compound rather than cancel—making a discipline of
measurement reliability a prerequisite, not an afterthought, for credible research.

\clearpage

\bibliographystyle{apa}

\bibliography{Casting_bib.bib}

\clearpage

\section*{Exhibits and Tables}

\clearpage

\inputregtable{Assets/tables/summary_stats_main}

\clearpage

\begin{table}[htbp]
\centering
\small
\begin{threeparttable}
\caption{Inter-Provider Agreement by Construct: Distribution of Pairwise Spearman Correlations}
\label{tab:corr_summary}
\begin{tabular}{lccccc}
\toprule
Construct & $N$ & Min & Mean & Median & Max \\
\midrule
\textbf{Sentiment} & \textbf{1,946} & \textbf{0.77} & \textbf{0.85} & \textbf{0.86} & \textbf{0.90} \\
Innovation & 1,946 & 0.59 & 0.76 & 0.75 & 0.87 \\
Uncertainty & 1,946 & 0.54 & 0.64 & 0.62 & 0.75 \\
Climate Risk & 1,946 & 0.36 & 0.58 & 0.53 & 0.79 \\
Specificity & 1,946 & 0.34 & 0.54 & 0.51 & 0.70 \\
Management Clarity & 1,946 & 0.38 & 0.53 & 0.51 & 0.66 \\
Political Risk & 1,946 & 0.30 & 0.52 & 0.53 & 0.64 \\
Corporate Culture & 1,938 & 0.33 & 0.49 & 0.48 & 0.66 \\
Teamwork & 1,939 & 0.37 & 0.49 & 0.49 & 0.62 \\
Quality & 1,942 & 0.27 & 0.44 & 0.41 & 0.62 \\
Respect & 1,939 & 0.27 & 0.41 & 0.34 & 0.58 \\
Integrity & 1,942 & 0.14 & 0.32 & 0.29 & 0.54 \\
\textit{Specificity Ratio} & \textit{1,878} & \textit{0.10} & \textit{0.23} & \textit{0.23} & \textit{0.43} \\
\midrule
Mean & & 0.37 & 0.52 & 0.50 & 0.67 \\
Median & & 0.34 & 0.52 & 0.51 & 0.66 \\
\bottomrule
\end{tabular}
\begin{tablenotes}[flushleft]\footnotesize
\item \textit{Notes:} For each construct we compute the Spearman rank correlation between every pair of the LLM providers scoring it (the non-LLM benchmark is excluded) across all transcripts, then summarize the resulting pairwise correlations. $N$ is the number of transcripts for which all providers have a score. Constructs are ordered by mean pairwise correlation; the most (least) consistent construct is shown in \textbf{bold} (\textit{italics}).
\end{tablenotes}
\end{threeparttable}
\end{table}

\clearpage

\begingroup\catcode95=12\begin{table}[htbp]
\centering
\small
\begin{threeparttable}
\caption{Cross-Model Commonality: PCA Summary}
\label{tab:pca_summary}
\begin{tabular}{lcc}
\toprule
Construct & PC1 Variance Explained (\%) & \# PCs for 90\% \\
\midrule
\textbf{Sentiment} & \textbf{84.95} & \textbf{3} \\
Climate Risk & 75.71 & 3 \\
Innovation & 74.28 & 4 \\
Uncertainty & 65.28 & 5 \\
Political Risk & 64.24 & 4 \\
Specificity & 58.89 & 5 \\
Corporate Culture & 57.05 & 5 \\
Teamwork & 56.85 & 6 \\
Management Clarity & 55.45 & 5 \\
Quality & 52.67 & 6 \\
Respect & 51.87 & 6 \\
Integrity & 45.99 & 6 \\
\textit{Specificity Ratio} & \textit{32.49} & \textit{6} \\
\midrule
Mean & 59.67 & 4.9 \\
Median & 57.05 & 5.0 \\
\bottomrule
\end{tabular}
\begin{tablenotes}[flushleft]\footnotesize
\item \textit{Notes:} PCA on z-standardized scores from the LLMs scoring each construct. PC1 variance = cross-model agreement; higher is stronger convergent validity.
\end{tablenotes}
\end{threeparttable}
\end{table}
\endgroup

\clearpage

\begingroup\catcode95=12\begin{table}[htbp]
\centering
\small
\begin{threeparttable}
\caption{Variance Decomposition of LLM Scores: Provider Calibration vs.\ Transcript Ambiguity}
\label{tab:reml_decomposition}
\begin{tabular}{lcccccc}
\toprule
Construct & Transcript & Provider & Residual & $G$ & $\Phi$ & $N$ \\
 & (\% of Total) & (\% of Total) & (\% of Total) & & & \\
\midrule
Sentiment & 74.0\% & 9.9\% & 16.1\% & 0.970 & 0.952 & 13,622 \\
Innovation & 52.0\% & 24.6\% & 23.4\% & 0.940 & 0.884 & 13,622 \\
Climate Risk & 58.5\% & 10.1\% & 31.4\% & 0.929 & 0.908 & 13,622 \\
Uncertainty & 20.9\% & 62.3\% & 16.9\% & 0.896 & 0.648 & 13,622 \\
Specificity & 42.4\% & 15.0\% & 42.6\% & 0.875 & 0.838 & 13,622 \\
Corporate Culture & 30.2\% & 36.3\% & 33.5\% & 0.863 & 0.752 & 13,566 \\
Teamwork & 30.1\% & 36.2\% & 33.7\% & 0.862 & 0.751 & 13,573 \\
Political Risk & 47.9\% & 4.9\% & 47.2\% & 0.835 & 0.821 & 9,730 \\
Quality & 28.4\% & 32.3\% & 39.3\% & 0.835 & 0.735 & 13,594 \\
Respect & 20.6\% & 45.9\% & 33.4\% & 0.812 & 0.645 & 13,573 \\
Management Clarity & 25.5\% & 29.1\% & 45.4\% & 0.797 & 0.706 & 13,622 \\
Integrity & 12.8\% & 58.8\% & 28.4\% & 0.760 & 0.508 & 13,594 \\
Specificity Ratio & 3.9\% & 68.3\% & 27.9\% & 0.492 & 0.219 & 13,146 \\
\midrule
\textit{Average} & 34.4\% & 33.4\% & 32.2\% & 0.836 & 0.720 & 13,269 \\
\bottomrule
\end{tabular}
\begin{tablenotes}[flushleft]\footnotesize
\item \textit{Notes:} Variance components of the crossed call $\times$ provider random-effects model, estimated per construct from the two-way analysis-of-variance mean squares on the calls scored by every provider ($\hat\sigma^2\sb{r}=MS\sb{r}$, $\hat\sigma^2\sb{t}=(MS\sb{t}-MS\sb{r})/k$, $\hat\sigma^2\sb{p}=(MS\sb{p}-MS\sb{r})/n\sb{t}$; negative estimates set to zero); for this balanced design these coincide with the REML estimates. $G=\sigma^2\sb{t}/(\sigma^2\sb{t}+\sigma^2\sb{r}/k)$ is the relative generalizability coefficient and $\Phi=\sigma^2\sb{t}/(\sigma^2\sb{t}+\sigma^2\sb{p}/k+\sigma^2\sb{r}/k)$ the absolute dependability coefficient of the $k$-model average, with $k$ the number of providers scoring the construct. Note that $k$ is equal to five for political risk.  $\Phi$ additionally penalizes stable provider-level calibration differences, so it measures reliability of score levels rather than of rankings. $N$ is calls $\times$ providers.
\end{tablenotes}
\end{threeparttable}
\end{table}
\endgroup

\clearpage

\inputregtable{Assets/tables/regression_dispersion_std}

\clearpage

\begin{table}
\centering
\small
\caption{LLM Disagreement and Post-Call Market \& Analyst Disagreement}
\label{tab:postcall_disagreement}
\begin{adjustbox}{max width=.92\textwidth}
\begin{threeparttable}
\begin{tabular}{lccccc}
\toprule
 & Analyst Disp. & Forec.\ Error & Bid--Ask & Abn.\ Volume & Abn.\ Ret.\ Var. \\
 & (1) & (2) & (3) & (4) & (5) \\
\midrule
LLM Dispersion (Z-score) & -0.0019 & -0.0041 & -0.0000 & 0.0099 & 0.4880 \\
 & (-1.23) & (-0.63) & (-1.03) & (0.12) & (0.69) \\
\multicolumn{6}{l}{\textit{Controls}} \\
Pre-Call Analyst Count & 0.0003 & 0.0005 & 0.0000$^{**}$ & 0.0621$^{***}$ & 0.2044 \\
 & (0.78) & (0.35) & (2.33) & (3.36) & (1.43) \\
Pre-Call Analyst Disp. & 0.5763$^{***}$ & 0.7262$^{***}$ & 0.0002$^{*}$ & -0.4429 & -15.8447$^{***}$ \\
 & (9.63) & (5.70) & (1.90) & (-0.51) & (-3.18) \\
$\ln(\text{Assets})$ & -0.0051$^{**}$ & 0.0084 & -0.0000$^{***}$ & -0.4003$^{***}$ & -2.0895$^{***}$ \\
 & (-2.05) & (0.98) & (-4.48) & (-4.35) & (-3.04) \\
Book-to-Market & 0.0437$^{**}$ & 0.1955$^{***}$ & -0.0001$^{**}$ & 0.1491 & -2.8404 \\
 & (2.56) & (2.70) & (-2.55) & (0.41) & (-1.43) \\
Tobin's $Q$ & -0.0012$^{*}$ & -0.0060$^{**}$ & 0.0000 & -0.0817$^{*}$ & -0.5260$^{**}$ \\
 & (-1.65) & (-1.99) & (0.07) & (-1.96) & (-2.24) \\
SUE & -0.0005 & 0.0018 & 0.0000 & -0.0492 & -0.1616 \\
 & (-0.67) & (0.97) & (0.68) & (-1.47) & (-0.49) \\
\midrule
Observations & 1,889 & 1,826 & 1,890 & 1,890 & 1,890 \\
Adj.\ $R^2$ & 0.467 & 0.197 & 0.055 & 0.069 & 0.045 \\
Industry FE, Quarter FE & Yes & Yes & Yes & Yes & Yes \\
\bottomrule
\end{tabular}
\begin{tablenotes}[flushleft]\footnotesize
\item \textit{Notes:} Each column is a separate OLS of a post-call disagreement outcome at the call level (one observation per call) on the call-level LLM dispersion (the average across the thirteen constructs of the standard deviation of the seven provider-standardized scores, standardized). Outcomes: log analyst forecast dispersion; forecast error ($|$post-call consensus $-$ actual$|$/price $\times 100$, from the I/B/E/S detail file); abnormal quoted bid--ask spread; abnormal volume; abnormal return variance (each summed over the $(0,1)$ window in excess of its estimation-window mean). Controls: pre-call analyst count and dispersion, ln(Assets), book-to-market, Tobin's Q, SUE. The dependent variables, SUE and the analyst variables are winsorized at 1/99\%. $t$-statistics in parentheses, SE clustered by firm. $^{*}\,p<0.10$; $^{**}\,p<0.05$; $^{***}\,p<0.01$.
\end{tablenotes}
\end{threeparttable}
\end{adjustbox}
\end{table}

\clearpage

\inputregtable{Assets/tables/main_table_CAR01_Carhart_industry}

\clearpage

\inputregtable{Assets/tables/cutoff_robustness}

\clearpage

\begin{table}[htbp]
\centering
\small
\begin{threeparttable}
\caption{Knowledge-Cutoff Robustness: Per-Construct Difference-in-Differences on Disagreement}
\label{tab:cutoff_did_construct}
\begin{tabular*}{\textwidth}{@{\extracolsep{\fill}}lcc}
\toprule
Construct & DeepSeek $\times$ post & Qwen $\times$ post\\
\midrule
Corp.\ culture & 0.0748$^{***}$ & -0.0402\\
 & (3.35) & (-1.54)\\
Quality & 0.0668$^{***}$ & -0.0220\\
 & (2.88) & (-0.82)\\
Mgmt clarity & 0.0573$^{***}$ & 0.0847$^{***}$\\
 & (2.65) & (3.36)\\
Teamwork & 0.0536$^{**}$ & -0.0411$^{*}$\\
 & (2.33) & (-1.65)\\
Respect & 0.0408$^{*}$ & -0.0402\\
 & (1.72) & (-1.60)\\
Innovation & 0.0335$^{**}$ & -0.0059\\
 & (2.33) & (-0.29)\\
Specificity & 0.0196 & -0.0098\\
 & (1.00) & (-0.38)\\
Integrity & 0.0184 & -0.0979$^{***}$\\
 & (0.75) & (-3.56)\\
Climate risk & -0.0013 & -0.0633$^{***}$\\
 & (-0.09) & (-2.74)\\
\textbf{Sentiment} & \textbf{-0.0030} & \textbf{-0.0138}\\
 & \textbf{(-0.25)} & \textbf{(-0.97)}\\
Uncertainty & -0.0202 & -0.0253\\
 & (-1.06) & (-1.28)\\
Specificity ratio & -0.0205 & 0.0026\\
 & (-0.70) & (0.09)\\
Political risk & -0.0266 & ---\\
 & (-1.24) & \\
\bottomrule
\end{tabular*}
\begin{tablenotes}[flushleft]\footnotesize
\item \textit{Notes:} Each construct is a separate OLS $|\text{dev}|_{ij}=\sum_{s}\beta_s\,(\text{straddler}_{s,i}\times\text{post}_{s,t})+\text{provider FE}+\text{month FE}+\varepsilon$, where $|\text{dev}|$ is the absolute deviation from the leave-one-out 7-model consensus on $z$-scores taken within provider$\times$construct (so provider calibration is netted out) and ``post'' $=$ call after the straddler's own cutoff month (DeepSeek: from August 2024; Qwen: from July 2024). Each $\beta_s$ is the extra change in that model's disagreement at its own cutoff beyond the common calendar trend; because the two cutoffs differ by a month, July 2024 calls are post-cutoff for one straddler and pre-cutoff for the other. Cells marked --- are provider--construct series excluded from the sample (Section~\ref{sec:llm_measures}). A contamination/look-ahead story predicts the news-grounded \emph{sentiment} construct should move most; instead it is among the \emph{least} affected for both straddlers. $t$-statistics in parentheses, SE clustered by firm. $^{*}\,p<0.10$; $^{**}\,p<0.05$; $^{***}\,p<0.01$.
\end{tablenotes}
\end{threeparttable}
\end{table}

\appendix

\begin{appendices}

\clearpage
\section*{Appendix Table}

\setcounter{table}{0}
\renewcommand{\thetable}{A\arabic{table}}
\singlespacing
\begin{longtable}{>{\arraybackslash}m{.30\textwidth} >{\arraybackslash}m{.67\textwidth}}
\caption{\textbf{Definition of variables}}\label{tab:varlist} \\
\def\sym#1{\ifmmode^{#1}\else\(^{#1}\)\fi}
\centering
&\\ \hline\hline
\textbf{Variable} & \textbf{Definition}  \\
\hline\hline
\multicolumn{2}{l}{\textbf{LLM-based variables}} \\ \hline
LLM Score & the score assigned by an LLM provider to the Q\&A text of a call for a given construct, on the scale requested in the prompt (0--1 for all constructs except sentiment, which is scored from $-1$ to $+1$; the specificity ratio is defined as specific words over total words). The full prompts are reported in the Appendix. Scores are winsorized at the 1st and 99th percentiles within construct and provider, and z-standardized within construct and provider where indicated \\ \hline
LLM Confidence & the self-reported confidence (0--1) that each provider returns together with its score \\ \hline
LLM Dispersion (Z) & the standard deviation of the seven providers' scores for the same call and construct, computed after z-standardizing each provider's scores within construct \\ \hline
Justif.\ Cosine Sim & the average pairwise cosine similarity of sentence embeddings of the seven providers' written justifications for the same call and construct \\ \hline
Justif.\ Jaccard Sim & the average pairwise word-overlap (Jaccard) similarity of the seven providers' written justifications for the same call and construct \\ \hline
\multicolumn{2}{l}{\textbf{Firm-level variables}} \\ \hline
$ln(Assets)$ & the natural logarithm of total assets (from Compustat) \\ \hline
$BTM$ & book-to-market ratio; total common/ordinary equity divided by the market value of equity (from Compustat) \\ \hline
$Q$ & Tobin's Q; the book value of assets minus book value of common equity plus the market value of common equity, divided by the total book value of assets (from Compustat) \\ \hline
$SUE$ & standardized unexpected earnings, from the I/B/E/S surprise history file, matched to the call's earnings announcement date \\ \hline
$CAR(0,1)$ & the cumulative abnormal return over trading days $(0,+1)$ around the call, from a Carhart four-factor model estimated on CRSP daily returns \\ \hline
Analyst Disp.\ (Q1) & the natural logarithm of one plus the coefficient of variation (standard deviation over the absolute mean) of I/B/E/S analyst EPS forecasts for the next fiscal quarter, from the first consensus published within 60 days after the call \\ \hline
Analyst Count (Q1) & the number of analyst estimates in that first post-call consensus \\ \hline
Pre-Call Analyst Disp.\ / Count & the same dispersion and count measures, from the last consensus published within 60 days before the call \\ \hline
{Forecast Error} & {the absolute difference between the post-call consensus (the mean of analysts' latest I/B/E/S detail-file forecasts issued within 30 days after the call, requiring at least three analysts) and the realized EPS for that quarter, scaled by the pre-call share price and multiplied by 100} \\ \hline
Abn.\ Bid-Ask / Volume / Ret.\ Var. & abnormal bid-ask spread, trading volume, and return variance over the $(0,+1)$ announcement window, each relative to the firm's estimation-window baseline (from CRSP) \\ \hline
\multicolumn{2}{l}{\textbf{Textual variables (management answers in the Q\&A)}} \\ \hline
Fog Index & the Gunning fog readability index of the answers \\ \hline
$ln(Word\ Count)$ & the natural logarithm of the total number of words in the answers \\ \hline
Financial Jargon & the ratio of finance-jargon terms to total words \\ \hline
Numbers & the ratio of numeric tokens to total words \\ \hline
Forward-Looking & the ratio of forward-looking terms to total words; forward-looking terms from \cite{Matsumoto_etal_AccRev2011} and \cite{bozanic2018management} \\ \hline
Specificity & the ratio of specific tokens---named entities, numeric values, monetary amounts, dates, and durations (Stanza named-entity recognition plus rule-based matching)---to total words; this is the non-LLM benchmark of the LLM specificity ratio \\ \hline
\hline
\end{longtable}
\doublespacing

\end{appendices}

\clearpage
\setcounter{page}{1}

\vfill
\begin{singlespace}
\begin{center}
\vspace*{2em}
{\Huge\bfseries Online Appendix\par}
\bigskip
\rule{\linewidth}{0.4pt}
\bigskip\bigskip
{\Large\bfseries \papertitle\par}
\ifANONYMOUS\else
\bigskip
{\large Hamid Boustanifar \qquad Sasan Mansouri\par}
\fi
\end{center}
\end{singlespace}
\thispagestyle{empty}

\bigskip

\section*{}

\setcounter{figure}{0}
\setcounter{table}{0}

\renewcommand{\thesection}{\Alph{section}}
\setcounter{section}{0}

\counterwithin{table}{section}
\counterwithin{figure}{section}
\counterwithin{equation}{section}

\renewcommand{\thetable}{OA.\Alph{section}.\arabic{table}}
\renewcommand{\thefigure}{OA.\Alph{section}.\arabic{figure}}
\renewcommand{\theequation}{OA.\Alph{section}.\arabic{equation}}

\clearpage

\clearpage
\appendix

\section{Correlation Heatmaps}
\label{app:heatmaps}

\begin{figure}[H]
  \centering
  \begin{subfigure}[b]{0.48\textwidth}
    \includegraphics[width=\textwidth]{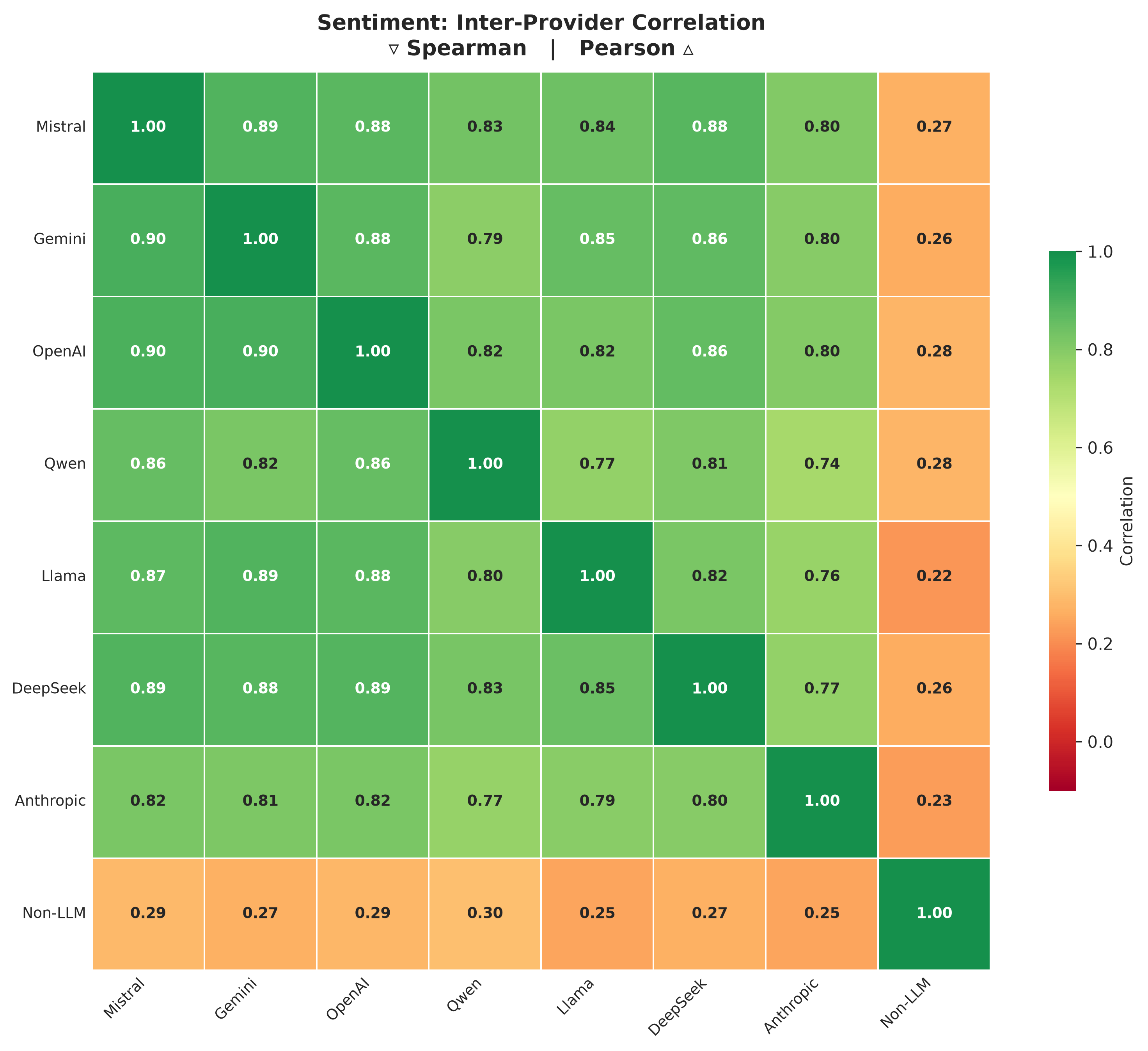}
    \caption{Score correlations}
  \end{subfigure}\hfill
  \begin{subfigure}[b]{0.48\textwidth}
    \includegraphics[width=\textwidth]{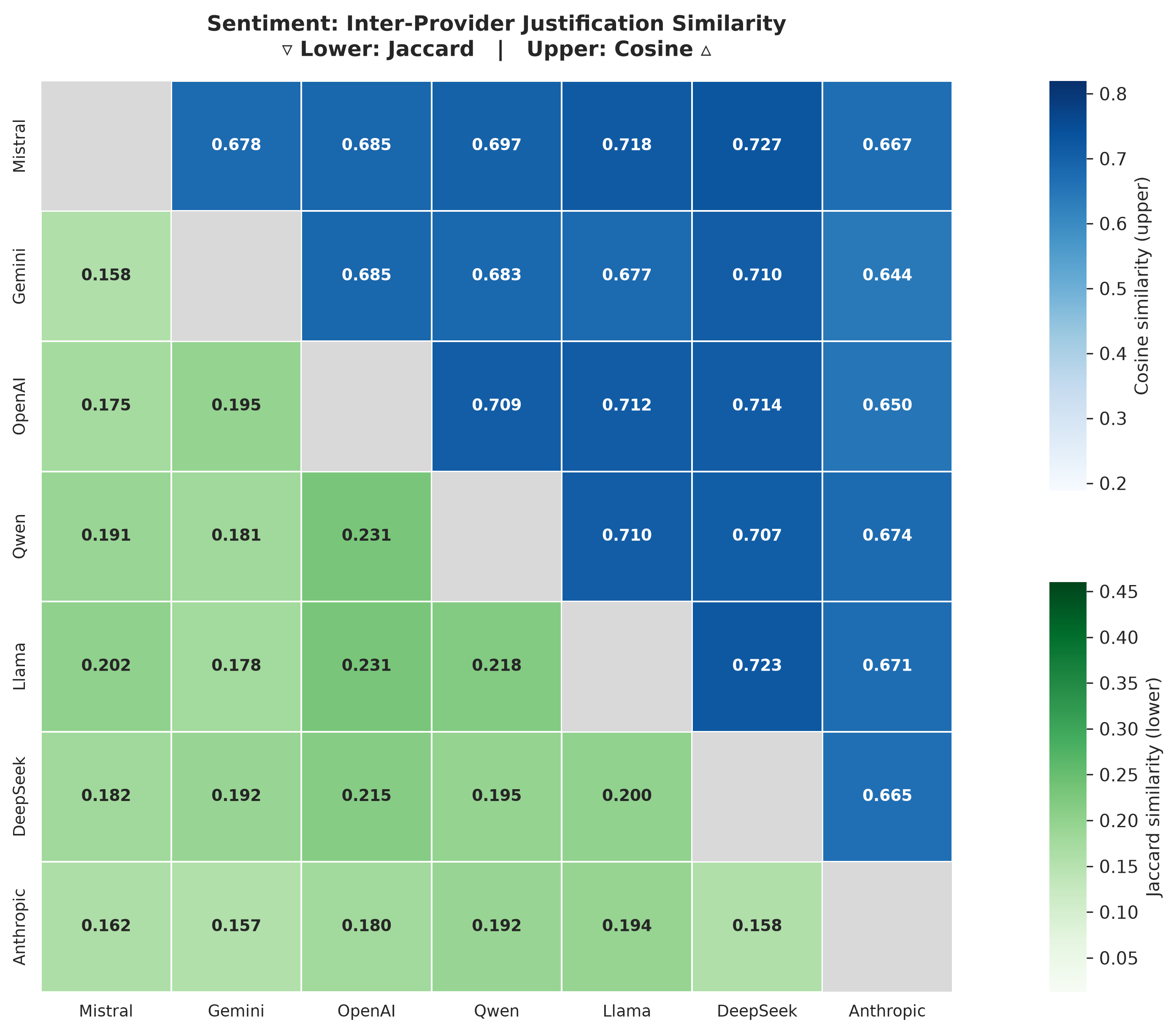}
    \caption{Textual similarity of the score justifications}
  \end{subfigure}
  \caption{Sentiment}
\end{figure}

\begin{figure}[H]
  \centering
  \begin{subfigure}[b]{0.48\textwidth}
    \includegraphics[width=\textwidth]{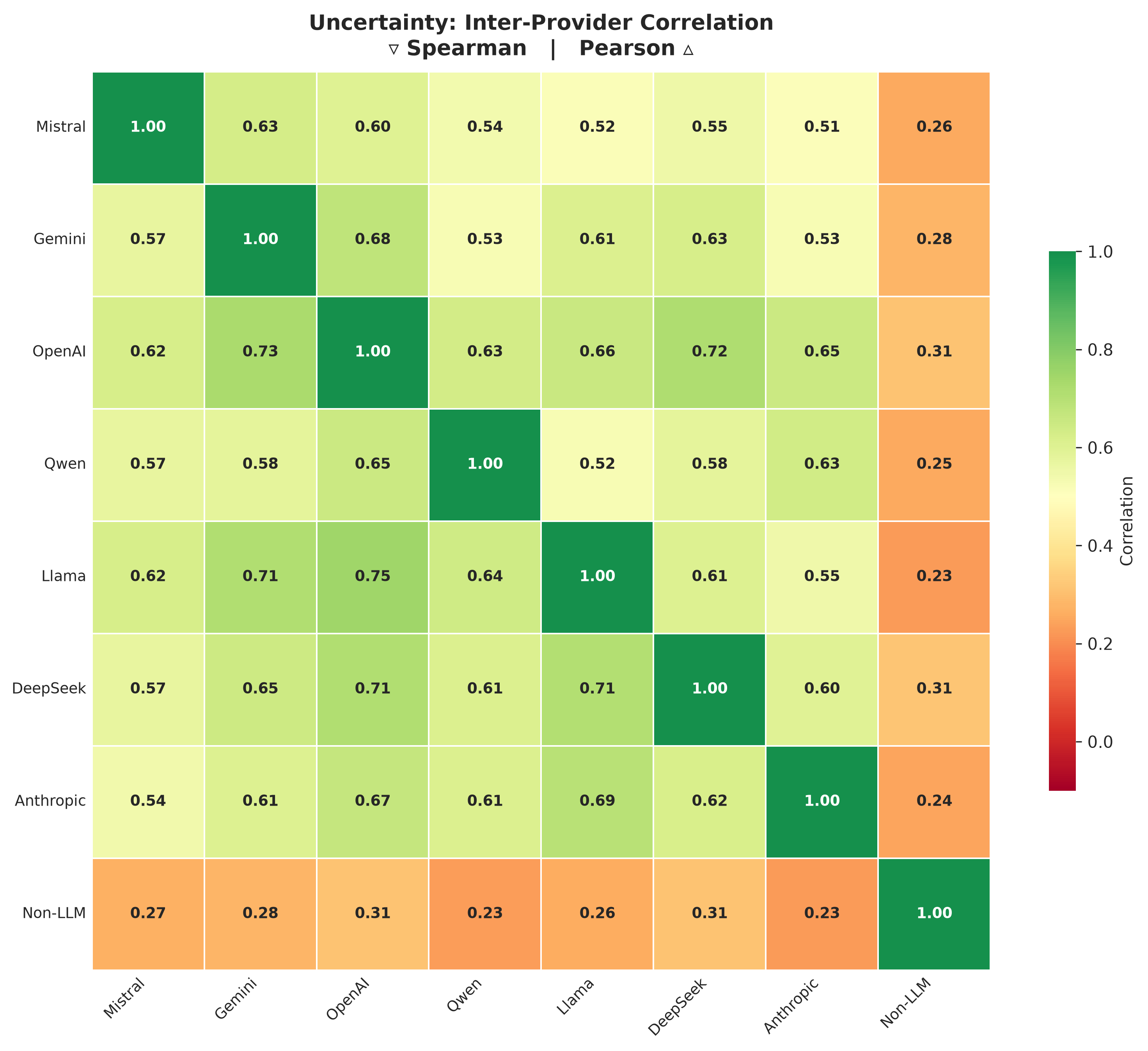}
    \caption{Score correlations}
  \end{subfigure}\hfill
  \begin{subfigure}[b]{0.48\textwidth}
    \includegraphics[width=\textwidth]{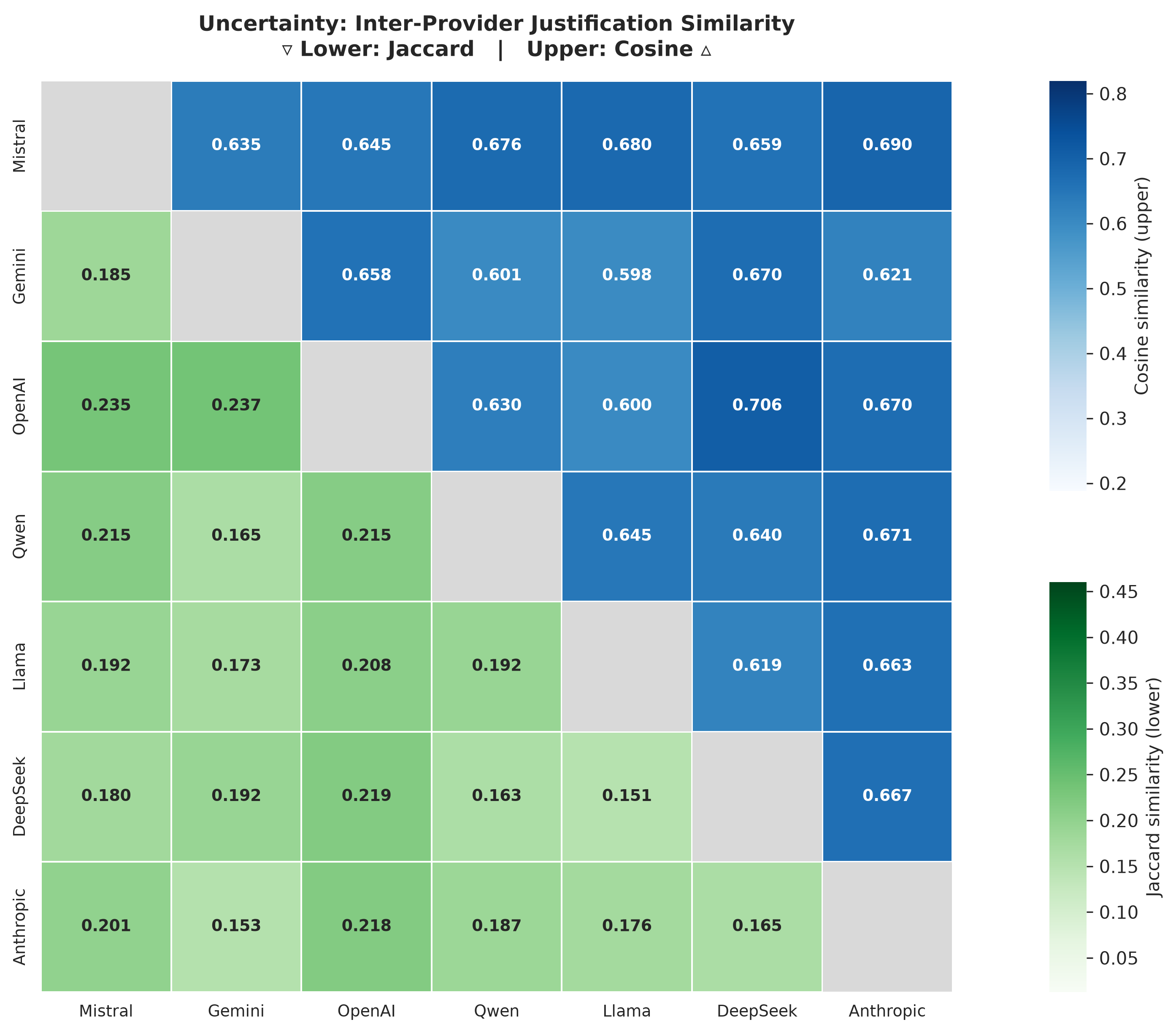}
    \caption{Textual similarity of the score justifications}
  \end{subfigure}
  \caption{Uncertainty}
\end{figure}

\begin{figure}[H]
  \centering
  \begin{subfigure}[b]{0.48\textwidth}
    \includegraphics[width=\textwidth]{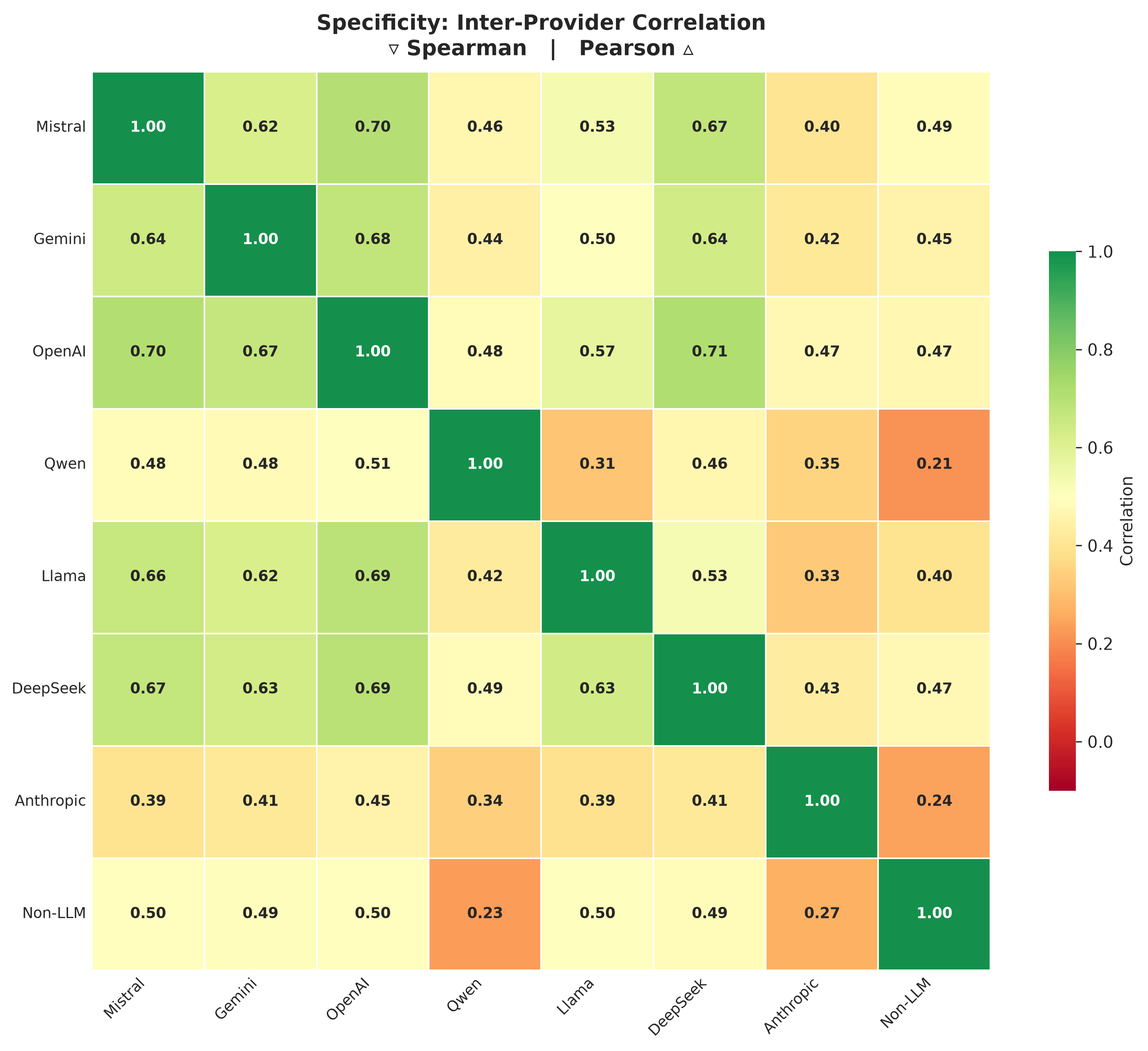}
    \caption{Score correlations}
  \end{subfigure}\hfill
  \begin{subfigure}[b]{0.48\textwidth}
    \includegraphics[width=\textwidth]{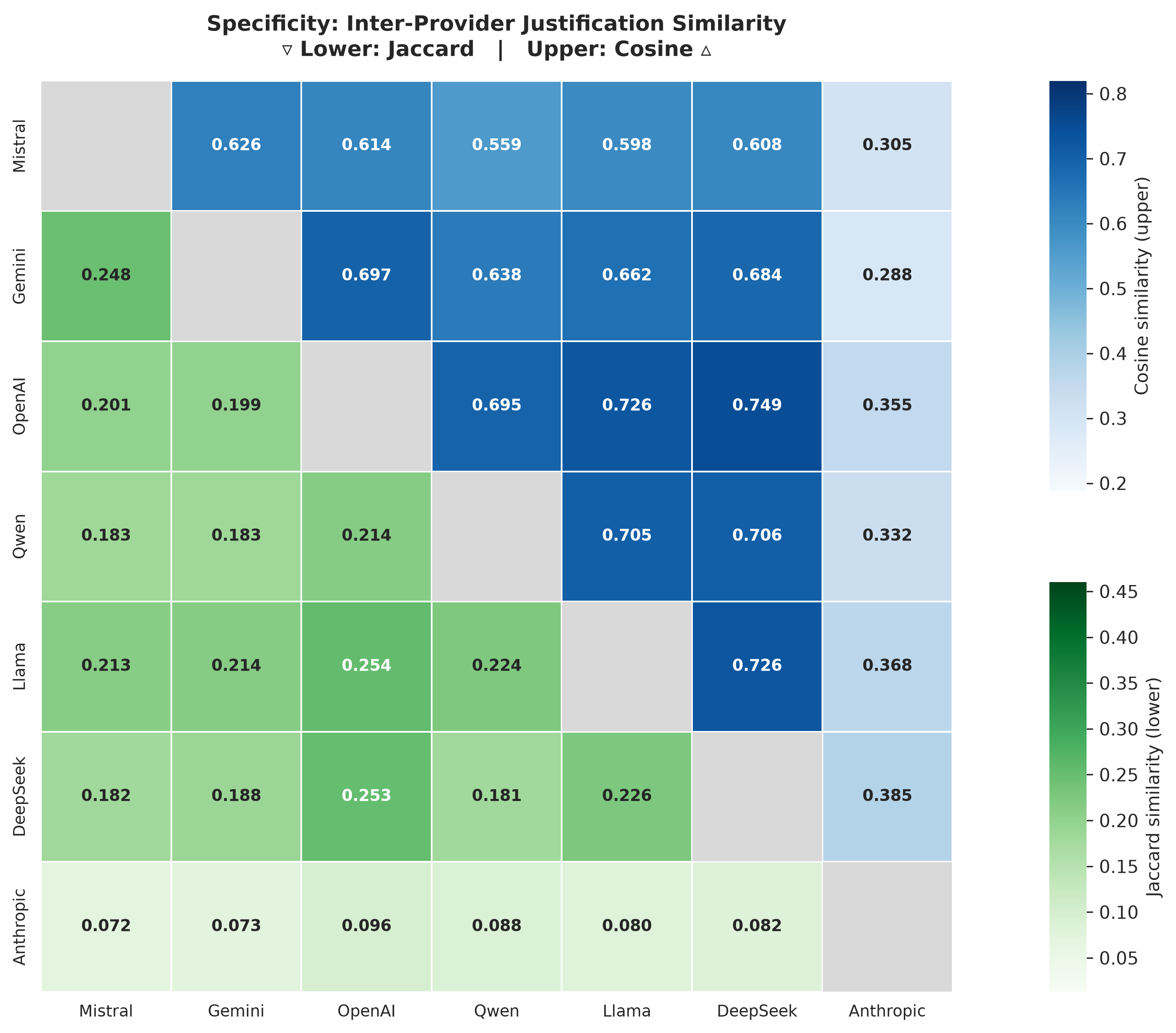}
    \caption{Textual similarity of the score justifications}
  \end{subfigure}
  \caption{Specificity}
\end{figure}

\begin{figure}[H]
  \centering
  \begin{subfigure}[b]{0.48\textwidth}
    \includegraphics[width=\textwidth]{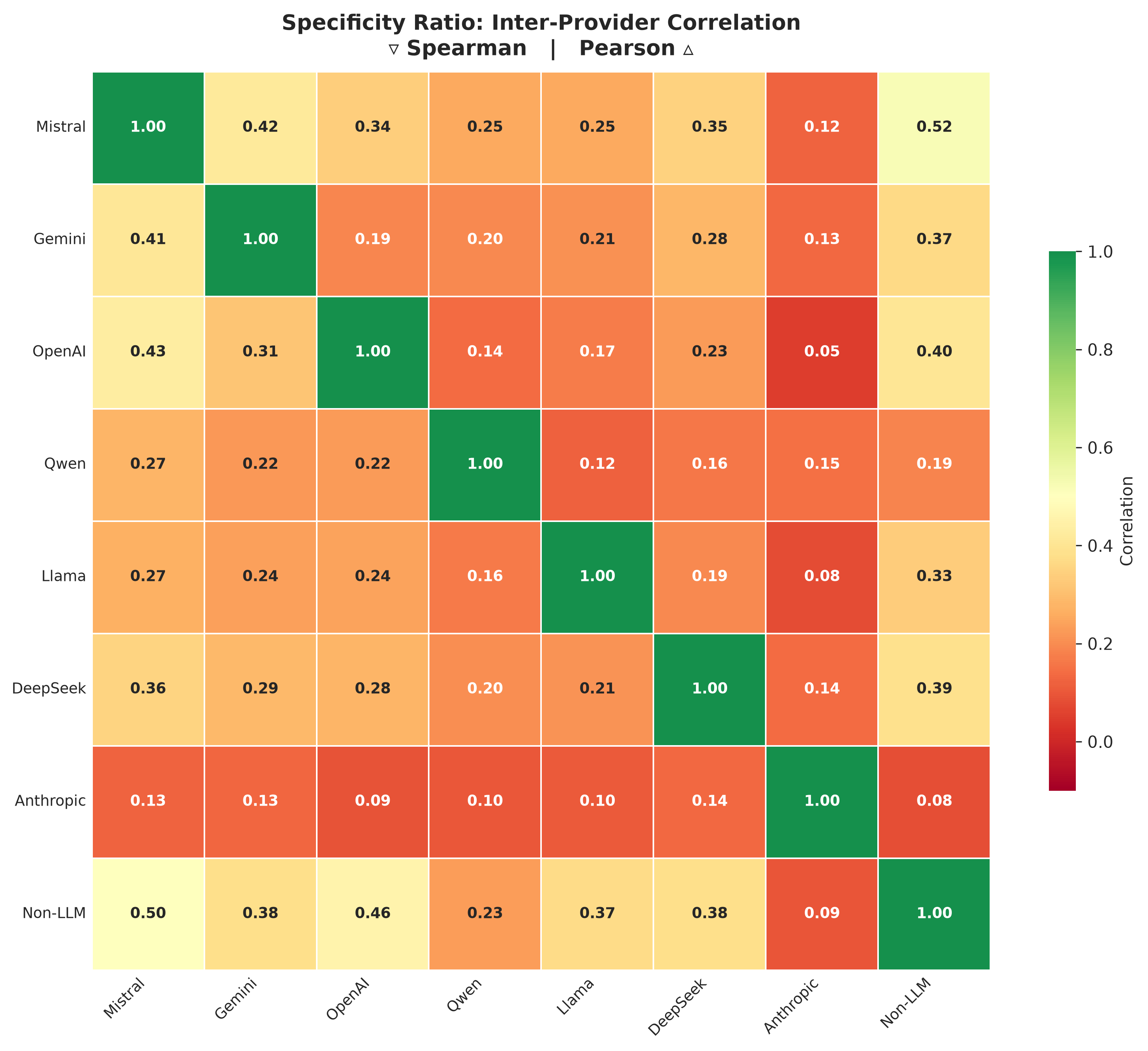}
    \caption{Score correlations}
  \end{subfigure}\hfill
  \begin{subfigure}[b]{0.48\textwidth}
    \includegraphics[width=\textwidth]{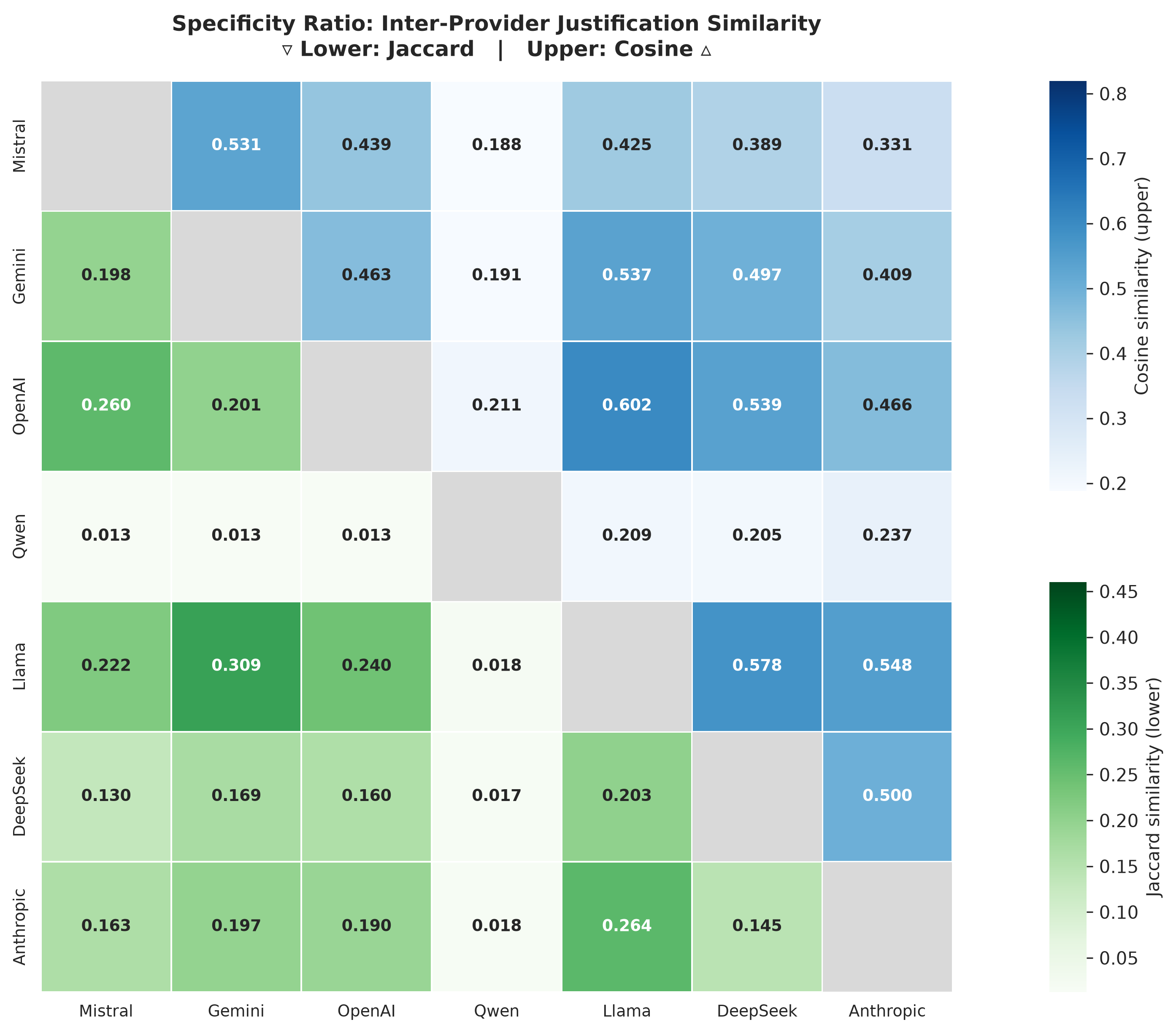}
    \caption{Textual similarity of the score justifications}
  \end{subfigure}
  \caption{Specificity Ratio}
\end{figure}

\begin{figure}[H]
  \centering
  \begin{subfigure}[b]{0.48\textwidth}
    \includegraphics[width=\textwidth]{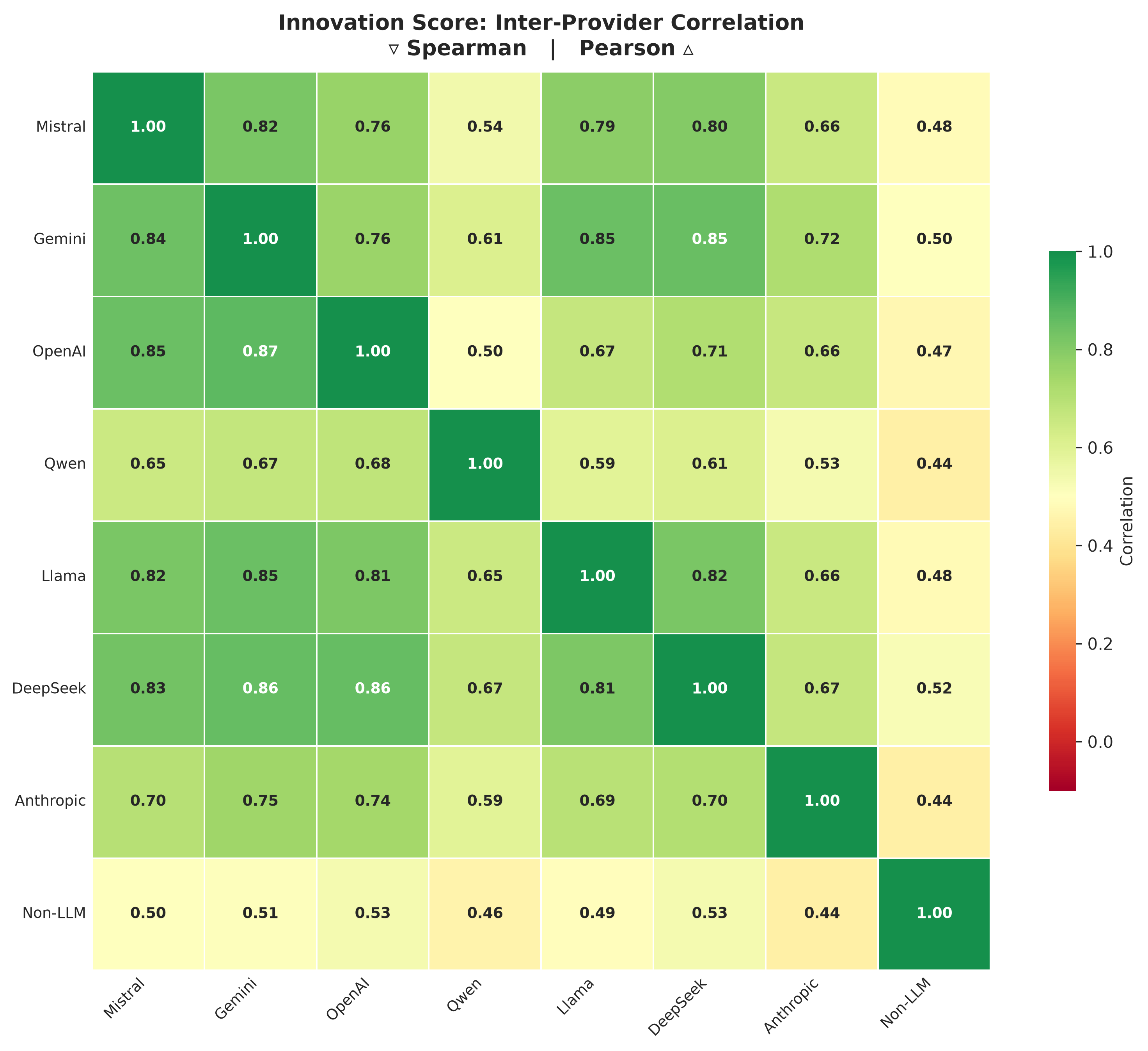}
    \caption{Score correlations}
  \end{subfigure}\hfill
  \begin{subfigure}[b]{0.48\textwidth}
    \includegraphics[width=\textwidth]{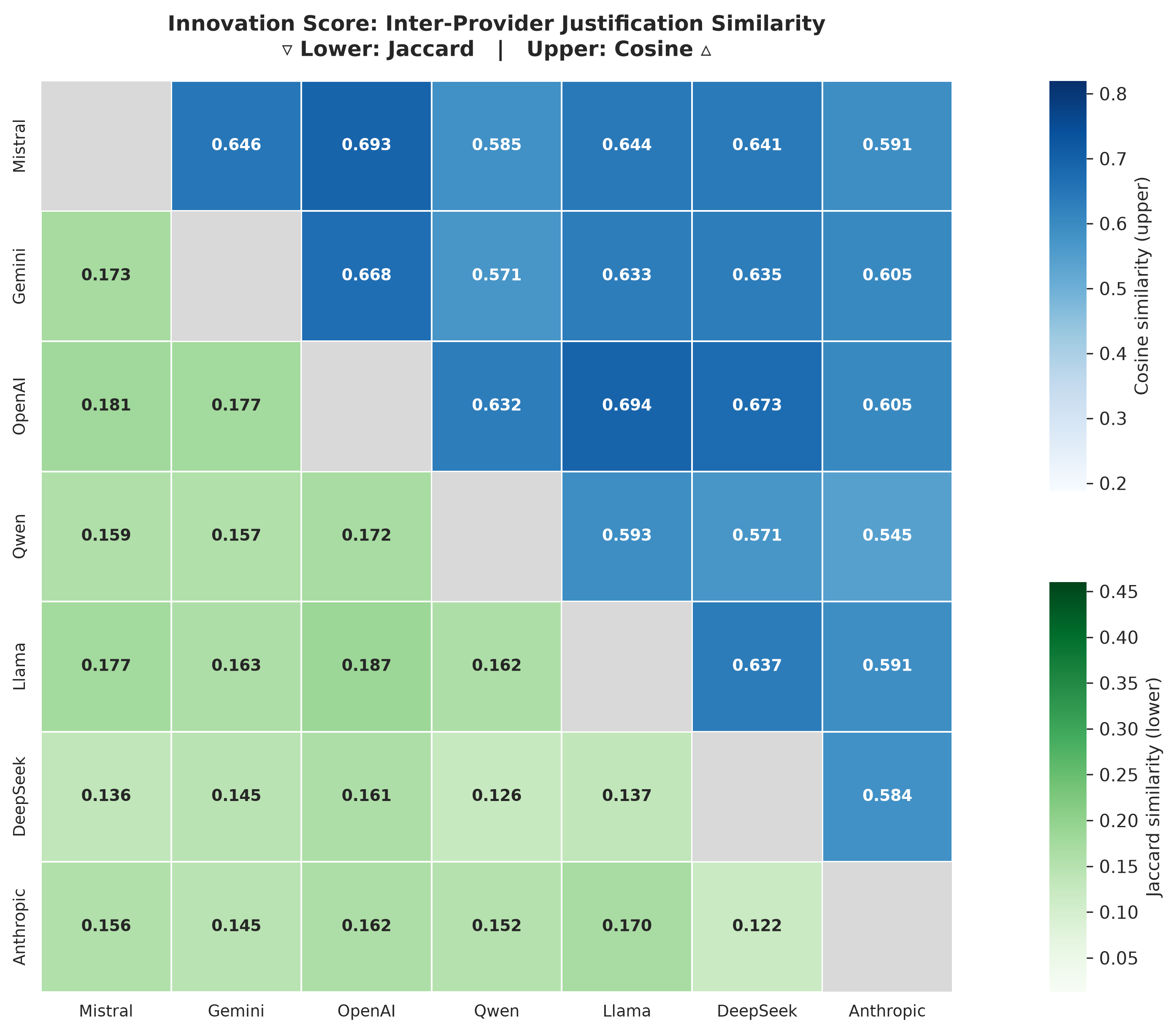}
    \caption{Textual similarity of the score justifications}
  \end{subfigure}
  \caption{Innovation Score}
\end{figure}

\begin{figure}[H]
  \centering
  \begin{subfigure}[b]{0.48\textwidth}
    \includegraphics[width=\textwidth]{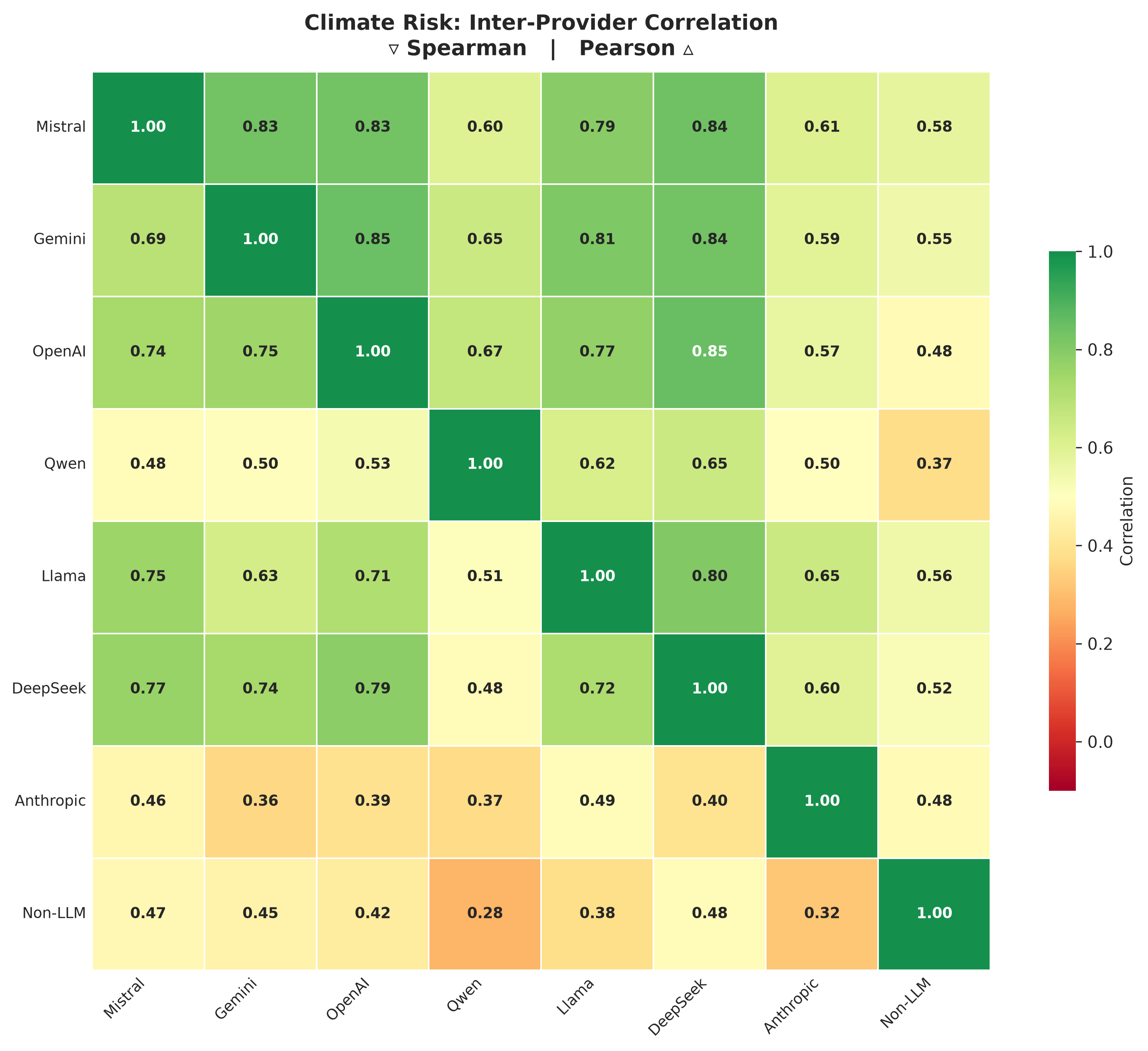}
    \caption{Score correlations}
  \end{subfigure}\hfill
  \begin{subfigure}[b]{0.48\textwidth}
    \includegraphics[width=\textwidth]{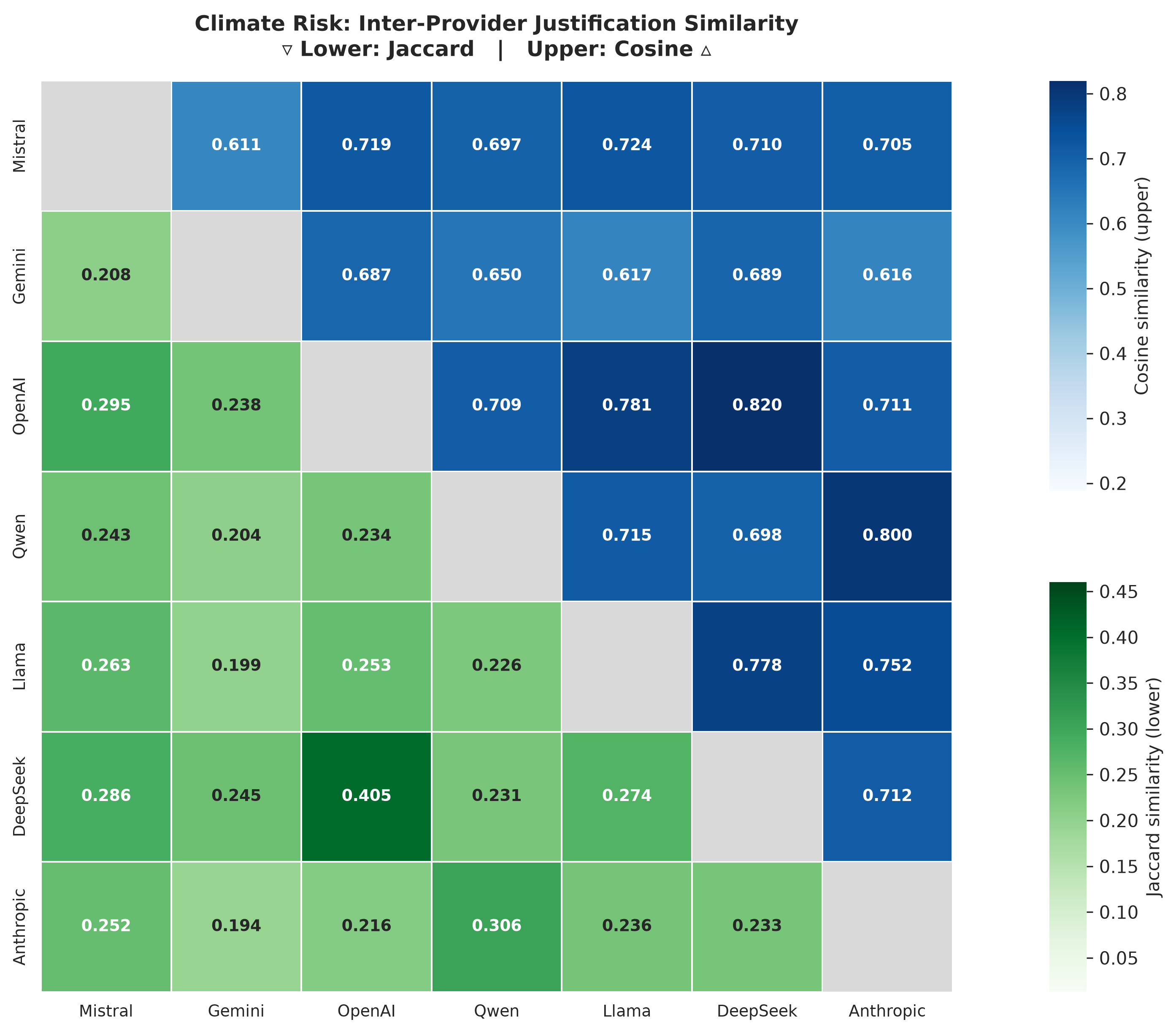}
    \caption{Textual similarity of the score justifications}
  \end{subfigure}
  \caption{Climate Risk}
\end{figure}

\begin{figure}[H]
  \centering
  \begin{subfigure}[b]{0.48\textwidth}
    \includegraphics[width=\textwidth]{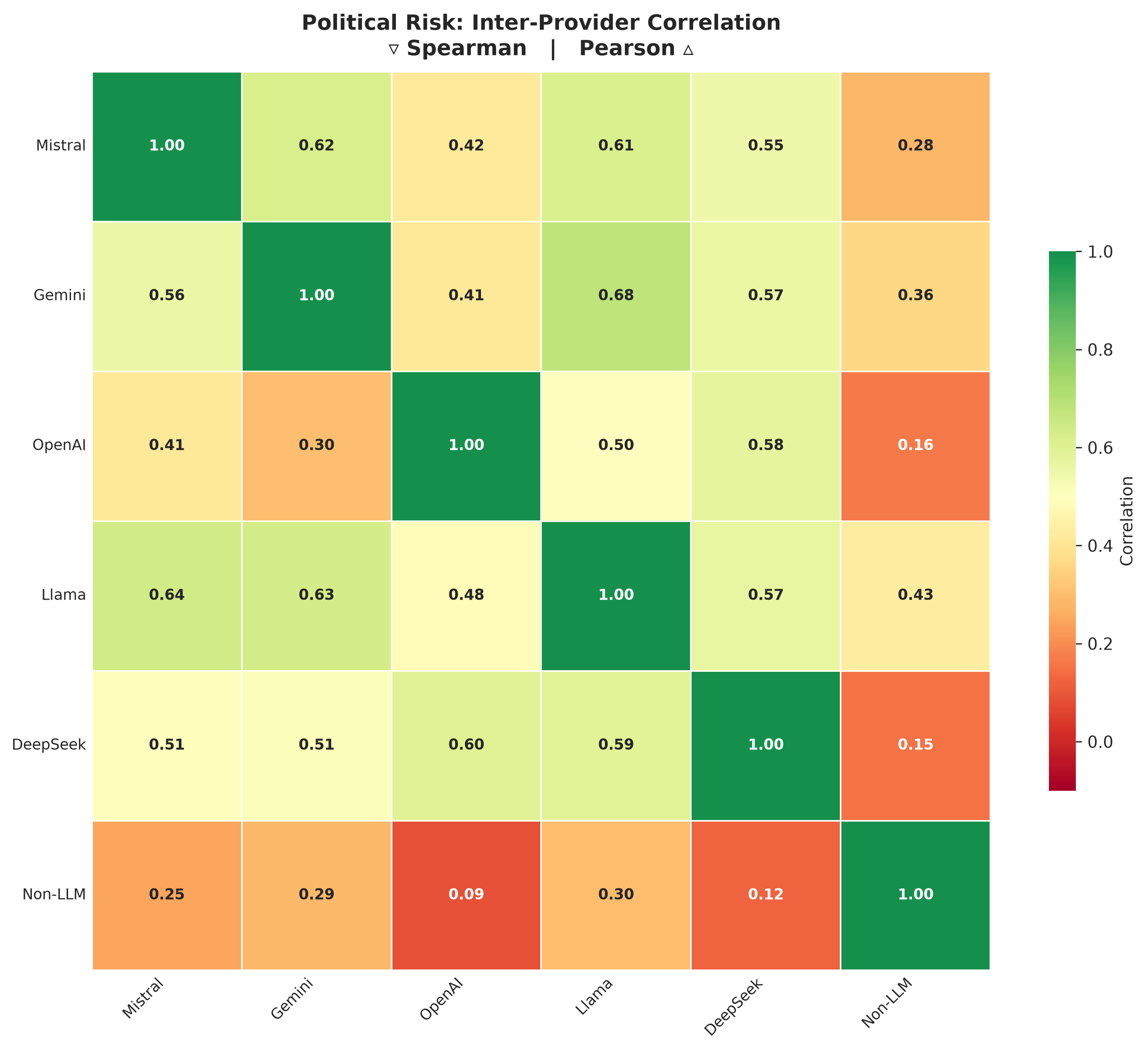}
    \caption{Score correlations}
  \end{subfigure}\hfill
  \begin{subfigure}[b]{0.48\textwidth}
    \includegraphics[width=\textwidth]{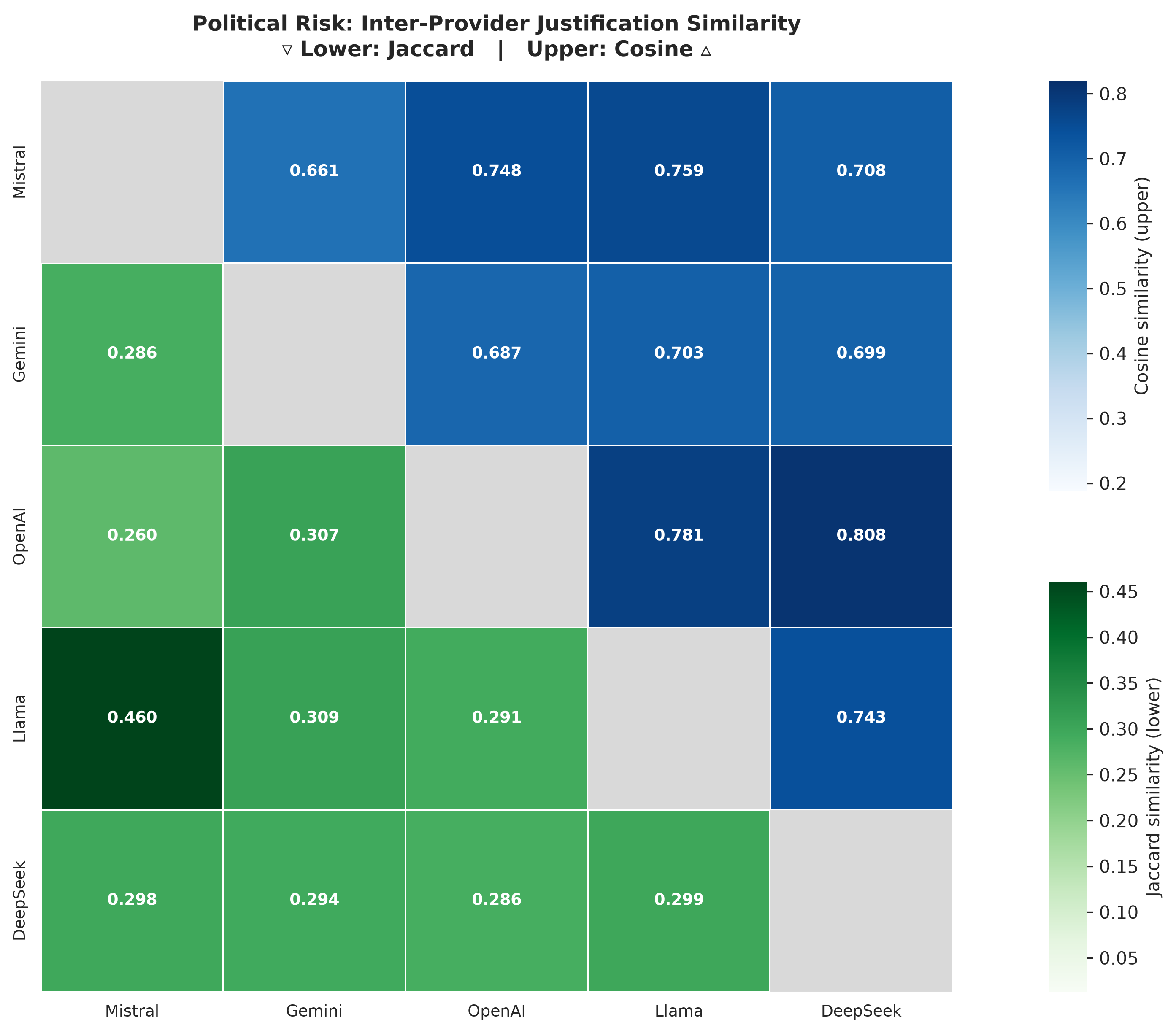}
    \caption{Textual similarity of the score justifications}
  \end{subfigure}
  \caption{Political Risk}
\end{figure}

\begin{figure}[H]
  \centering
  \begin{subfigure}[b]{0.48\textwidth}
    \includegraphics[width=\textwidth]{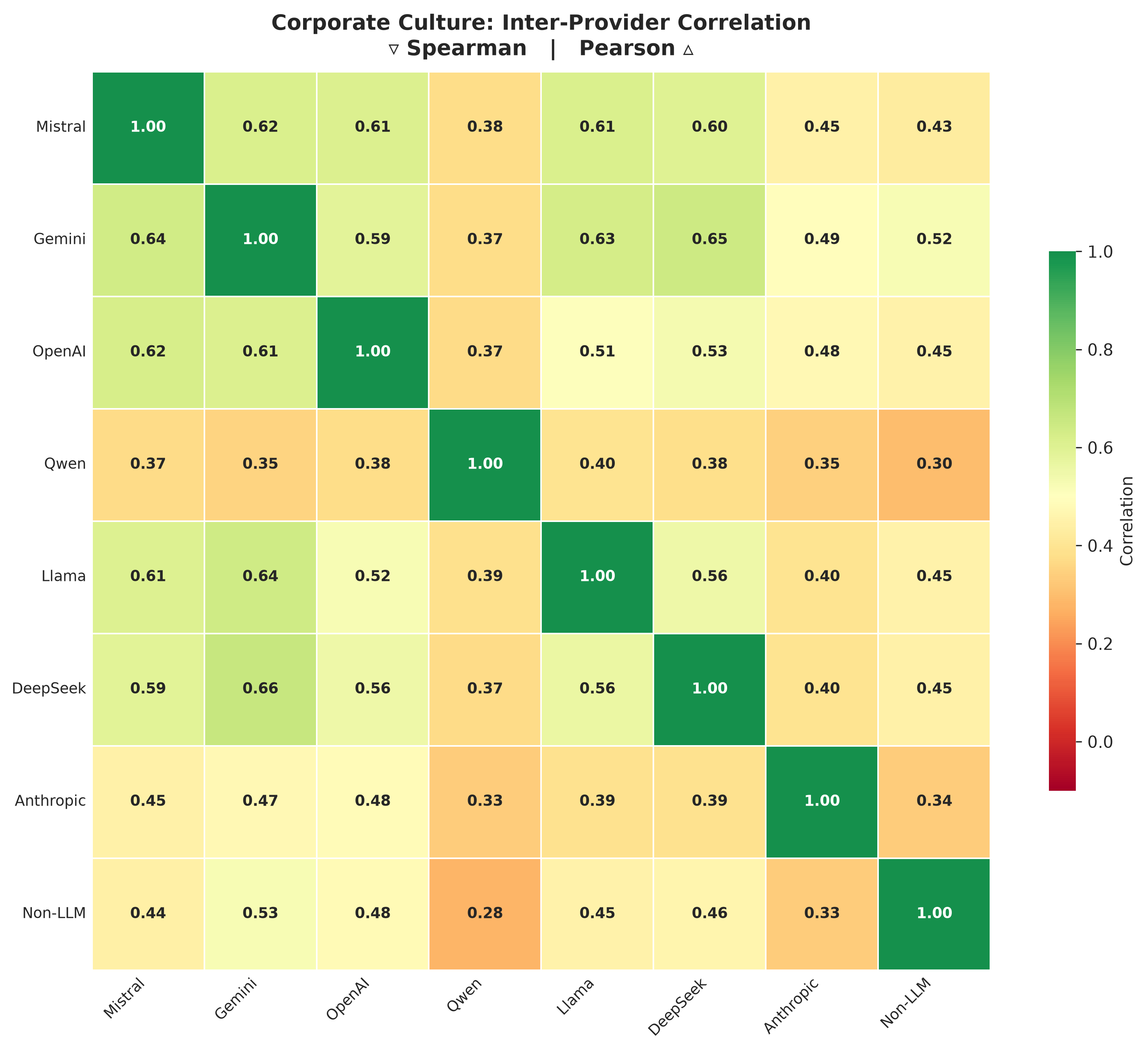}
    \caption{Score correlations}
  \end{subfigure}\hfill
  \begin{subfigure}[b]{0.48\textwidth}
    \includegraphics[width=\textwidth]{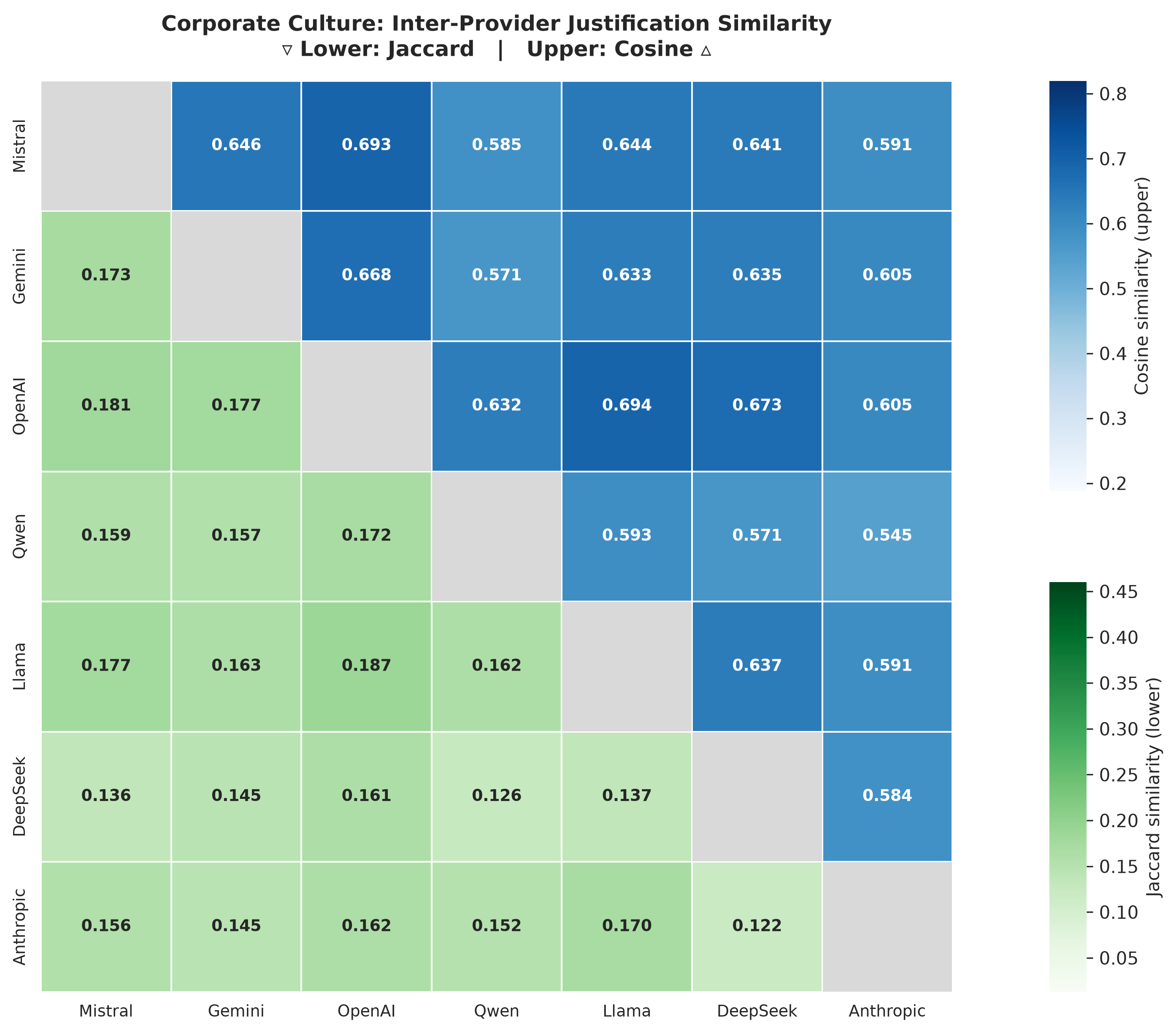}
    \caption{Textual similarity of the score justifications}
  \end{subfigure}
  \caption{Corporate Culture}
\end{figure}

\begin{figure}[H]
  \centering
  \begin{subfigure}[b]{0.48\textwidth}
    \includegraphics[width=\textwidth]{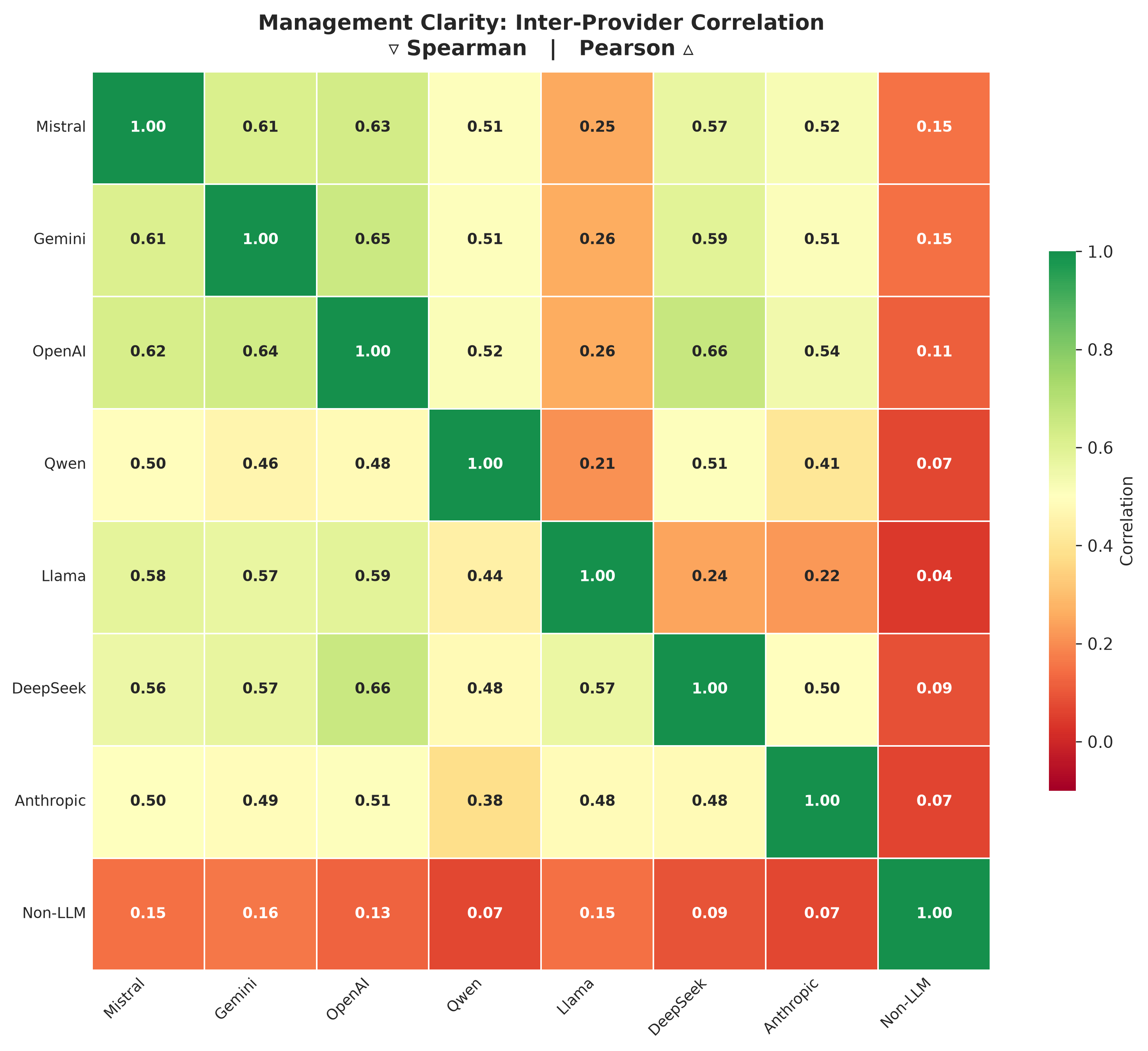}
    \caption{Score correlations}
  \end{subfigure}\hfill
  \begin{subfigure}[b]{0.48\textwidth}
    \includegraphics[width=\textwidth]{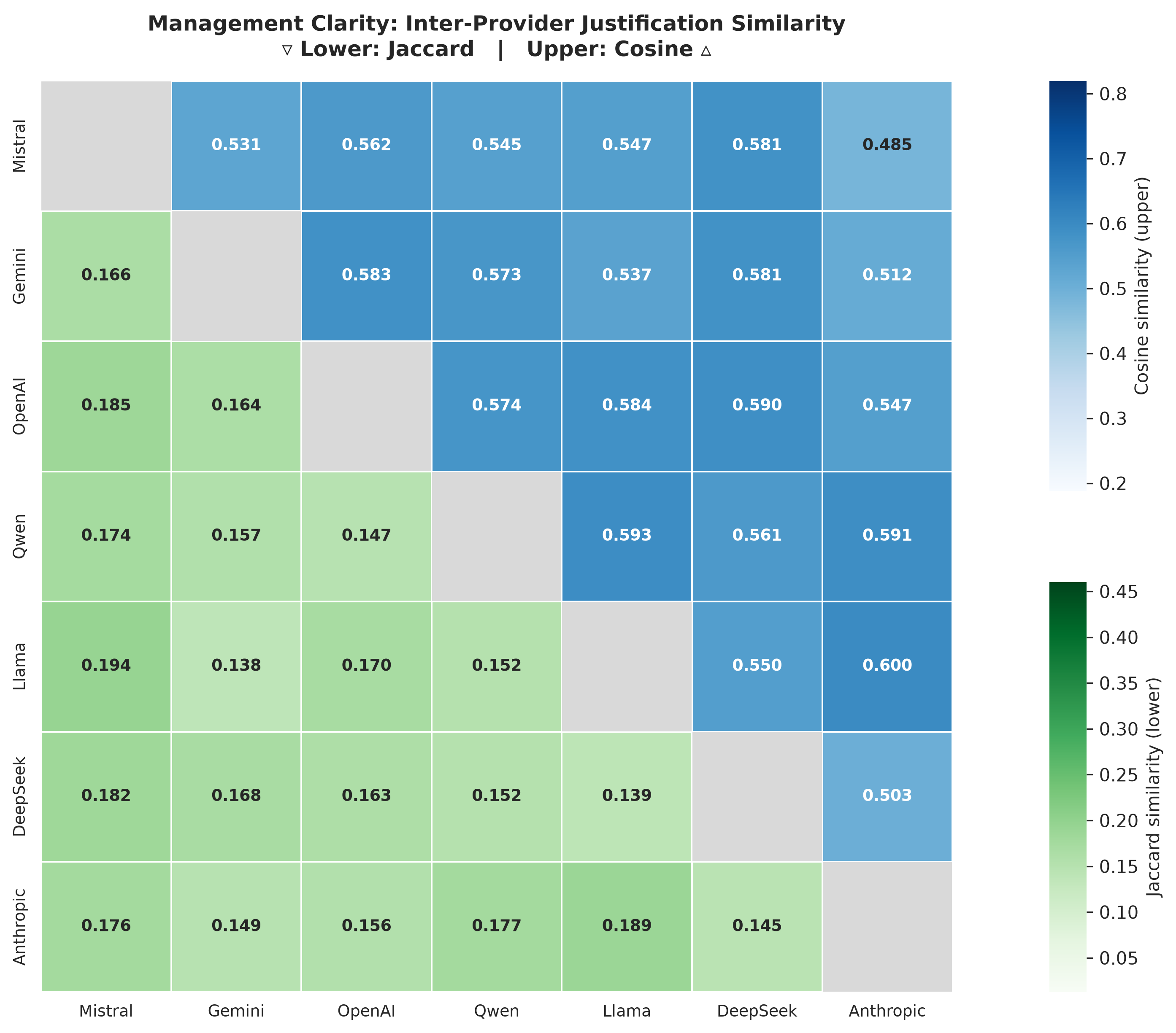}
    \caption{Textual similarity of the score justifications}
  \end{subfigure}
  \caption{Management Clarity}
\end{figure}

\begin{figure}[H]
  \centering
  \begin{subfigure}[b]{0.48\textwidth}
    \includegraphics[width=\textwidth]{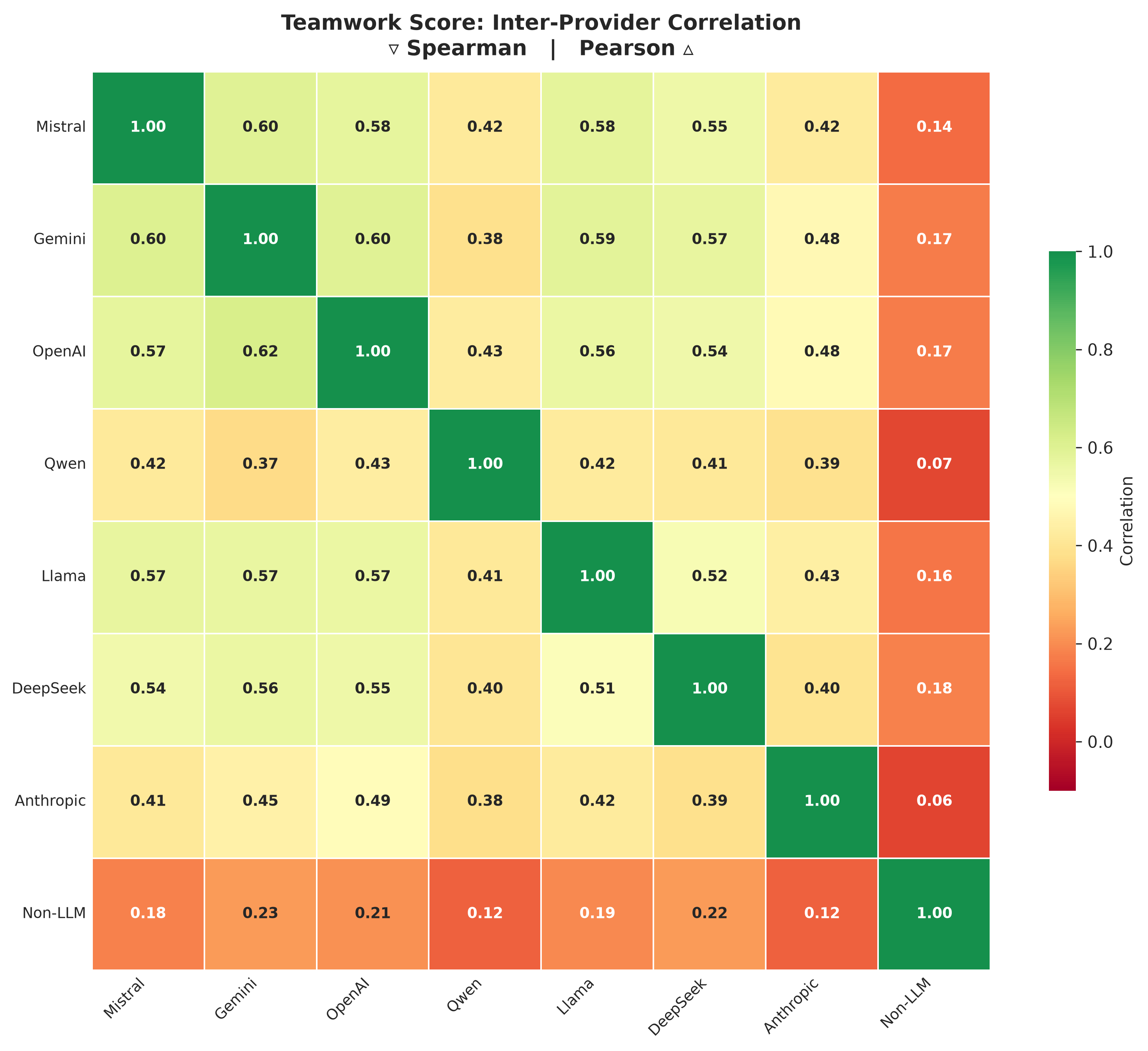}
    \caption{Score correlations}
  \end{subfigure}\hfill
  \begin{subfigure}[b]{0.48\textwidth}
    \includegraphics[width=\textwidth]{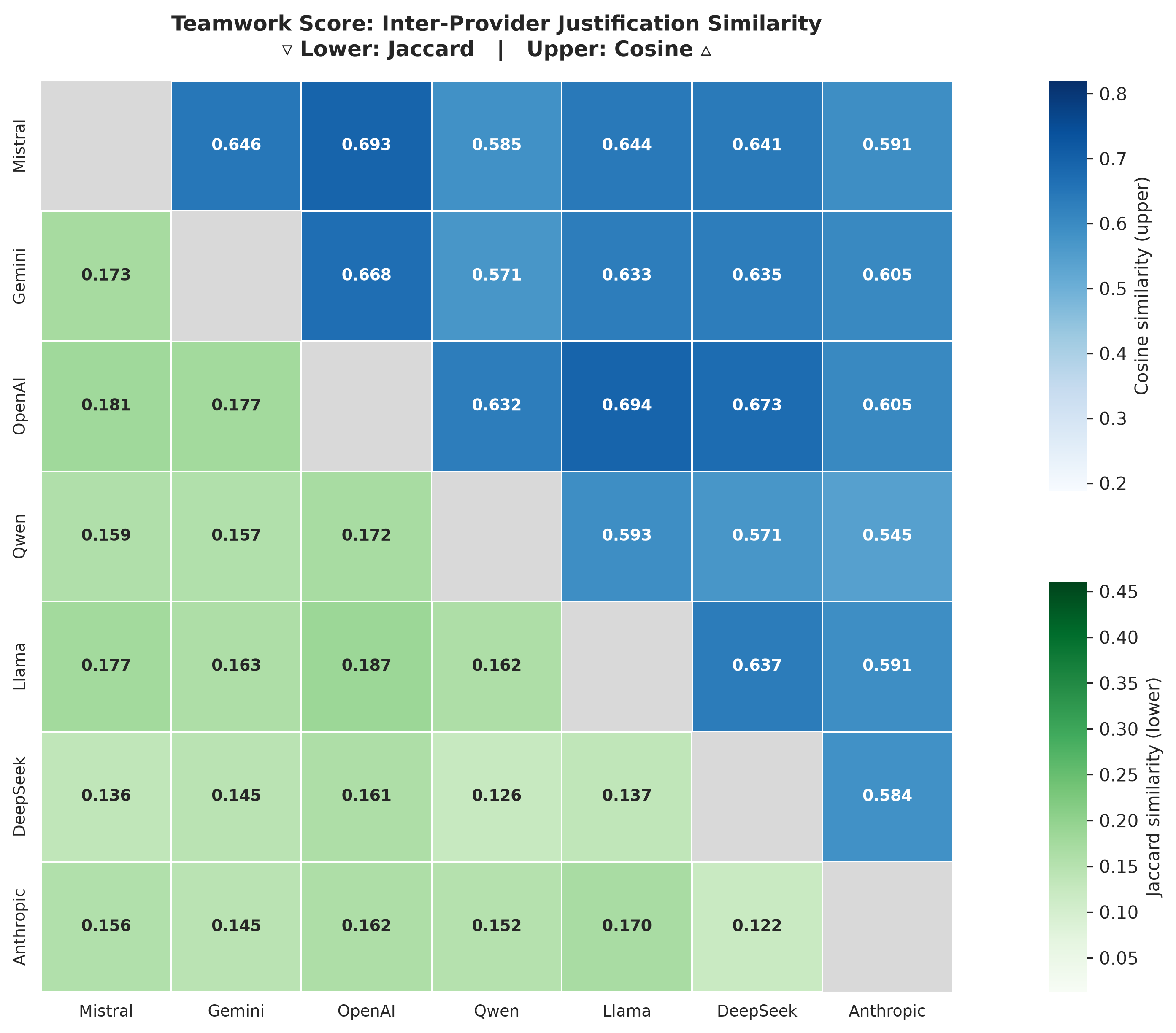}
    \caption{Textual similarity of the score justifications}
  \end{subfigure}
  \caption{Teamwork Score}
\end{figure}

\begin{figure}[H]
  \centering
  \begin{subfigure}[b]{0.48\textwidth}
    \includegraphics[width=\textwidth]{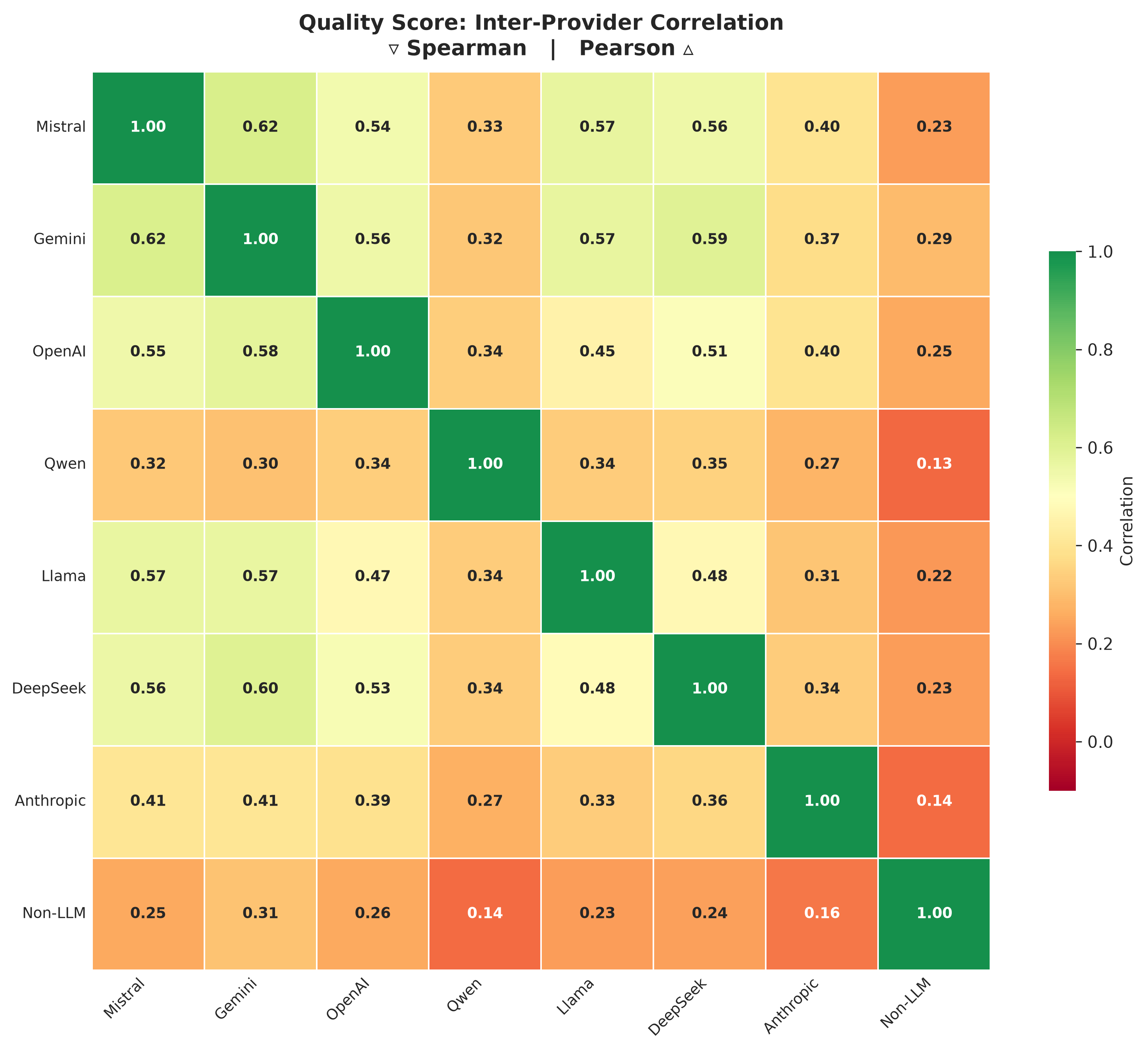}
    \caption{Score correlations}
  \end{subfigure}\hfill
  \begin{subfigure}[b]{0.48\textwidth}
    \includegraphics[width=\textwidth]{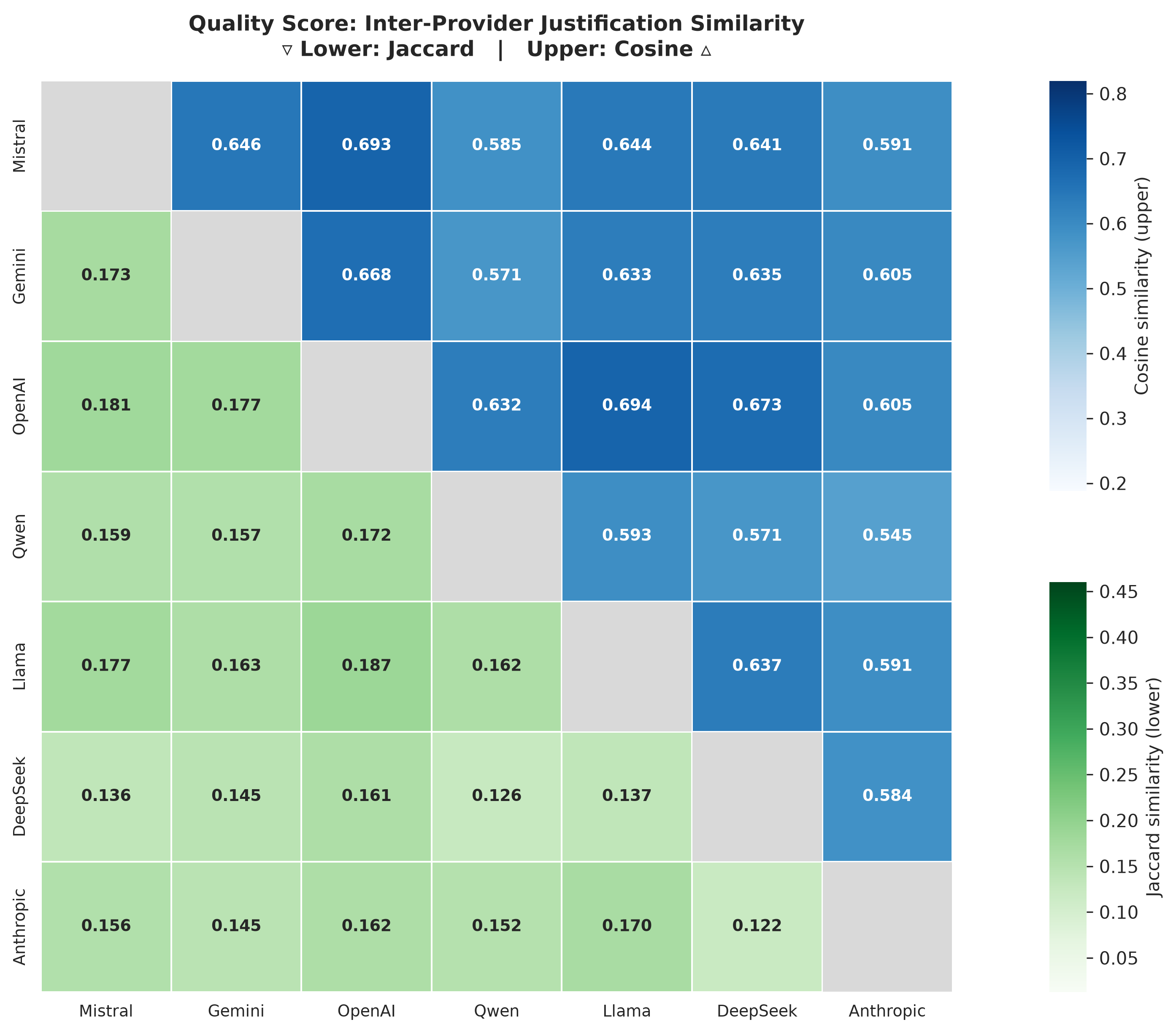}
    \caption{Textual similarity of the score justifications}
  \end{subfigure}
  \caption{Quality Score}
\end{figure}

\begin{figure}[H]
  \centering
  \begin{subfigure}[b]{0.48\textwidth}
    \includegraphics[width=\textwidth]{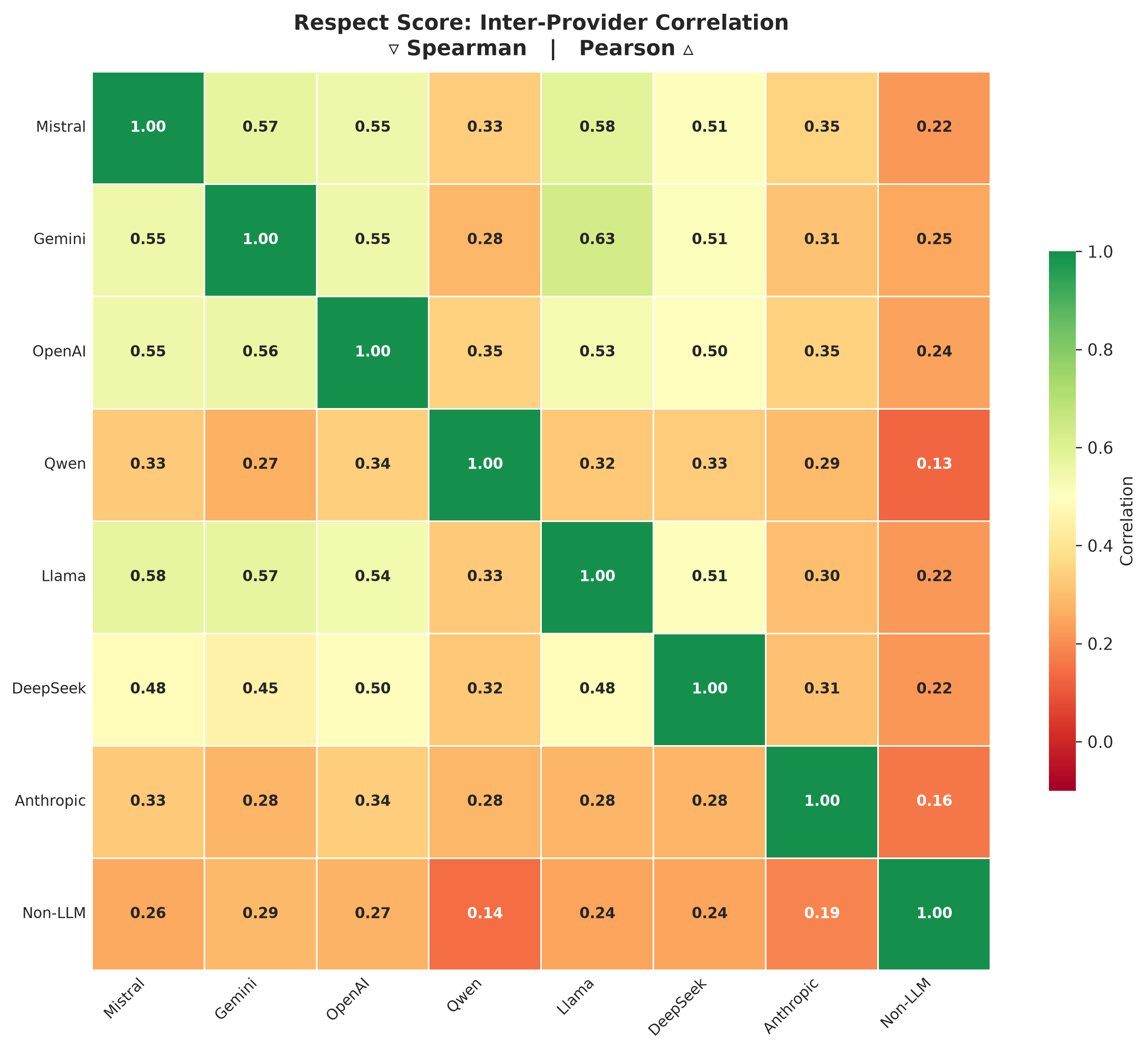}
    \caption{Score correlations}
  \end{subfigure}\hfill
  \begin{subfigure}[b]{0.48\textwidth}
    \includegraphics[width=\textwidth]{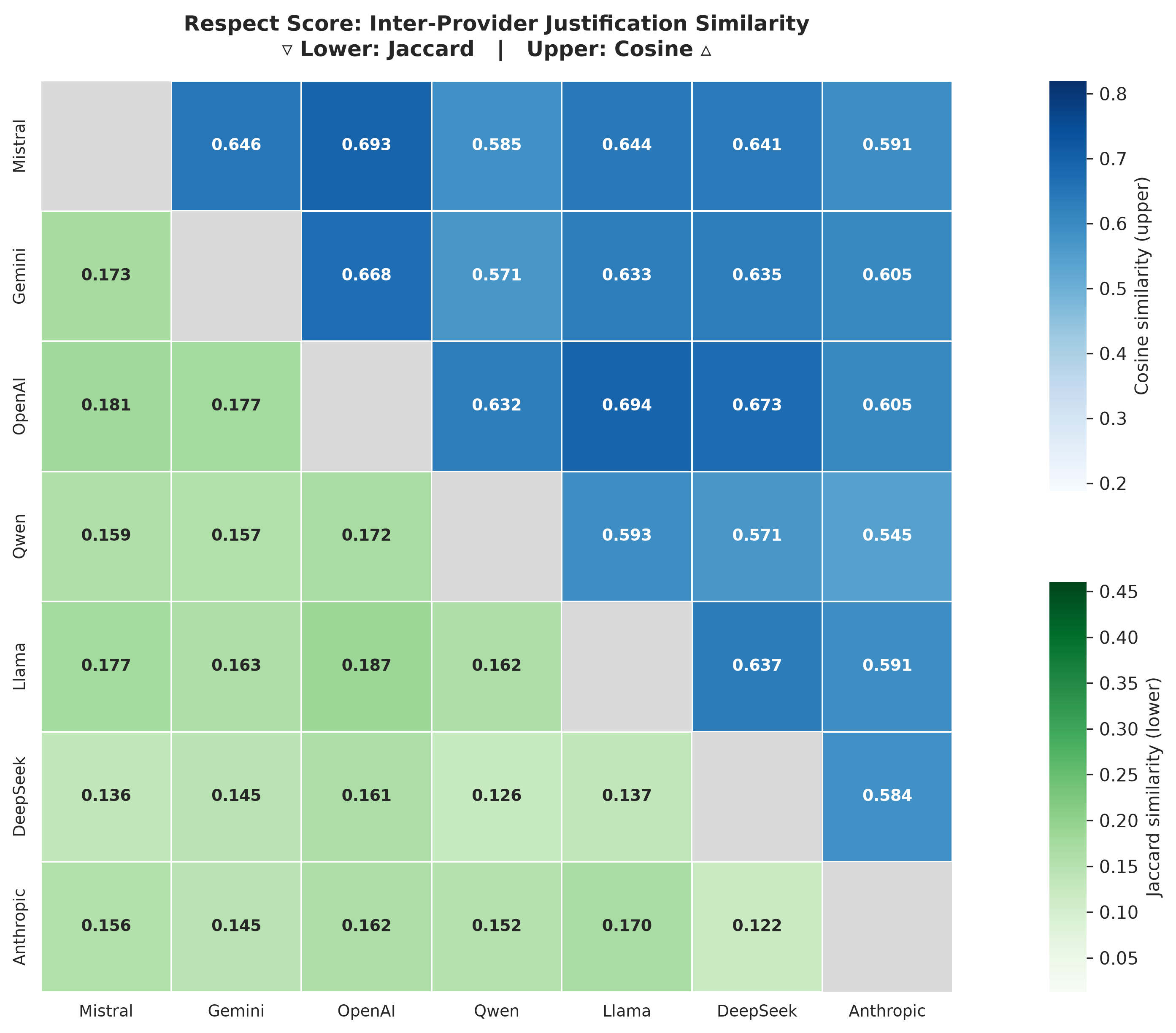}
    \caption{Textual similarity of the score justifications}
  \end{subfigure}
  \caption{Respect Score}
\end{figure}

\begin{figure}[H]
  \centering
  \begin{subfigure}[b]{0.48\textwidth}
    \includegraphics[width=\textwidth]{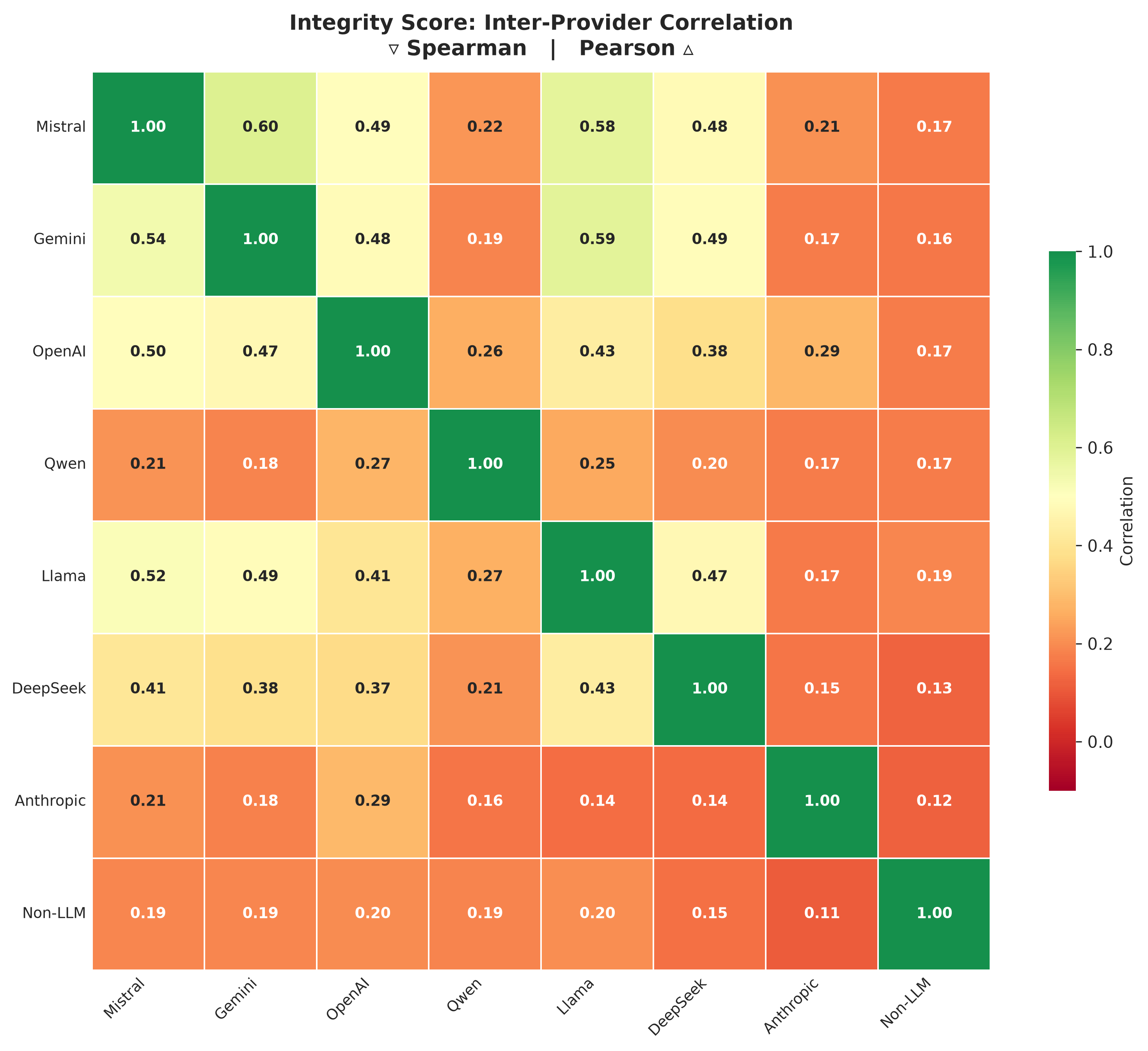}
    \caption{Score correlations}
  \end{subfigure}\hfill
  \begin{subfigure}[b]{0.48\textwidth}
    \includegraphics[width=\textwidth]{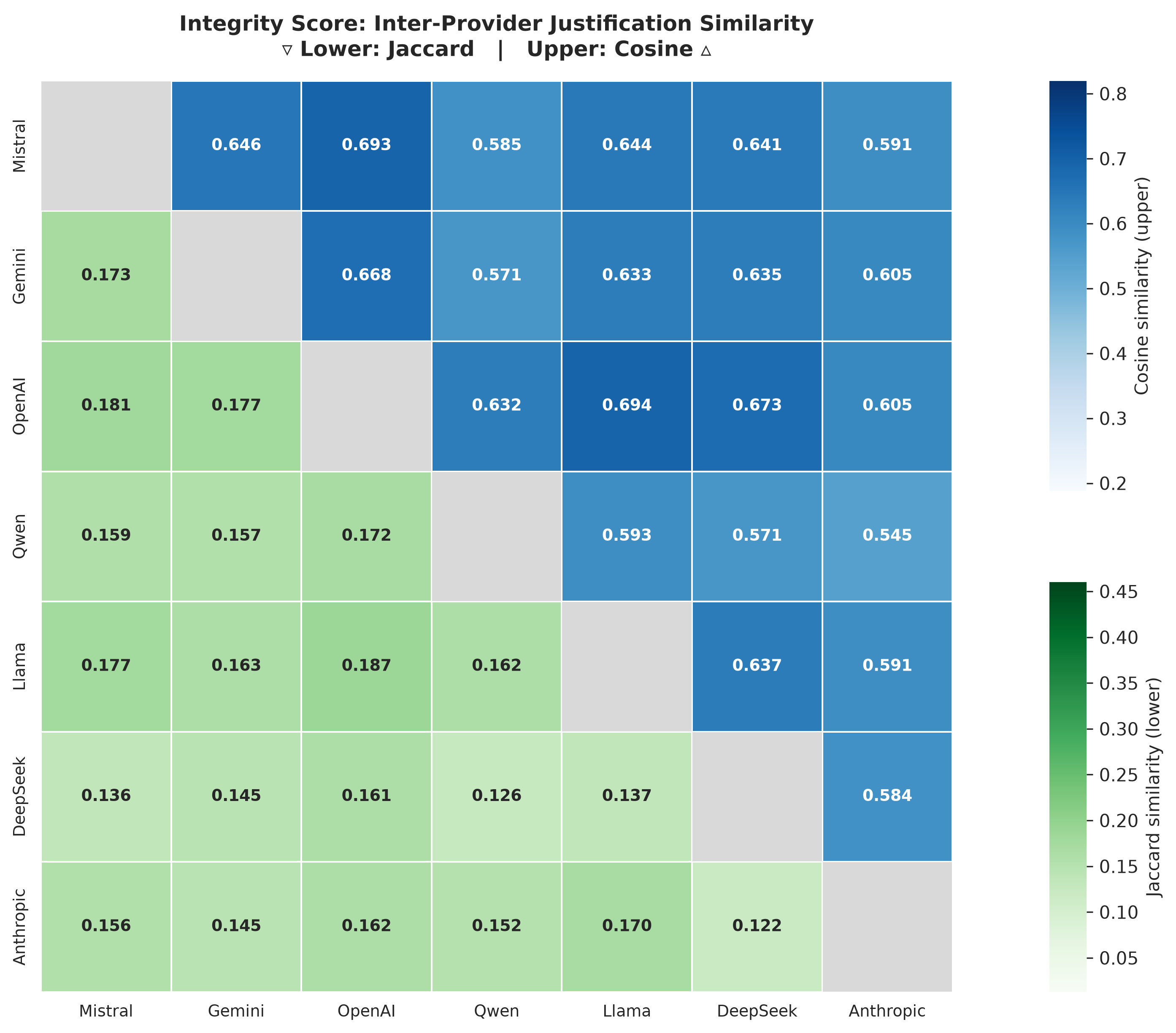}
    \caption{Textual similarity of the score justifications}
  \end{subfigure}
  \caption{Integrity Score}
\end{figure}

\clearpage

\section{Summary Statistics --- Full Detail}
\label{app:sumstats}

\begin{ThreePartTable}
\begin{TableNotes}[flushleft]\footnotesize
  \item \textit{Notes:} Summary statistics for variables in the main analyses. Variable definitions appear in Table~\ref{tab:varlist}. LLM scores are z-standardized within task. Firm-level and textual variables repeat across the 13 constructs in this pooled panel; call-level statistics are in Table~\ref{tab:summary_stats}.
\end{TableNotes}
\small
\begin{longtable}{lrrrrrrrr}

  \caption{Summary Statistics --- Full Detail (incl.\ per-construct LLM scores)}
  \label{tab:summary_stats_full} \\

  \toprule
   & N & Mean & Std & Min & P25 & Median & P75 & Max \\
  \midrule
  \endfirsthead
  \multicolumn{9}{l}{\textit{(Table \ref{tab:summary_stats_full} continued)}} \\[2pt]
  \toprule
   & N & Mean & Std & Min & P25 & Median & P75 & Max \\
  \midrule
  \endhead
  \midrule
  \multicolumn{9}{r}{\textit{continued on next page}} \\
  \endfoot
  \bottomrule
  \insertTableNotes
  \endlastfoot
  \multicolumn{9}{l}{\textit{LLM-related variables (call $\times$ construct level)}} \\[2pt]
  LLM Dispersion (Z) & 25,298 & 0.618 & 0.315 & 0.067 & 0.415 & 0.583 & 0.784 & 2.787 \\
  Justif. Cosine Sim & 25,298 & 0.618 & 0.102 & 0.204 & 0.564 & 0.624 & 0.682 & 0.931 \\
  Justif. Jaccard Sim & 25,298 & 0.183 & 0.061 & 0.010 & 0.150 & 0.169 & 0.193 & 0.584 \\
  \\[-6pt]
  \multicolumn{9}{l}{\textit{Firm-level variables (call level)}} \\[2pt]
  ln(Assets) & 25,285 & 10.348 & 1.281 & 7.288 & 9.498 & 10.246 & 11.102 & 15.253 \\
  Book-to-Market & 25,246 & 0.328 & 0.308 & -0.638 & 0.113 & 0.257 & 0.483 & 2.380 \\
  Tobin's Q & 25,246 & 2.892 & 2.712 & 0.760 & 1.345 & 1.961 & 3.386 & 34.202 \\
  SUE & 24,674 & 2.107 & 3.176 & -5.769 & 0.279 & 1.634 & 3.300 & 14.696 \\
  CAR(0,1) & 25,155 & 0.020 & 6.540 & -20.733 & -3.922 & -0.106 & 4.069 & 17.321 \\
  Analyst Disp. (Q1) & 25,155 & 0.070 & 0.103 & 0.003 & 0.019 & 0.037 & 0.076 & 0.693 \\
  Analyst Count (Q1) & 25,181 & 15.126 & 6.578 & 3.000 & 11.000 & 15.000 & 19.000 & 36.000 \\
  Pre-Call Analyst Disp. & 25,129 & 0.071 & 0.111 & 0.003 & 0.019 & 0.037 & 0.074 & 0.799 \\
  Pre-Call Analyst Count & 25,142 & 15.989 & 6.420 & 4.000 & 11.000 & 16.000 & 20.000 & 38.000 \\
  \\[-6pt]
  \multicolumn{9}{l}{\textit{Textual variables (call level)}} \\[2pt]
  Fog Index & 25,298 & 7.381 & 0.835 & 5.112 & 6.809 & 7.301 & 7.868 & 11.633 \\
  ln(Word Count) & 25,298 & 8.243 & 0.352 & 4.682 & 8.067 & 8.278 & 8.461 & 9.091 \\
  Financial Jargon & 25,298 & 0.063 & 0.011 & 0.032 & 0.056 & 0.063 & 0.070 & 0.110 \\
  Numbers & 25,298 & 0.012 & 0.005 & 0.000 & 0.008 & 0.011 & 0.015 & 0.036 \\
  Forward-Looking & 25,298 & 0.057 & 0.012 & 0.023 & 0.049 & 0.056 & 0.064 & 0.187 \\
  Specificity & 25,298 & 0.013 & 0.004 & 0.004 & 0.011 & 0.013 & 0.016 & 0.040 \\
  \\[-6pt]
  \multicolumn{9}{l}{\textit{LLM scores (z-scored, by task)}} \\[2pt]
  \quad \textit{Climate Risk} \\[1pt]
  \quad\quad Mistral & 1,946 & -0.000 & 1.000 & -0.511 & -0.511 & -0.511 & 0.582 & 3.863 \\
  \quad\quad Gemini & 1,946 & -0.000 & 1.000 & -0.751 & -0.751 & -0.056 & -0.056 & 4.118 \\
  \quad\quad OpenAI & 1,946 & -0.000 & 1.000 & -0.633 & -0.633 & -0.633 & 0.634 & 4.117 \\
  \quad\quad Qwen & 1,946 & 0.000 & 1.000 & -0.501 & -0.501 & -0.501 & 0.499 & 4.497 \\
  \quad\quad Llama & 1,946 & 0.000 & 1.000 & -0.339 & -0.339 & -0.339 & -0.339 & 5.121 \\
  \quad\quad DeepSeek & 1,946 & 0.000 & 1.000 & -0.594 & -0.594 & -0.594 & 0.383 & 5.048 \\
  \quad\quad Anthropic & 1,946 & -0.000 & 1.000 & -0.233 & -0.233 & -0.233 & -0.233 & 4.292 \\
  \\[-6pt]
  \quad \textit{Corporate Culture} \\[1pt]
  \quad\quad Mistral & 1,938 & 0.000 & 1.000 & -2.057 & -0.774 & 0.152 & 0.650 & 1.933 \\
  \quad\quad Gemini & 1,946 & -0.000 & 1.000 & -2.064 & -0.681 & -0.012 & 0.635 & 2.662 \\
  \quad\quad OpenAI & 1,946 & 0.000 & 1.000 & -2.994 & -0.461 & 0.298 & 0.636 & 1.565 \\
  \quad\quad Qwen & 1,946 & -0.000 & 1.000 & -2.457 & -0.704 & 0.173 & 0.757 & 2.101 \\
  \quad\quad Llama & 1,946 & -0.000 & 1.000 & -2.189 & -0.760 & 0.097 & 0.669 & 2.170 \\
  \quad\quad DeepSeek & 1,946 & 0.000 & 1.000 & -1.801 & -0.849 & 0.035 & 0.715 & 2.143 \\
  \quad\quad Anthropic & 1,946 & -0.000 & 1.000 & -3.011 & -0.942 & 0.207 & 0.590 & 1.816 \\
  \\[-6pt]
  \quad \textit{Innovation} \\[1pt]
  \quad\quad Mistral & 1,946 & -0.000 & 1.000 & -1.725 & -1.327 & 0.263 & 1.059 & 1.456 \\
  \quad\quad Gemini & 1,946 & 0.000 & 1.000 & -1.782 & -1.004 & -0.226 & 0.942 & 1.720 \\
  \quad\quad OpenAI & 1,946 & -0.000 & 1.000 & -2.392 & -0.243 & 0.294 & 0.563 & 1.368 \\
  \quad\quad Qwen & 1,946 & -0.000 & 1.000 & -1.615 & -0.245 & -0.245 & 0.441 & 3.868 \\
  \quad\quad Llama & 1,946 & 0.000 & 1.000 & -1.589 & -0.816 & -0.816 & 0.730 & 1.503 \\
  \quad\quad DeepSeek & 1,946 & 0.000 & 1.000 & -1.604 & -0.756 & -0.120 & 0.940 & 1.788 \\
  \quad\quad Anthropic & 1,946 & -0.000 & 1.000 & -2.791 & -0.085 & -0.085 & 0.998 & 1.539 \\
  \\[-6pt]
  \quad \textit{Integrity} \\[1pt]
  \quad\quad Mistral & 1,942 & -0.000 & 1.000 & -1.664 & -0.433 & -0.433 & 0.798 & 2.644 \\
  \quad\quad Gemini & 1,946 & 0.000 & 1.000 & -0.664 & -0.664 & -0.664 & 0.236 & 4.739 \\
  \quad\quad OpenAI & 1,946 & 0.000 & 1.000 & -2.608 & -0.232 & -0.232 & 0.560 & 2.144 \\
  \quad\quad Qwen & 1,946 & 0.000 & 1.000 & -2.371 & -0.880 & 0.115 & 0.612 & 1.855 \\
  \quad\quad Llama & 1,946 & -0.000 & 1.000 & -1.246 & -0.609 & 0.028 & 0.028 & 3.848 \\
  \quad\quad DeepSeek & 1,946 & -0.000 & 1.000 & -0.999 & -0.999 & -0.379 & 0.241 & 3.030 \\
  \quad\quad Anthropic & 1,946 & 0.000 & 1.000 & -3.323 & -0.468 & -0.468 & 0.959 & 1.245 \\
  \\[-6pt]
  \quad \textit{Management Clarity} \\[1pt]
  \quad\quad Mistral & 1,946 & -0.000 & 1.000 & -1.688 & 0.118 & 0.118 & 0.118 & 1.924 \\
  \quad\quad Gemini & 1,946 & -0.000 & 1.000 & -2.797 & -0.051 & -0.051 & 1.048 & 1.048 \\
  \quad\quad OpenAI & 1,946 & -0.000 & 1.000 & -2.738 & -0.100 & -0.100 & 1.219 & 1.219 \\
  \quad\quad Qwen & 1,946 & -0.000 & 1.000 & -2.328 & -0.777 & 0.774 & 0.774 & 0.774 \\
  \quad\quad Llama & 1,946 & -0.000 & 1.000 & -4.846 & 0.021 & 0.021 & 0.697 & 0.899 \\
  \quad\quad DeepSeek & 1,946 & -0.000 & 1.000 & -2.712 & -0.526 & 0.567 & 0.567 & 1.660 \\
  \quad\quad Anthropic & 1,946 & 0.000 & 1.000 & -2.360 & -0.542 & -0.542 & 1.276 & 2.185 \\
  \\[-6pt]
  \quad \textit{Political Risk} \\[1pt]
  \quad\quad Mistral & 1,946 & -0.000 & 1.000 & -0.394 & -0.394 & -0.394 & -0.394 & 2.880 \\
  \quad\quad Gemini & 1,946 & 0.000 & 1.000 & -0.656 & -0.656 & -0.656 & 0.127 & 4.041 \\
  \quad\quad OpenAI & 1,946 & 0.000 & 1.000 & -0.294 & -0.294 & -0.294 & -0.294 & 5.757 \\
  \quad\quad Llama & 1,946 & 0.000 & 1.000 & -0.394 & -0.394 & -0.394 & -0.394 & 5.077 \\
  \quad\quad DeepSeek & 1,946 & -0.000 & 1.000 & -0.435 & -0.435 & -0.435 & -0.435 & 4.351 \\
  \\[-6pt]
  \quad \textit{Quality} \\[1pt]
  \quad\quad Mistral & 1,942 & -0.000 & 1.000 & -2.468 & -0.633 & -0.021 & 0.591 & 2.426 \\
  \quad\quad Gemini & 1,946 & -0.000 & 1.000 & -2.014 & -0.963 & 0.088 & 0.614 & 2.190 \\
  \quad\quad OpenAI & 1,946 & 0.000 & 1.000 & -2.446 & -0.556 & 0.389 & 0.862 & 1.807 \\
  \quad\quad Qwen & 1,946 & -0.000 & 1.000 & -1.873 & -0.869 & 0.134 & 0.636 & 2.141 \\
  \quad\quad Llama & 1,946 & 0.000 & 1.000 & -2.277 & -0.784 & -0.287 & 0.709 & 1.704 \\
  \quad\quad DeepSeek & 1,946 & -0.000 & 1.000 & -2.021 & -0.646 & -0.095 & 1.006 & 1.831 \\
  \quad\quad Anthropic & 1,946 & -0.000 & 1.000 & -2.921 & -1.180 & 0.562 & 0.562 & 0.910 \\
  \\[-6pt]
  \quad \textit{Respect} \\[1pt]
  \quad\quad Mistral & 1,939 & 0.000 & 1.000 & -2.101 & -0.741 & -0.060 & 0.620 & 2.660 \\
  \quad\quad Gemini & 1,946 & -0.000 & 1.000 & -0.884 & -0.884 & 0.232 & 0.232 & 3.583 \\
  \quad\quad OpenAI & 1,946 & 0.000 & 1.000 & -2.289 & -0.888 & 0.162 & 0.862 & 1.913 \\
  \quad\quad Qwen & 1,946 & -0.000 & 1.000 & -2.005 & -0.652 & 0.250 & 0.701 & 1.828 \\
  \quad\quad Llama & 1,946 & 0.000 & 1.000 & -1.432 & -0.846 & -0.259 & 0.914 & 3.261 \\
  \quad\quad DeepSeek & 1,946 & -0.000 & 1.000 & -1.223 & -0.883 & -0.033 & 0.817 & 2.856 \\
  \quad\quad Anthropic & 1,946 & 0.000 & 1.000 & -2.392 & -0.876 & 0.639 & 0.639 & 2.155 \\
  \\[-6pt]
  \quad \textit{Sentiment} \\[1pt]
  \quad\quad Mistral & 1,946 & -0.000 & 1.000 & -3.385 & -0.779 & 0.524 & 0.524 & 1.393 \\
  \quad\quad Gemini & 1,946 & -0.000 & 1.000 & -2.878 & -0.856 & 0.431 & 0.983 & 1.351 \\
  \quad\quad OpenAI & 1,946 & -0.000 & 1.000 & -2.571 & -0.581 & 0.215 & 1.011 & 1.807 \\
  \quad\quad Qwen & 1,946 & -0.000 & 1.000 & -3.447 & -0.836 & 0.469 & 0.469 & 1.340 \\
  \quad\quad Llama & 1,946 & 0.000 & 1.000 & -3.454 & 0.081 & 0.081 & 0.912 & 1.120 \\
  \quad\quad DeepSeek & 1,946 & -0.000 & 1.000 & -3.206 & -0.781 & 0.432 & 0.836 & 1.240 \\
  \quad\quad Anthropic & 1,946 & 0.000 & 1.000 & -2.683 & -0.545 & 0.168 & 0.643 & 1.831 \\
  \\[-6pt]
  \quad \textit{Specificity} \\[1pt]
  \quad\quad Mistral & 1,946 & -0.000 & 1.000 & -2.039 & -0.671 & 0.356 & 0.698 & 1.383 \\
  \quad\quad Gemini & 1,946 & -0.000 & 1.000 & -2.943 & -0.497 & 0.318 & 0.318 & 1.541 \\
  \quad\quad OpenAI & 1,946 & -0.000 & 1.000 & -2.831 & -0.149 & -0.149 & 0.745 & 1.639 \\
  \quad\quad Qwen & 1,946 & 0.000 & 1.000 & -2.619 & -0.198 & 0.609 & 0.609 & 1.416 \\
  \quad\quad Llama & 1,946 & -0.000 & 1.000 & -4.594 & -0.345 & 0.409 & 0.409 & 1.232 \\
  \quad\quad DeepSeek & 1,946 & -0.000 & 1.000 & -2.141 & -0.290 & 0.081 & 0.821 & 1.562 \\
  \quad\quad Anthropic & 1,946 & -0.000 & 1.000 & -2.915 & -0.501 & -0.501 & 1.108 & 1.108 \\
  \\[-6pt]
  \quad \textit{Specificity Ratio} \\[1pt]
  \quad\quad Mistral & 1,946 & -0.000 & 1.000 & -1.811 & -0.459 & -0.213 & 0.771 & 2.123 \\
  \quad\quad Gemini & 1,946 & 0.000 & 1.000 & -2.191 & -0.338 & -0.232 & 0.912 & 2.004 \\
  \quad\quad OpenAI & 1,946 & -0.000 & 1.000 & -0.783 & -0.369 & -0.231 & -0.002 & 5.514 \\
  \quad\quad Qwen & 1,946 & -0.000 & 1.000 & -1.228 & -0.503 & -0.503 & 0.223 & 3.687 \\
  \quad\quad Llama & 1,878 & 0.000 & 1.000 & -1.337 & -0.764 & -0.286 & 0.442 & 3.013 \\
  \quad\quad DeepSeek & 1,946 & -0.000 & 1.000 & -1.750 & -0.395 & -0.395 & 0.535 & 2.395 \\
  \quad\quad Anthropic & 1,946 & -0.000 & 1.000 & -1.210 & -0.441 & -0.441 & 1.096 & 2.633 \\
  \\[-6pt]
  \quad \textit{Teamwork} \\[1pt]
  \quad\quad Mistral & 1,939 & 0.000 & 1.000 & -2.530 & -0.875 & 0.228 & 0.779 & 1.883 \\
  \quad\quad Gemini & 1,946 & -0.000 & 1.000 & -1.576 & -0.822 & -0.069 & 0.684 & 2.944 \\
  \quad\quad OpenAI & 1,946 & -0.000 & 1.000 & -2.581 & -0.685 & 0.073 & 0.832 & 1.590 \\
  \quad\quad Qwen & 1,946 & 0.000 & 1.000 & -1.855 & -0.857 & 0.141 & 0.640 & 2.137 \\
  \quad\quad Llama & 1,946 & -0.000 & 1.000 & -1.996 & -0.926 & 0.145 & 0.680 & 2.285 \\
  \quad\quad DeepSeek & 1,946 & -0.000 & 1.000 & -1.885 & -0.991 & -0.098 & 0.497 & 2.284 \\
  \quad\quad Anthropic & 1,946 & 0.000 & 1.000 & -2.422 & -1.056 & 0.310 & 0.310 & 1.676 \\
  \\[-6pt]
  \quad \textit{Uncertainty} \\[1pt]
  \quad\quad Mistral & 1,946 & -0.000 & 1.000 & -3.382 & -0.160 & -0.160 & -0.160 & 1.451 \\
  \quad\quad Gemini & 1,946 & -0.000 & 1.000 & -3.774 & -0.433 & 0.680 & 0.680 & 1.794 \\
  \quad\quad OpenAI & 1,946 & -0.000 & 1.000 & -2.172 & -1.246 & 0.608 & 0.608 & 1.534 \\
  \quad\quad Qwen & 1,946 & 0.000 & 1.000 & -1.902 & -0.947 & 0.008 & 0.962 & 3.827 \\
  \quad\quad Llama & 1,946 & 0.000 & 1.000 & -3.956 & -1.006 & 0.398 & 0.609 & 1.505 \\
  \quad\quad DeepSeek & 1,946 & 0.000 & 1.000 & -2.034 & -0.712 & 0.611 & 0.611 & 1.273 \\
  \quad\quad Anthropic & 1,946 & 0.000 & 1.000 & -1.198 & -0.844 & -0.135 & 1.282 & 2.345 \\
  \\[-6pt]

\end{longtable}
\end{ThreePartTable}

\clearpage
\section{Dispersion Regression Tables}
\label{app:dispersion}

\inputregtable{Assets/tables/regression_dispersion_range}

\clearpage
\section{Self-Reported Confidence}
\label{app:confidence}

Each model reports a confidence level (0--1) with every score. Panel (a) shows
the pooled distribution: confidence is tightly clustered---about 30\% of all
scores carry a confidence of exactly 0.80, more than four-fifths lie between 0.70 and 0.95, and
provider medians range from 0.80 to 0.90---so the field carries limited
cross-sectional information. Panel (b) compares inter-provider agreement on all
calls with agreement on the subset of calls where \textit{both} models in a pair
report confidence at or above their own median for that construct---a loose,
provider-relative cut. Even this mild form of confidence screening does not
improve agreement: across the eleven constructs for which the conditional
correlations are computable, the mean pairwise Spearman correlation falls from
0.52 to 0.48, and no construct improves meaningfully. For climate risk and
political risk the high-confidence subsample is degenerate---models are most
confident precisely on calls where they assign the construct a score of
zero---so almost no provider pair retains any score variation. The decline is
monotone in the strictness of the cut: a fixed threshold of 0.8, under which all
thirteen constructs remain computable, yields a mean of 0.46; requiring
confidence strictly above each provider's own median yields 0.35; and a
top-quartile cut yields 0.33 over the six constructs that remain computable. Consistent with this, a model's confidence is essentially uninformative
about its own deviation from the seven-model consensus (mean Spearman
correlation of $-0.08$ across provider--construct cells). Self-reported
confidence therefore does not identify a subset of calls on which LLM-based
measures can be trusted to agree, no matter where the threshold is drawn.

\begin{figure}[H]
  \centering
  \includegraphics[width=\textwidth]{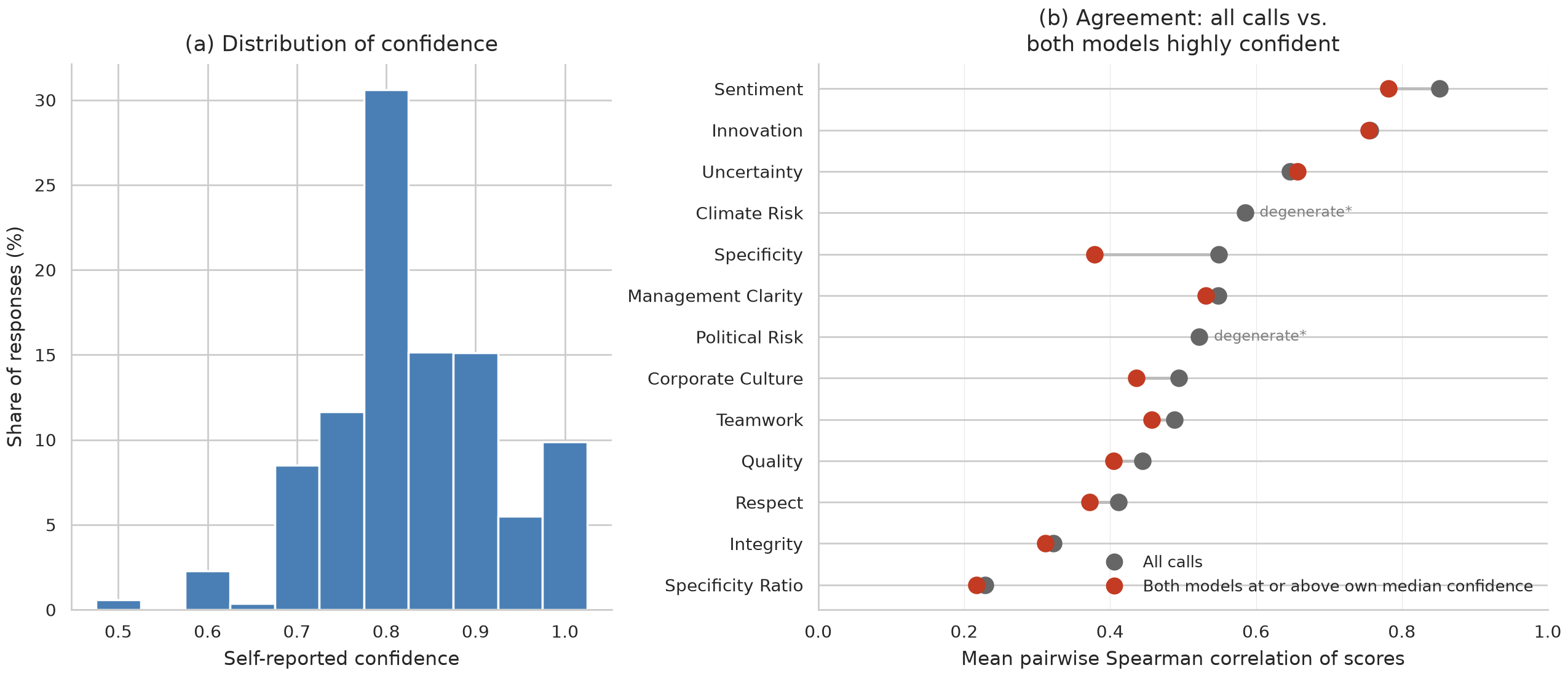}
  \caption{Self-reported confidence: distribution and conditional agreement.
  Panel (a): pooled distribution of the confidence field across the seven
  providers and thirteen constructs (a single invalid response with confidence
  $>1$ is excluded).
  Panel (b): mean pairwise Spearman correlation of scores per construct, using
  all calls (gray) versus only calls where both models of a pair report
  confidence at or above their own provider--construct median (red);
  ``degenerate'' marks constructs where nearly all high-confidence pairs lack
  score variation. Stricter definitions of high confidence lower the
  conditional correlations further.}
  \label{fig:confidence}
\end{figure}

\clearpage
\FloatBarrier
\section{LLM Models}
\label{app:models}

Table~\ref{tab:llm_models} lists the seven large language models used to score
every earnings-call text, together with the exact model identifier and the
route through which each model was queried. The set is held fixed across all
calls and all constructs, so any disagreement among the models reflects the
models themselves rather than differences in coverage. Six of the seven models
are accessed through the OpenRouter gateway using an OpenAI-compatible
interface; OpenAI's GPT is queried directly through the native OpenAI API. All
models are prompted identically (see \autoref{app:prompts}) with the same
decoding settings.

\begin{table}[H]
  \centering
  \footnotesize
  \caption{Large language models used in the analysis.}
  \label{tab:llm_models}
  \begin{adjustbox}{max width=\textwidth}
  \begin{tabular}{lllll}
    \toprule
    Provider & Developer & Family & Model identifier & Access route \\
    \midrule
    OpenAI    & OpenAI    & GPT      & \texttt{gpt-4o-mini}                              & OpenAI API (native SDK) \\
    Anthropic & Anthropic & Claude   & \texttt{anthropic/claude-3-haiku}                 & OpenRouter \\
    Gemini    & Google    & Gemini   & \texttt{google/gemini-2.5-flash-lite}             & OpenRouter \\
    Llama     & Meta      & Llama    & \texttt{meta-llama/llama-3.3-70b-instruct}        & OpenRouter \\
    Mistral   & Mistral   & Mistral  & \texttt{mistralai/mistral-small-24b-instruct-2501}& OpenRouter \\
    DeepSeek  & DeepSeek  & DeepSeek & \texttt{deepseek/deepseek-chat}                   & OpenRouter \\
    Qwen      & Alibaba   & Qwen     & \texttt{qwen/qwen-2.5-7b-instruct}                & OpenRouter \\
    \bottomrule
  \end{tabular}
  \end{adjustbox}

  \vspace{0.5em}
  \begin{minipage}{0.95\textwidth}
    \footnotesize\textit{Notes.} ``Provider'' is the short label used to
    identify each model throughout the tables and figures. ``Model identifier''
    is the exact string passed to the API. OpenRouter
    (\texttt{openrouter.ai}) is an OpenAI-compatible gateway that routes
    requests to the underlying model provider; OpenAI's GPT is queried directly
    through the native OpenAI SDK. The seven models come from seven independent
    developers and are held fixed across every earnings call and every textual
    construct.
  \end{minipage}
\end{table}

\clearpage
\FloatBarrier
\section{Prompt Templates}
\label{app:prompts}

The following templates were used verbatim for all LLM providers.
The placeholder \texttt{\{text\}} is replaced with the earnings-call
transcript segment at inference time.

\subsection{Sentiment}
\label{app:prompt:sentiment}

\begin{Verbatim}[fontsize=\footnotesize,breaklines=true,frame=single]
You are an analyst tasked with quantifying sentiment from a passage of text.

Important constraints:

- Base your assessment exclusively on the text provided below.
- Do not use any external knowledge, background information, or assumptions
  about the firm, industry, economy, or time period.
- Do not infer intent beyond what is explicitly stated in the text.

Sentiment Definition:

- Sentiment captures the overall positivity vs. negativity expressed in the
  text regarding firm performance, prospects, or conditions.

Sentiment Scale:

- Sentiment Score: -1.00 (very negative) to +1.00 (very positive),
  where 0.00 is neutral.

Confidence definition:

- Confidence reflects how certain you are that the numeric sentiment score
  accurately captures the overall positivity versus negativity expressed in
  the provided text.
- Confidence is not about how extreme the sentiment is, only about the
  certainty of the measurement given the text.

JSON Output format (strict):

- Sentiment Score: <number>
- Confidence: <number between 0 and 1>
- Justification: <1-2 sentences citing specific wording from the text>

Text to analyze:

{text}
\end{Verbatim}

\subsection{Uncertainty}
\label{app:prompt:uncertainty}

\begin{Verbatim}[fontsize=\footnotesize,breaklines=true,frame=single]
You are an analyst tasked with quantifying uncertainty about the firm's
future outlook from a passage of text.

Important constraints:

- Use only the text provided below.
- Do not incorporate any external data, historical outcomes, or industry
  knowledge.

Uncertainty Score Definition (critical):

- Uncertainty reflects the degree of ambiguity, lack of precision, hedging,
  conditional language, or unresolved outcomes regarding the firm's future.
- This is NOT about whether expectations and the outlook are positive or
  negative.

Uncertainty Score Scale:

- 0.00 (very certain / precise outlook) to 1.00
  (very uncertain / ambiguous outlook).

Confidence definition:

- Confidence reflects how certain you are that the numeric uncertainty score
  accurately captures the degree of uncertainty about the firm's future
  outlook as expressed in the provided text.
- Confidence is not about whether expectations are positive or negative,
  only about the certainty of measuring uncertainty given the text.

JSON Output format (strict):

- Uncertainty Score: <number>
- Confidence: <number between 0 and 1>
- Justification: <1-2 sentences citing specific wording from the text>

Text to analyze:

{text}
\end{Verbatim}

\subsection{Climate Risk}
\label{app:prompt:climate}

\begin{Verbatim}[fontsize=\footnotesize,breaklines=true,frame=single]
You are an analyst tasked with quantifying climate risk exposure as expressed
in a passage of text.

Important constraints:

- Rely only on the text provided below.
- Do not use external information about climate change, regulation,
  geography, or the firm.

Climate risk exposure Definition:

- Climate risk exposure captures the extent to which the text indicates
  exposure to climate-related risks, including physical risks (e.g., extreme
  weather), transition risks (e.g., regulation, carbon costs), or
  climate-related operational disruptions.
- The measure reflects presence and salience in the text, not whether the
  firm is managing the risk well.

Climate risk exposure Scale:

- Climate Risk Exposure Score: 0.00 (no climate risk mentioned) to 1.00
  (climate risk is a major, explicit concern).

Confidence definition:

- Confidence reflects how certain you are that the numeric score accurately
  captures climate risk exposure as expressed in the provided text.
- Confidence is not about how large or severe the climate risk is, only
  about the certainty of the measurement given the text.

JSON Output format (strict):

- Climate Risk Exposure Score: <number>
- Confidence: <number between 0 and 1>
- Justification: <1-2 sentences citing specific wording from the text>

Text to analyze:

{text}
\end{Verbatim}

\subsection{Management Clarity}
\label{app:prompt:clarity}

\begin{Verbatim}[fontsize=\footnotesize,breaklines=true,frame=single]
You are an analyst tasked with quantifying management clarity in responses
from a passage of text.

Important constraints:

- Base your judgment only on the provided text.
- Do not infer intent or competence beyond what is observable in the wording.
- Do not use external benchmarks or expectations.

Management clarity Definition:

- Management clarity reflects how clear, direct, specific, and internally
  consistent the responses are.
- High clarity: concrete answers, specific explanations, minimal hedging,
  clear structure.
- Low clarity: vague language, evasive answers, excessive qualifiers, or
  failure to directly address issues.

Management clarity Scale:

- 0.00 (very unclear / evasive) to 1.00 (very clear and direct).

Confidence definition:

- Confidence reflects how certain you are that the numeric score accurately
  captures management clarity as expressed in the provided text.
- Confidence is not about how extreme the clarity/lack of clarity is, only
  about the certainty of the measurement given the text.

JSON Output format (strict):

- Management Clarity Score: <number>
- Confidence: <number between 0 and 1>
- Justification: <1-2 sentences citing specific wording from the text>

Text to analyze:

{text}
\end{Verbatim}

\subsection{Political Risk}
\label{app:prompt:political}

\begin{Verbatim}[fontsize=\footnotesize,breaklines=true,frame=single]
You are an analyst tasked with quantifying political risk exposure as
expressed in a passage of text.

Important constraints:

- Rely only on the text provided below.
- Do not use external knowledge about politics, countries, regulations, or
  current events.
- If the text contains no explicit reference to political risk (e.g.,
  elections, sanctions, tariffs, trade restrictions, government instability,
  policy uncertainty, geopolitical conflict, nationalization, government
  shutdowns), assign a score of exactly 0.00.

Political risk exposure Definition:

- Political risk exposure captures the extent to which the text indicates
  exposure to political or geopolitical forces that could affect the firm's
  performance or outlook.
- The measure reflects presence and salience in the text, not whether the
  risk is managed well.

Political risk exposure Scale:

- Political Risk Exposure Score: 0.00 (no political risk mentioned) to 1.00
  (political risk is a major, explicit concern).

Confidence definition:

- Confidence reflects how certain you are that the numeric score accurately
  captures political risk exposure as expressed in the provided text.

Output format (strict):

- Political Risk Exposure Score: <number>
- Confidence: <number between 0 and 1>
- Justification: <1-2 sentences citing specific wording from the text>

Text to analyze:

{text}
\end{Verbatim}

\subsection{Corporate Culture}
\label{app:prompt:culture}

\begin{Verbatim}[fontsize=\footnotesize,breaklines=true,frame=single]
You are an analyst tasked with quantifying corporate culture signals from
a passage of text.

Important constraints:

- Use only the text provided below.
- Do not use external knowledge about the firm or industry.
- Score only what is explicitly signaled in the wording (values, norms,
  behaviors, incentives).

Corporate culture Definition:

- Corporate culture reflects the norms and values signaled by management's
  language about how the organization behaves and what it prioritizes.

Dimensions (each scored 0.00 to 1.00):

- Innovation: emphasis on experimentation, learning, R&D, speed,
  new products/processes.
- Integrity: emphasis on ethics, compliance, transparency, doing the
  right thing.
- Quality: emphasis on reliability, craftsmanship, excellence, safety,
  customer outcomes.
- Respect: emphasis on people, inclusion, wellbeing, fair treatment,
  listening.
- Teamwork: emphasis on collaboration, shared ownership, cross-functional
  alignment.

Overall Culture Score:

- Overall Corporate Culture Score (0.00 to 1.00) summarizes the strength
  and clarity of culture signals across dimensions.

Confidence definition:

- Confidence reflects how certain you are that the scores accurately reflect
  culture signals in the text (not whether the culture is "good").

Output format (strict):

- Innovation Score: <number>
- Integrity Score: <number>
- Quality Score: <number>
- Respect Score: <number>
- Teamwork Score: <number>
- Overall Corporate Culture Score: <number>
- Confidence: <number between 0 and 1>
- Justification: <2-3 sentences citing specific wording from the text>

Text to analyze:

{text}
\end{Verbatim}

\subsection{Specificity Ratio}
\label{app:prompt:specificity_ratio}

\begin{Verbatim}[fontsize=\footnotesize,breaklines=true,frame=single]
You are an analyst tasked with measuring the specificity of a passage of
text.

Important constraints:

- Base your assessment exclusively on the text provided below.
- Do not use any external knowledge or assumptions.

Definition of Specificity:

- Specificity refers to the extent to which the text contains concrete,
  verifiable, and precise information, as opposed to general or vague
  statements.
- In this task, "specific words" are words or phrases that belong to one
  of the following seven categories:
  1. Person names
  2. Geographic locations
  3. Organizations (including companies)
  4. Percentages
  5. Monetary values
  6. Dates
  7. Times or durations

Counting rules:

- Count as "specific" only those words or phrases that clearly fall into
  one of the seven categories listed above.
- Count each named entity as one unit, regardless of the number of words
  (e.g., "Apple Inc." = 1).
- Count each numeric expression as one unit (e.g., "5%", "$10 million",
  "2024", "three years" = 1 each).
- Count only instances explicitly present in the text.

Definitions for calculation:

- Number of Specific Words = total number of words or expressions belonging
  to the seven categories above.
- Total Words = total number of words in the text.

Specificity Score:

- Specificity Score is a number between 0.00 and 1.00, defined as:
- Specificity Score = (Number of Specific Words) / (Total Words)

Confidence definition:

- Confidence reflects how certain you are that the numeric scores accurately
  capture the specificity of the provided text.
- Confidence is not about whether specificity is high or low, but about how
  certain you are in applying the rules consistently.

Output format (strict):

- Number of Specific Words: <integer>
- Specificity Score: <number between 0 and 1>
- Confidence: <number between 0 and 1>
- Justification: <list all instances of specific content you counted,
  separated by commas; if none, write "None">

Text to analyze:

{text}
\end{Verbatim}

\subsection{Specificity}
\label{app:prompt:specificity}

\begin{Verbatim}[fontsize=\footnotesize,breaklines=true,frame=single]
You are an analyst tasked with assessing the specificity of a passage of
text.

Important constraints:

- Base your assessment exclusively on the text provided below.
- Do not use any external knowledge or assumptions.

Definition of Specificity:

- Specificity refers to the extent to which the text contains concrete,
  verifiable, and precise information, as opposed to general or vague
  statements.
- In this context, specificity is driven by the presence of:
  - Person names
  - Geographic locations
  - Organizations (including companies)
  - Percentages
  - Monetary values
  - Dates
  - Times or durations

Guidance:

- Texts with many such elements should receive higher specificity scores.
- Texts that are general, abstract, or lack concrete details should receive
  lower scores.
- Base your assessment on the overall prevalence of these elements in the
  text.

Specificity Score Scale:

- Specificity Score: 0.00 (very vague / no concrete details) to 1.00
  (highly specific / dense with concrete details).

Confidence definition:

- Confidence reflects how certain you are that the numeric score accurately
  captures the specificity of the provided text.
- Confidence is not about whether specificity is high or low, but about how
  certain you are in judging the prevalence of specific elements from the
  text alone.

Output format (strict):

- Specificity Score: <number between 0 and 1>
- Confidence: <number between 0 and 1>
- Justification: <1-2 sentences explaining what types of specific elements
  were present or missing>

Text to analyze:

{text}
\end{Verbatim}

\end{document}